# Sparse Autoencoders Can Capture Language-Specific Concepts Across Diverse Languages

**Lyzander Marciano Andrylie[1], Inaya Rahmanisa[1], Mahardika Krisna Ihsani[2]**
**Alfan Farizki Wicaksono[1], Haryo Akbarianto Wibowo[2], Alham Fikri Aji[2]**

[1]Faculty of Computer Science, Universitas Indonesia
[2]Department of Natural Language Processing, MBZUAI
{lyzander.marciano,inaya.rahmanisa}@ui.ac.id, alfan@cs.ui.ac.id
{mahardika.ihsani,haryo.wibowo,alham.fikri}@mbzuai.ac.ae

## Abstract

Understanding the multilingual mechanisms of large language models (LLMs) provides insight into how they process different languages, yet this remains challenging. Existing studies often focus on individual neurons, but their polysemantic nature makes it difficult to isolate language-specific units from cross-lingual representations. To address this, we explore sparse autoencoders (SAEs) for their ability to learn monosemantic features that represent concrete and abstract concepts across languages in LLMs. While some of these features are language-independent, the presence of language-specific features remains underexplored. In this work, we introduce *SAE-LAPE*, a method based on feature activation probability, to identify language-specific features within the feed-forward network. We find that many such features predominantly appear in the middle to late layers of the model and are interpretable. These features influence the model's multilingual performance and language output, and can be used for language identification with performance comparable to fastText, along with more interpretability. Our code and complete figures are available here.

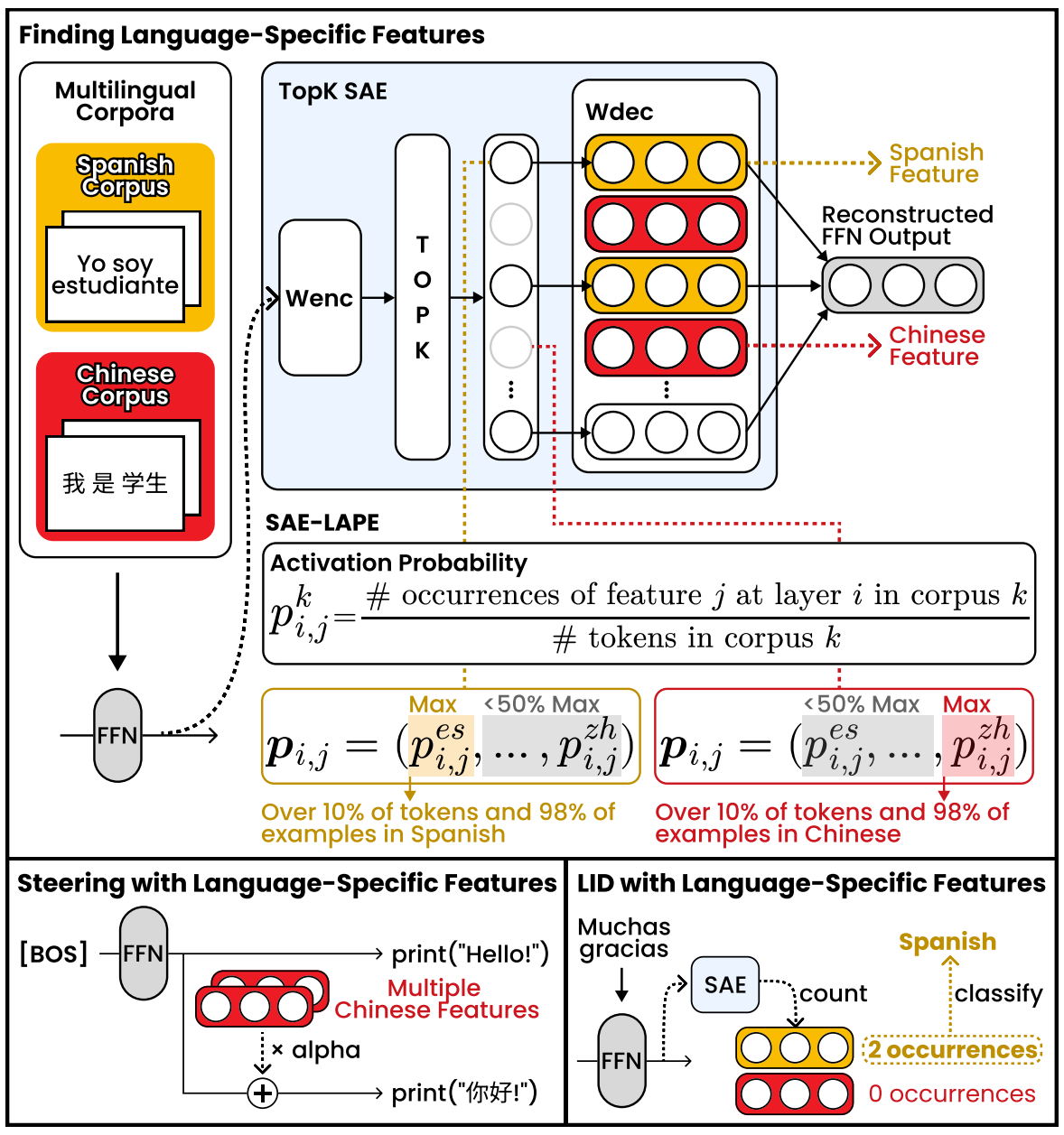

Figure 1: An illustration of language-specific features. **Top:** Finding language-specific features that are often activated in a target language using *SAE-LAPE*. **Bottom Left:** Steering with multiple language-specific features simultaneously by multiplying them with a scaling factor $\alpha$ and adding the result to the FFN output. **Bottom Right:** Language identification (LID) by counting the occurrences of language-specific features in an SAE.

## 1 Introduction

Understanding the multilingual mechanisms of LLMs offers insight into how they process language across diverse linguistic contexts. However, such mechanisms remain challenging to interpret. Existing studies (Tang et al., 2024; Kojima et al., 2024) primarily focus on individual neurons, yet these neurons are often polysemantic, meaning they may activate in response to multiple unrelated concepts (Elhage et al., 2022). This characteristic limits both interpretability and reliability, as it becomes difficult to isolate language-specific units from cross-lingual representations.

A promising approach to address the issue of the polysemous nature of neurons is the use of sparse autoencoders (SAEs), an unsupervised method that can learn monosemantic features in the form of vectors that represent single concepts (Huben et al., 2024). SAEs have recently been shown to be capable of detecting particular concepts in LLMs, such as "Golden Gate Bridge"-related concepts, abstract grammatical concepts shared across languages (Templeton et al., 2024; Brinkmann et al., 2025), known entities (Ferrando et al., 2025), and "artificial text"-related concepts used for classifying such text (Kuznetsov et al., 2025). While some features appear to be language-agnostic, only a few studies have explored the existence of language-specific features, as discussed in Section 6.

In this study, we investigate the presence of language-specific features by identifying them via SAEs and using them as units for analyzing the multilinguality of LLMs. By adapting the *language activation probability entropy (LAPE)* method proposed by Tang et al. (2024), we introduce *SAE-LAPE*, a method to identify language-specific features in SAEs trained on the feed-forward network (FFN) outputs of each layer. We further analyze the concepts represented by these features through automated interpretability techniques proposed by Paulo et al. (2025). Finally, we provide empirical evidence showing that the features are indeed language-specific. This leads to several findings:

- **Language-specific features exist and are interpretable.** These features, predominantly active in a single language, are primarily located in the middle to late layers. Additionally, these features are interpretable, as their interpretation scores fall within the acceptable range reported by Paulo et al. (2025) across models.
- **Language-specific features influence both the perplexity and the language output of LLMs.** Suppressing these features increases perplexity in the targeted language with relatively little impact on others, while activating them has a minimal effect on that language but can influence others. Moreover, activating these features can shift the output toward the targeted language.
- **Language-specific features can be used as a language classifier.** Because these features activate on aspects of particular languages, we use them for a language identification (LID) task; this approach achieves results comparable to fastText (Joulin et al., 2017), yet is more interpretable since each feature tends to be monosemantic with its own interpretation.

# 2 Background

## 2.1 TopK SAEs and JumpReLU SAEs

SAEs are an unsupervised method used to discover interpretable feature directions in LLMs by reconstructing model activations as a sparse linear combination of learned overcomplete basis vectors representing these directions. The sparsity constraint encourages the model to represent each activation using only a small number of these feature directions, leading to more monosemous representations. To achieve this, SAEs use an encoder to map the activations $\boldsymbol{x} \in \mathbb{R}^d$ into a sparse, higher-dimensional representation $\boldsymbol{z} \in \mathbb{R}^n$, and a decoder to project this representation back into the original activations. This process can be described as follows:

$$\boldsymbol{z} = \text{ReLU}(\boldsymbol{W}_{\text{enc}}(\boldsymbol{x} - \boldsymbol{b}_{\text{dec}}) + \boldsymbol{b}_{\text{enc}})\,,$$

$$\hat{\boldsymbol{x}} = \boldsymbol{W}_{\text{dec}}\boldsymbol{z} + \boldsymbol{b}_{\text{dec}}\,,$$

where $\boldsymbol{W}_{\text{enc}} \in \mathbb{R}^{n \times d}$ is an encoder matrix; $\boldsymbol{W}_{\text{dec}} \in \mathbb{R}^{d \times n}$ is a decoder matrix; $\boldsymbol{b}_{\text{enc}} \in \mathbb{R}^n$ and $\boldsymbol{b}_{\text{dec}} \in \mathbb{R}^d$ are the bias vectors for the encoder and decoder, respectively; $\hat{\boldsymbol{x}}$ is the reconstruction of $\boldsymbol{x}$; and $\text{ReLU}(x) = \max(0, x)$ is an activation function (Glorot et al., 2011). We define $\boldsymbol{z}_j$ as the activation of feature $j$ and $\boldsymbol{d}^j = \boldsymbol{W}_{\text{dec}}[:, j] \in \mathbb{R}^d$ as the direction of feature $j$.

In this study, we use TopK SAEs (Gao et al., 2025), an improved variant that provides a better reconstruction-sparsity trade-off and increases feature monosemanticity. TopK SAEs enforce sparsity by retaining only the $k$ largest values in the SAE activations $\boldsymbol{z} \in \mathbb{R}^n$ and zeroing the rest. In our setup, ReLU is applied prior to TopK to ensure non-negative activations. Moreover, we use JumpReLU SAEs (Rajamanoharan et al., 2024), which improve reconstruction fidelity by replacing the standard ReLU with JumpReLU (Erichson et al., 2020), defined as $\text{JumpReLU}(x) = xH(x - \theta)$, where $H$ is the Heaviside step function and $\theta$ is a learned threshold.

## 2.2 High Frequency Latents

High frequency latents (HFLs) are SAE features that appear in over 10% of tokens from the entire corpora, constituting a significant proportion (Sun et al., 2025). As noted by prior work, SAEs often still learn a few HFLs that may appear less interpretable, even under greater sparsity constraints (Cunningham and Conerly, 2024; Rajamanoharan et al., 2024; Sun et al., 2025). However, Sun et al. (2025) argued that these features may correspond to genuine dense language model features and can be somewhat interpretable, as suggested by their preliminary analyses.

# 3 Method

## 3.1 Finding Language-Specific Features

By adapting the method proposed by Tang et al. (2024), we introduce an approach called *SAE-LAPE* to identify language-specific features that are also HFLs in the FFN, where language-specific information resides (Tang et al., 2024; Kojima et al., 2024). Specifically, we provide an LLM with multilingual corpora, and we compute the activation probability of feature $\boldsymbol{d}^j$ from an SAE at layer $i$ for each language $k$ as follows:

$$p^k_{i,j} = \mathbb{E}\left(\mathbb{I}(\boldsymbol{z}^i_j > 0) \mid \text{language } k\text{'s corpus}\right),$$

where $\mathbb{E}$ is the expectation over the tokens in language $k$'s corpus; $\mathbb{I}$ is the indicator function, where a feature is considered active if its activation is nonzero; and $\boldsymbol{z}^i_j$ is the activation associated with feature $\boldsymbol{d}^j$ at layer $i$. Subsequently, we obtain the vector $\boldsymbol{p}_{i,j} = (p^1_{i,j}, \ldots, p^l_{i,j})$ for $l$ languages, and then apply L1-normalization to produce a valid probability distribution $\boldsymbol{p}'_{i,j} = (p'^1_{i,j}, \ldots, p'^l_{i,j})$. We compute the entropy of this distribution, referred to as *language activation probability entropy (LAPE)*, to measure the degree of language specificity, where a lower value indicates greater specificity:

$$\text{LAPE}_{i,j} = -\sum_{k=1}^{l} p'^k_{i,j} \ln(p'^k_{i,j}).$$

Thus far, our implementation follows LAPE (Tang et al., 2024). This method was originally used to identify language-specific neurons in the FFN activations (see Appendix A.2 for details). However, SAE activations are sparser than FFN activations. In addition, the features we aim to find are a type of HFL. Therefore, we need to modify the original method to address these issues.

To find HFLs, we consider only features active in more than 10% of tokens and at least $N$% of examples in at least one language-specific corpus, ensuring that the concepts are predominantly associated with those languages. Next, we dynamically identify features as specific to one or more languages when their activation probability $p^k_{i,j}$ is at least $T$% of $\max(\boldsymbol{p}_{i,j})$. Lastly, we select language-specific features as those specific to a single language.

### 3.2 Steering with Language-Specific Features

Steering with SAE features is a form of activation steering (Turner et al., 2024), in which model activations are modified during a forward pass using an SAE feature to influence downstream behavior. Generally, we can activate a concept in LLMs by adding a feature, or suppress it by subtracting one, at every token position (Templeton et al., 2024; Durmus et al., 2024):

$$\boldsymbol{x}_{\text{new}} = \boldsymbol{x} + \alpha \boldsymbol{z}^{\max}_j \boldsymbol{d}^j,$$

where $\alpha \in \mathbb{R}$ is a scaling factor and $\boldsymbol{z}^{\max}_j$ is the maximum value of the feature of interest $\boldsymbol{d}^j$ in the multilingual corpora.

In contrast to previous studies (e.g., Templeton et al., 2024; Brinkmann et al., 2025), where interventions were typically conducted on a single feature in a middle layer, we intervene on multiple features across multiple layers. We argue that this method works on language-specific features due to their tendency, as HFLs, to naturally and frequently co-occur within a specific language (see Section 5.1). We also use the same method with language-specific neurons as a baseline (see Appendix A.3).

### 3.3 LID with Language-Specific Features

Bricken et al. (2023) define the sensitivity of an SAE feature as the reliable activation of the feature when its associated concept is present in a given example. With language as the concept of interest, we use language-specific features for language classification to assess whether examples in a particular language predominantly activate the corresponding features over those of other languages.

Specifically, to classify languages using SAE features for a text input with FFN output $\boldsymbol{X} = (\boldsymbol{x}_1, \ldots, \boldsymbol{x}_t)$ consisting of $t$ tokens, we compute the score for each language $k$ as follows:

$$s_k = \sum_{i \in \mathcal{L}} \sum_{\boldsymbol{x} \in \boldsymbol{X}} \sum_{j \in \mathcal{F}^i_k} \mathbb{I}(\boldsymbol{z}^i_{\boldsymbol{x},j} > 0),$$

where $\mathcal{L}$ is the set of all layers; $\mathcal{F}^i_k$ is the set of SAE feature indices associated with language $k$ at layer $i$; $\boldsymbol{z}^i_{\boldsymbol{x},j}$ is the activation value of feature $\boldsymbol{d}^j$ for input $\boldsymbol{x}$ at layer $i$; and $\mathbb{I}(\boldsymbol{z}^i_{\boldsymbol{x},j} > 0)$ is the indicator function to ensure that only active features are considered. Then, we predict the language by selecting the index with the highest score in the vector $\boldsymbol{s} = (s_1, \ldots, s_l)$ over $l$ languages. We also use the same method with language-specific neurons as a baseline (see Appendix A.4).

## 4 Experimental Setup

### 4.1 Models

We adopt Llama 3.2 1B (Meta AI, 2024b) as the primary model in this study due to its multilingual capabilities, as it explicitly supports eight languages: English (en), German (de), French (fr), Italian (it), Portuguese (pt), Hindi (hi), Spanish (es), and Thai (th). It has also been trained on a broader collection of languages beyond these officially supported ones (Meta AI, 2024a). For this model, we use the publicly available TopK SAEs[1] released by EleutherAI

[1] https://huggingface.co/EleutherAI/sae-Llama-3.2-1B-131k

(2024) for Llama 3.2 1B, with $k = 32$ and an expansion factor of 64, resulting in 131K features per layer. These SAEs are trained on the RedPajama-v2 (Weber et al., 2024) multilingual corpus using the model's FFN outputs across all layers. See Appendix A.5 for the SAE configuration details.

We also use Gemma 2 (2B and 9B) (Team et al., 2024) in a subset of experiments to evaluate cross-model generalizability. Although primarily trained on English data, these models exhibit some multilingual capabilities. For both models, we use the JumpReLU SAEs from Gemma Scope[2] (Lieberum et al., 2024). We adopt the canonical variant, with 65K features per layer for Gemma 2 2B and 131K for Gemma 2 9B. These SAEs are trained mainly on English data using each model's FFN outputs across all layers.

### 4.2 Datasets

To identify and analyze language-specific features, we focus on 15 languages across three categories: (1) 8 languages explicitly supported by Llama 3.2; (2) 4 languages where the primary model performs reasonably well[3]: Bulgarian (bg), Russian (ru), Turkish (tr), and Vietnamese (vi); and (3) 3 languages for broader typological coverage: Japanese (ja), Korean (ko), and Chinese (zh).

To create a language-specific corpus, we use training sets from three datasets: FLORES+ (NLLB Team et al., 2024), PAWS-X (Yang et al., 2019), and XNLI (Conneau et al., 2018). This yields multilingual corpora with 1,000 to 3,000 examples (approximately 40,000 to 120,000 tokens) per language.

### 4.3 Interpreting Language-Specific Features

To interpret language-specific features across layers, we use automated interpretability tools from Paulo et al. (2025). Specifically, we ask an LLM to generate an explanation of a given feature based on a set of its activating examples. Then, another LLM is tasked with classifying whether the feature is active in new examples based on that explanation. We then report the classification accuracy. Please refer to Appendix A.7 for more details on this evaluation.

We apply two scoring methods: detection scoring, which evaluates the entire sentence, and fuzzing scoring, which focuses only on randomly selected highlighted words within the sentence. Both methods have a random baseline accuracy of 50% and a median accuracy of 70–80% across models, as reported by Paulo et al. (2025).

### 4.4 Experiment Design

We first identify language-specific features, then use them for steering and detection to provide evidence that they are truly language-specific.

**Finding Language-Specific Features.** We use *SAE-LAPE* to identify language-specific features by setting the hyperparameters described in Section 3.1 to $N = 98$, so that language-specific features are active for the majority of examples, and $T = 50$, to ensure that no other language has an activation probability close to half of the maximum, thus reinforcing their language specificity. We then analyze both their properties and interpretations.

**Steering with Language-Specific Features.** We conduct two experiments:

1. **Perplexity change**: We evaluate multilingual modeling capabilities by measuring perplexity (PPL) on the FLORES+ test set. We then compare the PPL of the model steered using the identified features with that of the language-specific neuron baseline to assess their influence.
2. **Unconditional text generation**: We steer the model with the identified features without an input prompt (i.e., only a `[BOS]` token) 100 times, each with a random seed from 0 to 99, to observe their effect on the model's output. In addition, we use Llama 3.2's default decoding strategy: random sampling with `temperature` $= 0.6$ and `top_p` $= 0.9$.

**LID with Language-Specific Features.** We evaluate whether the identified language-specific features can accurately detect the language of a given input and compare their performance against the language-specific neuron baseline and fastText (Joulin et al., 2017). For this purpose, we utilize the WiLI-2018 dataset (Thoma, 2018) to identify the 15 analyzed languages.

## 5 Results and Discussion

### 5.1 Finding Language-Specific Features

In total, using *SAE-LAPE*, we identified 540 language-specific features out of two million across layers in Llama 3.2 1B. Figure 2 shows that most features are concentrated in the middle to late layers[4], with similar trends observed in Gemma 2, as shown in Figures 18–19 in Appendix A.15.3.

[2] `https://huggingface.co/google/gemma-scope`

[3] Over 40% accuracy on the XNLI task, comparable to the explicitly supported ones (see Table 7 in Appendix A).

[4] This distribution resembles that of language-specific neurons, as shown in Figure 12 in Appendix A.

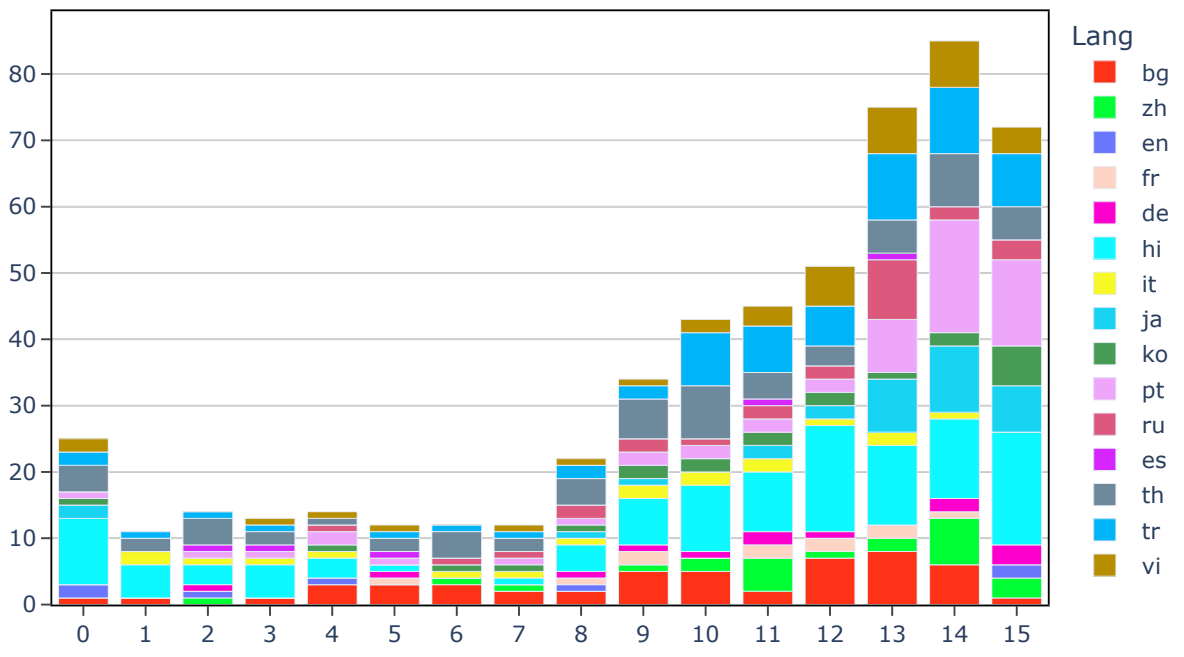

Figure 2: Distribution of language-specific features identified by *SAE-LAPE* across layers of Llama 3.2 1B. Most features appear in the middle to late layers, indicating that language-specific concepts are predominantly captured in deeper layers.

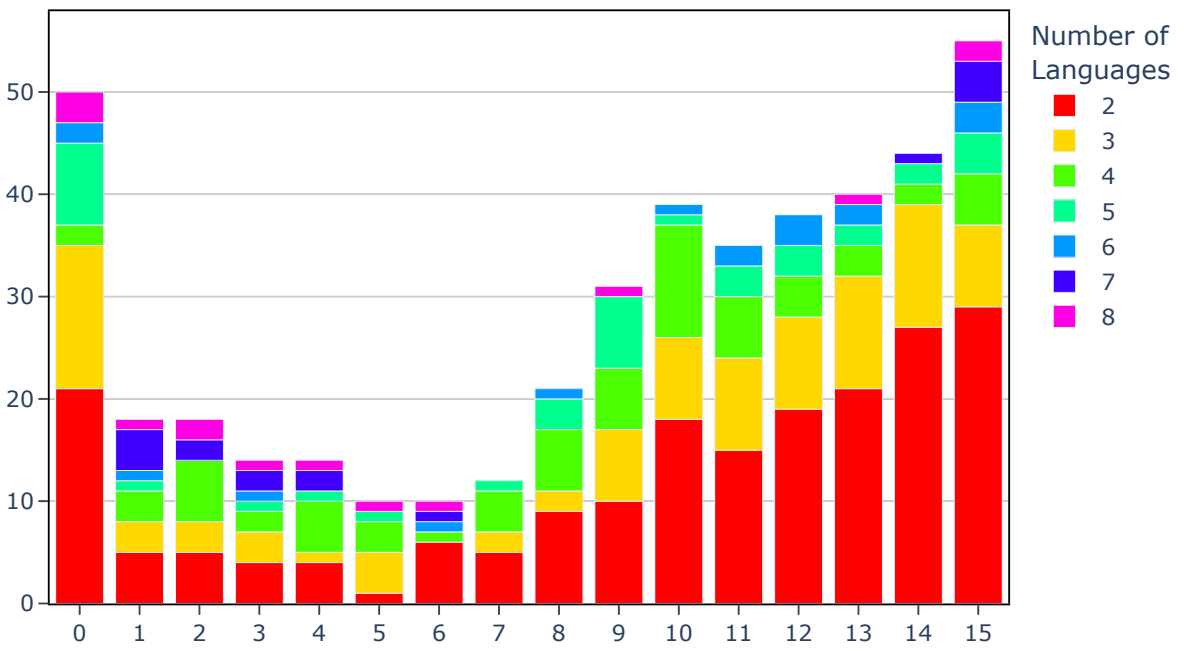

Figure 3: Distribution of language-shared features across 2–8 languages identified by *SAE-LAPE* in Llama 3.2 1B. Most features are concentrated in the middle to late layers, a trend similar to that of language-specific features.

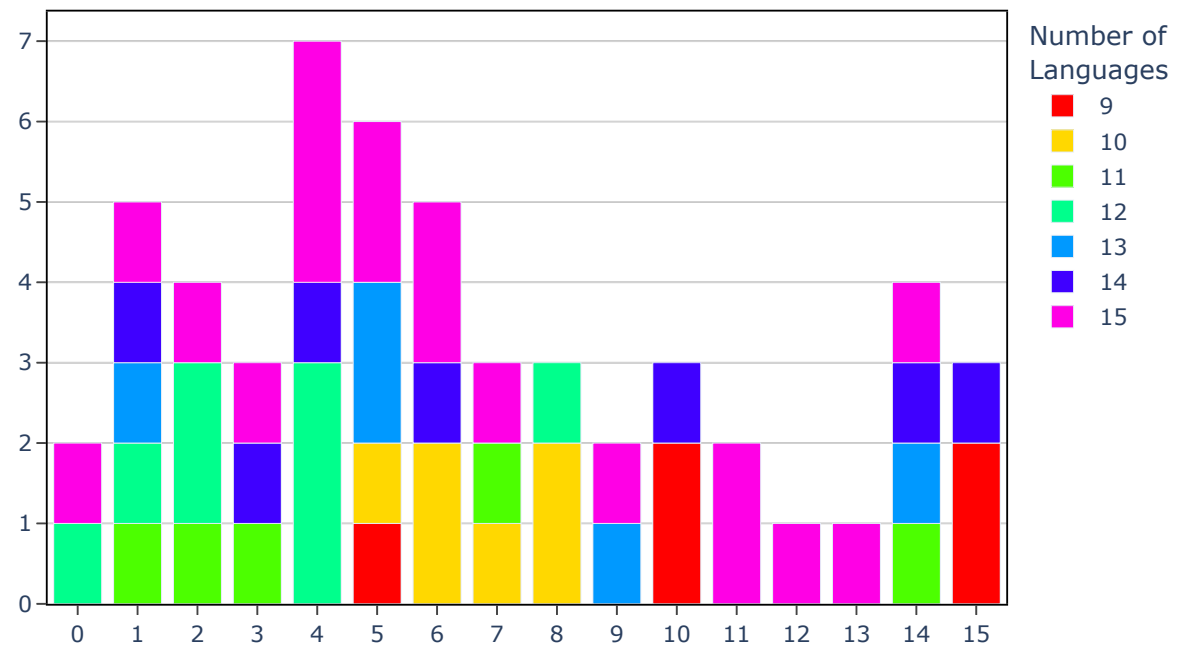

Figure 4: Distribution of language-shared features across 9–15 languages identified by *SAE-LAPE* in Llama 3.2 1B. Most features are concentrated in the early to middle layers, indicating that language-shared concepts are primarily captured in early to middle layers.

In addition, language-shared features across 2–8 languages (approximately half of the analyzed languages) exhibit a pattern consistent with that of language-specific features, as shown in Figure 3. However, for features shared across 9–15 languages, as shown in Figure 4, a contrasting pattern emerges, with these features concentrated in the early to middle layers. This phenomenon supports Kojima et al.; Tang et al.'s (2024; 2024) view that the early layers of decoder-based LLMs handle cross-lingual transfer by converting lexical and syntactic inputs into language-independent semantic representations, the middle layers process these representations, and the late layers translate them back to the target language's syntax and lexicon. See Appendix A.8 for details on language-shared features.

Further supporting this view, a logit lens-inspired analysis (nostalgebraist, 2020), which involved multiplying language-specific features by the token unembedding matrix, shows that features from shallower layers tend to map to more generic tokens (often in English or mixed-language), whereas features from deeper layers increasingly correspond to language-specific tokens (see Figures 35–43 in Appendix A). This progressive specialization suggests that linguistic information becomes increasingly language-specific as model depth increases. This method is detailed in Appendix A.9.

To better understand the nature of these features, we analyze their properties. We observe that language-specific features tend to co-occur naturally and frequently across tokens within a specific language (see Figures 44–46 in Appendix A), and some exhibit moderate to high Pearson correlations in their activation values (see Figures 47–49 in Appendix A). These properties enable the simultaneous steering of multiple language-specific features, as this approach aligns with their natural co-activation patterns.

### 5.1.1 Feature Interpretation Scores

Based on the automated evaluation for interpretable features proposed by Paulo et al. (2025), our SAE features are interpretable, as they have median accuracies of 0.75 and 0.76 for detection and fuzzing scoring, respectively, similar to the reported ones. Moreover, they achieve average accuracy, F1, precision, and recall scores of 0.74, 0.70, 0.85, and 0.64, respectively, across languages for both tasks (see Table 8 in Appendix A). This indicates that most of the generated explanations are faithful, as they can be used to determine whether the given sentences or tokens activate the corresponding features.

Moreover, we find that interpretation scores are moderately negatively correlated with entropy, indicating that more specific features are generally more interpretable (see Figures 50–51 in Appendix A). These scores also tend to have higher precision than

**Feature 67343**

Language: English
Model: Llama-3.2-1B
Layer: layers.15.mlp

Activation Range: 0.3-2.0 | 2.0-3.6 | 3.6-5.3 | 5.3-6.9 | 6.9-8.5 | 8.5-10.2 | 10.2-11.9 | 11.9-13.5 | 13.5-15.1 | 15.1-16.7

**Interpretation**

"The text contains a wide variety of conversational and narrative fragments, including idiomatic expressions, filler words, discourse markers, and references to people, places, and actions. There is a frequent use of pronouns, conjunctions, and auxiliary verbs, as well as sequences that reflect spoken language patterns such as repetitions, hesitations, and clarifications. Many segments highlight the structure of dialogue, question-answer formats, and descriptive or explanatory statements, often focusing on actions, states, or relationships between entities."

**English Examples**

**xnli-188.** yeah yeah glad to see y 'all taken care of well the i think what changed everything and uh is uh y 'all were y 'all the only ones that make any money for TI here in the last two years In the past few years , no one has made any money for TI .

**xnli-34.** Felicia 's Journey takes place behind the eyes of its central a young Irish girl , Felicia , who crosses the sea to England in a hopeful quest to find the father of her unborn child

**xnli-936.** was this uh done not by TI people was it It wasn 't done by TI people .

**xnli-230.** it 's um it 's it 's amazingly less expensive than it is It 's much more expensive that way .

**Feature 5889**

Language: English
Model: Llama-3.2-1B
Layer: layers.15.mlp

Activation Range: 0.2-1.4 | 1.4-2.7 | 2.7-3.9 | 3.9-5.1 | 5.1-6.4 | 6.4-7.6 | 7.6-8.8 | 8.8-10.0 | 10.0-11.3 | 11.3-12.5

**Interpretation**

"The highlighted tokens frequently correspond to structural elements in formal or factual writing, such as numbers, dates, conjunctions, prepositions, punctuation, and proper nouns, which are often used to organize, enumerate, or specify information in lists, references, or descriptive passages."

**English Examples**

**xnli-720.** Comments on classified or restricted reports should be transmitted in the manner agreed to by GAO and the agency . There are specific transmission methodologies for classified reports .

**xnli-575.** Analysis of this information revealed that no two of these organizations defined the design review process and its various elements in exactly the same manner . Many organizations have a radically different approach to their design review process .

**xnli-36.** there 's a uh a couple called um oh i 'm going to forgot his name now uh Dirkson I can 't remember their name

**xnli-589.** i 've had a chance to i had a chance to terminate and uh L O A and back to L O A I never had any opportunity to end LOA .

**xnli-732.** Advertisement , replied Tuppence promptly . Tuppence replied right away .

Figure 5: English-specific features in layer 15 of Llama 3.2 1B show the SAE's ability to capture English stylistic patterns learned by the model, with feature 67343 identifying conversational and narrative fragments and feature 5889 recognizing structures in formal or factual writing. See Figures 52–53 in Appendix A for extended examples.

recall because the explanations focus on top activating examples but are evaluated on a broader set, including weaker and non-activating ones (see Appendix A.7 for details). As a result, the explanations mainly reflect top examples from one or a few languages but miss weaker ones, which lowers recall.

### 5.1.2 Feature Interpretations

While many linguistic concepts from earlier layers remain in later layers, we focus on the distinctive aspects that emerge in deeper layers. A consistent, overall pattern observed across languages is hierarchical processing of linguistic information, where the model builds understanding progressively. In earlier layers, the model's features concentrate on foundational, low-level elements such as morphemes (e.g., suffixes, affixes, and word stems) as well as basic function words and punctuation. Progressing to middle layers, these features begin to integrate these basic components into larger, more cohesive units, such as multi-word phrases and clauses that convey specific actions, states, or relationships. Finally, in deeper layers, the model demonstrates a more abstract and sophisticated comprehension, with features that capture complex sentence-level structures, discourse patterns, pragmatic elements like conversational flow, and nuanced, context-specific meanings. See Appendix A.10 for language-specific interpretations.

Notably, we find highly specialized features that capture stylistic patterns in English. As shown in Figure 5, English features in layer 15 can distinguish between text styles, with feature 67343 identifying conversational and narrative fragments like idiomatic expressions, and feature 5889 recognizing structural elements of formal or factual writing, such as lists and references.

Moreover, features that capture language-specific context are also present. Multiple features in Korean (feature 125019 in layer 10 and feature 67845 in layer 12) are dedicated to recognizing historical and cultural concepts, specifically those related to King Sejong and the invention of the Hangul alphabet, as illustrated in Figure 54–55 in Appendix A. Similarly, features in other languages learn to identify key geopolitical and geographic entities, such as one in Thai for the word "Thailand" (layer 5, feature 118169), one in Portuguese for "Brazil" (layer 7, feature 118682), and one in Hindi for "Pakistan" (layer 10, feature 84378). These examples show

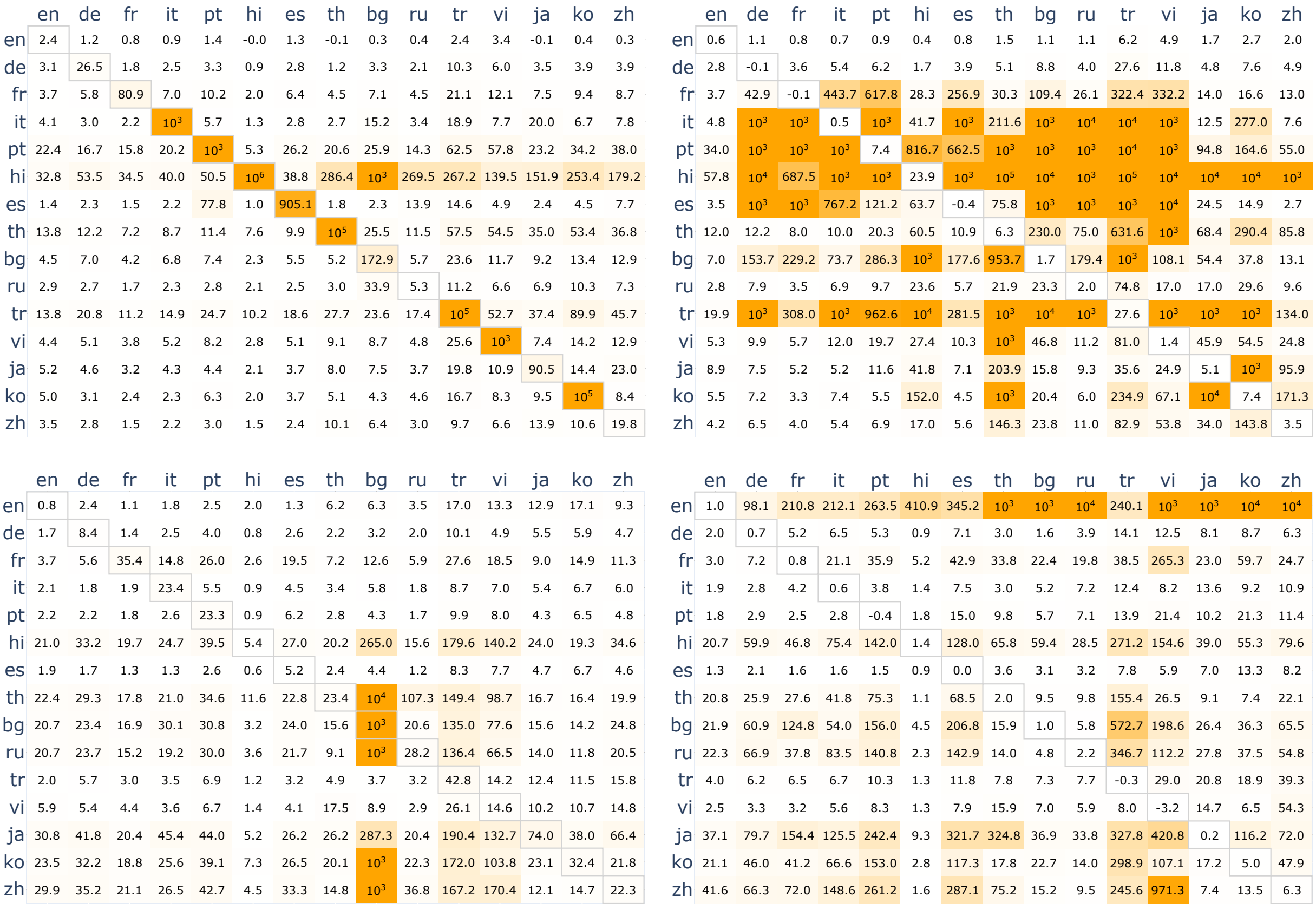

Figure 6: PPL changes in Llama 3.2 1B after suppressing language-specific features and neurons with $\alpha < 0$ and activating them with $\alpha > 0$. **Top**: Steering using **language-specific features** with $\alpha = -0.2$ (left) and $\alpha = 0.2$ (right). **Bottom**: Steering using **language-specific neurons** with $\alpha = -0.2$ (left) and $\alpha = 0.2$ (right). Each cell ($i$, $j$) shows the PPL change for language $j$ after intervening on language $i$. The heatmap color scale is capped at 1000. Notably, the heatmap shows that suppressing language-specific features mainly affects the PPL of the target language, with limited effects on other languages, and produces more isolated effects than language-specific neurons, indicating their greater monosemanticity. See Figures 26–28 in Appendix A.15.6 for results with $\alpha \in \{\pm 0.1, \pm 0.3, \pm 0.4\}$.

SAEs' ability to capture specific knowledge relevant to each language. Refer to Figures 56–58 in Appendix A for feature activation examples.

Interestingly, some features learn to recognize abstract patterns in other languages, even across different families. For example, Turkish-specific feature 97688 in layer 10 also identifies verb endings common to Romance languages (see Figure 59 in Appendix A). Another surprising instance is Japanese-specific feature 32154 in layer 8, which is strongly active for Bulgarian morphemes (see Figure 60 in Appendix A). However, since interpretations rely on each feature's top activating examples, they may be biased toward examples from other languages that activate the feature most strongly.

## 5.2 Steering with Language-Specific Features

### 5.2.1 Perplexity Change

We perform steering with language-specific features in Llama 3.2 1B using a scaling factor of $\alpha = \pm 0.2$ for each language. From Figure 6, we observe that suppressing language-specific features primarily affects the PPL of the targeted language while having a relatively minimal impact on the other languages. Conversely, activating language-specific features has a minimal impact on the targeted language, but could significantly disrupt the other languages. Similar effects are observed when steering with language-specific neurons, although the effects are less pronounced and less isolated with a negative scaling factor. Furthermore, as shown in Appendix A.15.6, we try other values and observe higher PPL changes with higher-magnitude factors.[5] The same trend also holds for Gemma 2.

We observe that intervention in English exhibits different results compared to that in other languages. We speculate that this is because English dominates the training data of both the model and the SAEs; as a result, English-specific features may be more granular and not always HFLs (see Appendix A.12 for details). Thus, steering with English-specific features may not significantly affect PPL.

[5] In contrast, activating and suppressing language-shared features across 15 languages have similar effects on PPL, as shown in Table 1 in Appendix A.

| Targeted Language | α | Output |
|---|---|---|
| - | - | Question:Write a Python code snippet to Display Low Nail care: Trimming Nails for Decision Making for Experts. Consider the Trust and implement appropriate if/else or switch/case statements |
| de | 0.4 | if "bathroom" in shared_space:<br>print("In der Badezimmer ist die Hygiene nicht so gut wie in den anderen Räumen.") |
| fr | 0.3 | Question:Write a Python code snippet to Display Low Hygiene during illness: Disinfecting Lesions et de l'hygiène des patients. Ce qui est la solution la plus simple? |
| es | 0.5 | Question:Write a Python code snippet that Determines Extreme Weather: Precipitación by City (data.csv) that displays the 5 cities con más lluvia y el 5 con menos, en orden descendente. |
| ru | 0.5 | Question:Create a Python script snippet that Determines Extreme Fasting: Diaper Rash на Англии на русизации. Сделайте ее как можно более простой и понятной. |
| ja | 0.3 | Question:Write a Python code snippet to DetermineLowFees: Paymentテストをおしまいでお考えください。テストコードをお持ちです。 |
| zh | 0.3 | Question:WriteaPythoncode snippettoDisplayaListofPythonLibraries.使用Python的库和包列表显示列表。请使用Python3.5。显示列表时，使用Python3.5的方法。使用Python3.5的方法。 |

Figure 7: Example outputs from Llama 3.2 1B with `seed` = 0 under steering with language-specific features at $\alpha > 0$, effectively activating language-specific concepts. The model is prompted with the `[BOS]` token and generates outputs using top-p sampling. Full outputs are shown in Figures 65–71 in Appendix A.

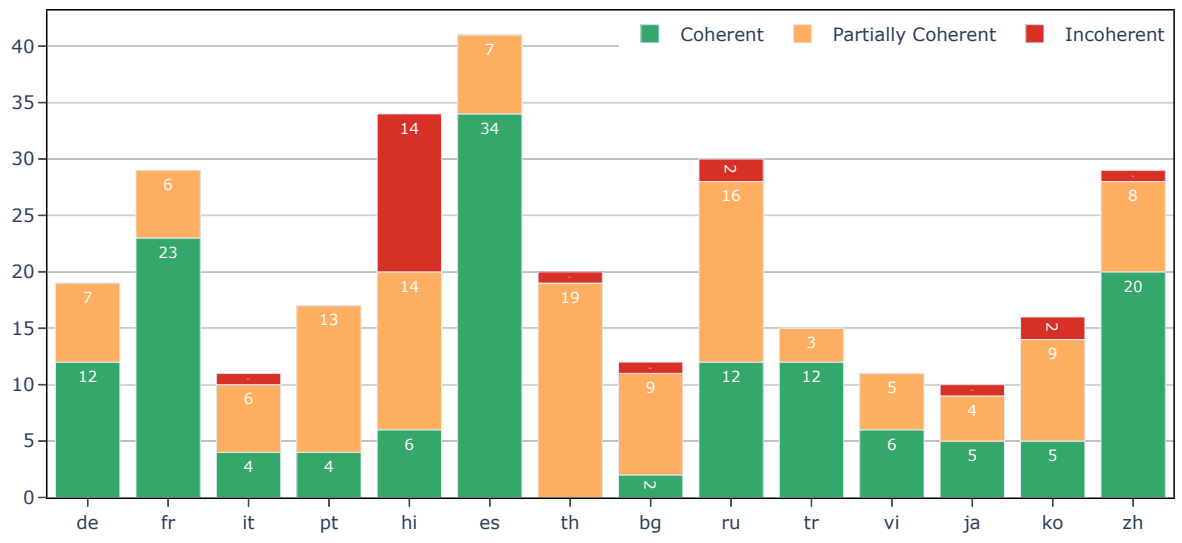

Figure 8: Number of generated texts from Llama 3.2 1B containing the target language across 100 generation attempts for each language under steering with language-specific features at various scaling factors $\alpha$. Notably, steering can shift the output toward the target language, but may also degrade generation quality. See Table 9 in Appendix A for details.

#### 5.2.2 Unconditional Text Generation

Figure 7 shows that steering with language-specific features can guide the model to generate text in the target language. Interestingly, when generating code, the model correctly switches between using the target language for descriptive text (e.g., comments or string literals) and English for syntax, as shown in Figures 72–74 in Appendix A.

However, we acknowledge that the intervention is not always successful and sometimes leads to degraded outputs. To quantitatively evaluate the results, we prompt an LLM with a rubric assessing two criteria: language change occurrence (unchange, change) and text coherence (incoherent, partially coherent, coherent). See Appendix A.13 for rubric details.

Figure 8 demonstrates that manipulating language-specific features enhances the likelihood of generating text in the target language during unconditional text generation. Unfortunately, we observe that further increasing the scaling factor, while increasing this likelihood, can lead to degenerate outputs, specifically a rise in incoherent and partially coherent generations, as shown in Tables 9–10 and Figures 76–78 in Appendix A. This suggests that we may need to develop better steering strategies for language-specific features, such as diminishing or dynamic steering (Scalena et al., 2024) or filtering steering features (Arad et al., 2025). We leave this for future work.

### 5.3 LID with Language-Specific Features

The activation of language-specific features mostly for one language suggests their potential for language classification. For Llama 3.2 1B, Figure 9 shows that the performance of the SAE-based LID is slightly below fastText (Joulin et al., 2017), while the neuron-based LID performs poorly. For Gemma 2, the results are more mixed: SAE-based and neuron-based LID are broadly comparable to fastText overall, but exhibit weaker performance on certain languages (see Appendix A.15.7).

The SAE-based LID, however, offers more interpretability by showing which features activate on specific tokens in a given input, along with the interpretations of those features. To demonstrate this, we analyze the following English sentence from WiLI-2018: "Hamilton has been the primary songwriter, guitarist, and vocalist for Brothers Past, as well as co-producer for all of their recorded releases." See Figure 79 in Appendix A for the full feature activation details. The lower layers (0–2) primarily recognize English function words, such as "has" and "been" (auxiliary verbs), "the" (article), "for" and "of" (prepositions), "and" and "as well as" (conjunction), and even punctuation marks like the comma (,). In middle layers (4–8), the activated features correspond to English phrase structures and common word combinations. Most notably, feature 34138 in layer 4 activates on multi-word segments forming complete semantic units. Higher-level features in layer 15 show activation across almost the entire sentence, demonstrating recognition of overall English structure, including punctuation patterns such as the use of commas in lists.

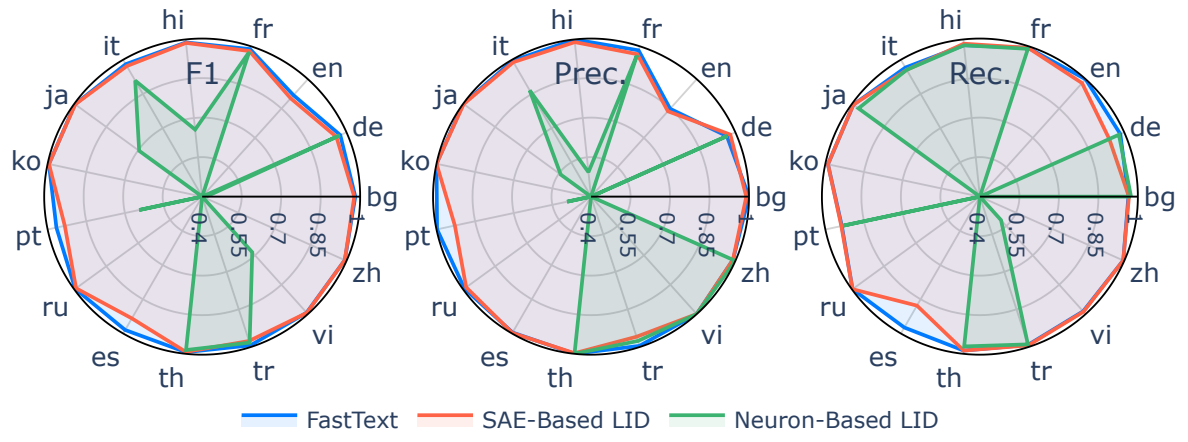

Figure 9: Results of SAE-based and neuron-based LID derived from Llama 3.2 1B and fastText LID, applied to WiLI-2018 across 15 languages. SAE-based LID achieves results comparable to fastText, while neuron-based LID underperforms. Details are provided in Table 11 in Appendix A.

Overall, these features help us understand which aspects of particular languages are detected across layers for a given input and why the classifier makes such predictions, highlighting the potential of using SAEs as concept classifiers. However, the current approach may be biased toward languages with more features. This bias is particularly evident in mixed-language inputs, as shown in the misclassification case in Appendix A.14, and is also reflected in the results for Gemma 2 in Appendix A.15.7. We leave this for future work.

## 6 Related Work

**Language-Specific Neurons.** Multilingual LLM analyses often treat individual neurons as the unit of analysis. Tang et al. (2024) proposed *language activation probability entropy* (LAPE) to detect language-specific neurons in the FFN activations and showed their impact on multilingual capabilities of LLMs. Kojima et al. (2024) showed that identifying and manipulating such neurons in decoder-based LLMs can control the output language. Wang et al. (2024) found that neuron activation patterns to vary across languages, shaped by task and example characteristics.

**Language-Shared and Language-Specific Features.** Few studies have explored the possible existence of such features in SAEs. Brinkmann et al. (2025) showed that abstract grammatical concepts are often captured by features shared among various languages. Jing et al. (2025) used counterfactual datasets to extract, evaluate, and manipulate linguistic features across four dimensions: morphology, syntax, semantics, and pragmatics. Deng et al. (2025) and Chou et al. (2025) proposed methods for identifying language-specific features in the residual stream using mean activation differences, following an approach similar to Soo et al. (2025). Both studies found that, at each layer, the top-$k$ features ranked by mean activation differences between a target language and other languages are highly language-specific. Deng et al. (2025) further showed that ablating these features at selected layers substantially increases the cross-entropy loss for the target language while leaving other languages largely unaffected. In contrast, Chou et al. (2025) demonstrated that the causal effect of steering these features varies across layers and can control the output language. Unlike prior work, we target language-specific features that are also HFLs using *SAE-LAPE* in the FFN, analyze both their properties and interpretations, manipulate these features simultaneously across all layers, and compare the results with a language-specific neuron baseline using *LAPE*.

## 7 Conclusion

This study offers new insights into the internal representations and multilingual mechanisms of LLMs through language-specific features from SAEs. These features mainly appear in the middle to late layers and are interpretable. They also impact the model's multilingual performance and output. Finally, the SAE-based classifier enables interpretable language identification, highlighting its potential for both interpretability and classification, with the latter possibly extending to tasks beyond language identification, such as safety.

## Limitations

**Language-Shared Features.** This study primarily focuses on language-specific features. While our preliminary analysis revealed interesting properties of language-shared features, they warrant further investigation in future work.

**Steering Methods.** We used the most commonly adopted steering method for SAEs in this study. To enhance the efficacy of language-specific features in supporting the multilingual abilities of LLMs, future research could explore new steering strategies, such as diminishing steering, dynamic steering, or filtering steering features. This is important, as steering multiple features simultaneously is not yet common practice and may be challenging.

**SAE-Based Classifier.** The current preliminary classifier is limited to language identification and may be biased toward languages with more features, which should be addressed to develop a more reliable classifier. Enhancing the use of SAEs as interpretable classifiers remains a promising direction for detecting specific concepts in LLMs, including tasks beyond language identification, such as safety monitoring.

## References

Dana Arad, Aaron Mueller, and Yonatan Belinkov. 2025. SAEs are good for steering – if you select the right features. In *Proceedings of the 2025 Conference on Empirical Methods in Natural Language Processing*, pages 10241–10259, Suzhou, China.

Steven Bills, Nick Cammarata, Dan Mossing, Henk Tillman, Leo Gao, Gabriel Goh, Ilya Sutskever, Jan Leike, Jeff Wu, and William Saunders. 2023. Language models can explain neurons in language models. https://openaipublic.blob.core.windows.net/neuron-explainer/paper/index.html.

Joseph Bloom and Johnny Lin. 2024. Understanding sae features with the logit lens. LessWrong blog post. Informal post from the ML Alignment & Theory Scholars Program.

Trenton Bricken, Adly Templeton, Joshua Batson, Brian Chen, Adam Jermyn, Tom Conerly, Nick Turner, Cem Anil, Carson Denison, Amanda Askell, Robert Lasenby, Yifan Wu, Shauna Kravec, Nicholas Schiefer, Tim Maxwell, Nicholas Joseph, Zac Hatfield-Dodds, Alex Tamkin, Karina Nguyen, and 6 others. 2023. Towards monosemanticity: Decomposing language models with dictionary learning. *Transformer Circuits Thread*.

Jannik Brinkmann, Chris Wendler, Christian Bartelt, and Aaron Mueller. 2025. Large language models share representations of latent grammatical concepts across typologically diverse languages. In *Proceedings of the 2025 Conference of the Nations of the Americas Chapter of the Association for Computational Linguistics: Human Language Technologies (Volume 1: Long Papers)*, pages 6131–6150, Albuquerque, New Mexico.

B. Bryson. 2015. *The Mother Tongue: English and How it Got that Way*. HarperCollins.

Cheng-Ting Chou, George Liu, Jessica Sun, Cole Blondin, Kevin Zhu, Vasu Sharma, and Sean O'Brien. 2025. Causal language control in multilingual transformers via sparse feature steering. In *ACL 2025 Student Research Workshop*.

Alexis Conneau, Ruty Rinott, Guillaume Lample, Adina Williams, Samuel Bowman, Holger Schwenk, and Veselin Stoyanov. 2018. XNLI: Evaluating cross-lingual sentence representations. In *Proceedings of the 2018 Conference on Empirical Methods in Natural Language Processing*, pages 2475–2485, Brussels, Belgium.

Hoagy Cunningham and Tom Conerly. 2024. Circuits updates – june 2024. Circuits Updates series, Anthropic.

Boyi Deng, Yu Wan, Baosong Yang, Yidan Zhang, and Fuli Feng. 2025. Unveiling language-specific features in large language models via sparse autoencoders. In *Proceedings of the 63rd Annual Meeting of the Association for Computational Linguistics (Volume 1: Long Papers)*, pages 4563–4608, Vienna, Austria.

Esin Durmus, Alex Tamkin, Jack Clark, Jerry Wei, Jonathan Marcus, Joshua Batson, Kunal Handa, Liane Lovitt, Meg Tong, Miles McCain, Oliver Rausch, Saffron Huang, Sam Bowman, Stuart Ritchie, Tom Henighan, and Deep Ganguli. 2024. Evaluating feature steering: A case study in mitigating social biases. https://anthropic.com/research/evaluating-feature-steering. Accessed: 2025-05-01.

EleutherAI. 2024. sae-llama-3.2-1b-131k. https://huggingface.co/EleutherAI/sae-Llama-3.2-1B-131k.

Nelson Elhage, Tristan Hume, Catherine Olsson, Nicholas Schiefer, Tom Henighan, Shauna Kravec, Zac Hatfield-Dodds, Robert Lasenby, Dawn Drain, Carol Chen, Roger Grosse, Sam McCandlish, Jared Kaplan, Dario Amodei, Martin Wattenberg, and Christopher Olah. 2022. Toy models of superposition. *Transformer Circuits Thread*.

N. Benjamin Erichson, Zhewei Yao, and Michael W. Mahoney. 2020. Jumprelu: A retrofit defense strategy for adversarial attacks. In *Proceedings of the 9th International Conference on Pattern Recognition Applications and Methods - ICPRAM*, pages 103–114. INSTICC, SciTePress.

Javier Ferrando, Oscar Balcells Obeso, Senthooran Rajamanoharan, and Neel Nanda. 2025. Do i know this entity? knowledge awareness and hallucinations in language models. In *The Thirteenth International Conference on Learning Representations*.

Leo Gao, Tom Dupre la Tour, Henk Tillman, Gabriel Goh, Rajan Troll, Alec Radford, Ilya Sutskever, Jan Leike, and Jeffrey Wu. 2025. Scaling and evaluating sparse autoencoders. In *The Thirteenth International Conference on Learning Representations*.

Leo Gao, Jonathan Tow, Baber Abbasi, Stella Biderman, Sid Black, Anthony DiPofi, Charles Foster, Laurence Golding, Jeffrey Hsu, Alain Le Noac'h, Haonan Li, Kyle McDonell, Niklas Muennighoff, Chris Ociepa, Jason Phang, Laria Reynolds, Hailey Schoelkopf, Aviya Skowron, Lintang Sutawika, and 5 others. 2024. The language model evaluation harness.

Xavier Glorot, Antoine Bordes, and Yoshua Bengio. 2011. Deep sparse rectifier neural networks. In *Proceedings of the Fourteenth International Conference on Artificial Intelligence and Statistics*, volume 15 of *Proceedings of Machine Learning Research*, pages 315–323, Fort Lauderdale, FL, USA. PMLR.

Robert Huben, Hoagy Cunningham, Logan Riggs Smith, Aidan Ewart, and Lee Sharkey. 2024. Sparse autoencoders find highly interpretable features in language models. In *The Twelfth International Conference on Learning Representations*.

Yi Jing, Zijun Yao, Hongzhu Guo, Lingxu Ran, Xiaozhi Wang, Lei Hou, and Juanzi Li. 2025. LinguaLens: Towards interpreting linguistic mechanisms of large language models via sparse auto-encoder. In *Proceedings of the 2025 Conference on Empirical Methods in Natural Language Processing*, pages 28232–28251, Suzhou, China.

Armand Joulin, Edouard Grave, Piotr Bojanowski, and Tomas Mikolov. 2017. Bag of tricks for efficient text classification. In *Proceedings of the 15th Conference of the European Chapter of the Association for Computational Linguistics: Volume 2, Short Papers*, pages 427–431, Valencia, Spain. Association for Computational Linguistics.

Takeshi Kojima, Itsuki Okimura, Yusuke Iwasawa, Hitomi Yanaka, and Yutaka Matsuo. 2024. On the multilingual ability of decoder-based pre-trained language models: Finding and controlling language-specific neurons. In *Proceedings of the 2024 Conference of the North American Chapter of the Association for Computational Linguistics: Human Language Technologies (Volume 1: Long Papers)*, pages 6919–6971, Mexico City, Mexico.

Kristian Kuznetsov, Laida Kushnareva, Anton Razzhigaev, Polina Druzhinina, Anastasia Voznyuk, Irina Piontkovskaya, Evgeny Burnaev, and Serguei Barannikov. 2025. Feature-level insights into artificial text detection with sparse autoencoders. In *Findings of the Association for Computational Linguistics: ACL 2025*, pages 25727–25748, Vienna, Austria.

Yongqi Leng and Deyi Xiong. 2025. Towards understanding multi-task learning (generalization) of LLMs via detecting and exploring task-specific neurons. In *Proceedings of the 31st International Conference on Computational Linguistics*, pages 2969–2987, Abu Dhabi, UAE. Association for Computational Linguistics.

Tom Lieberum, Senthooran Rajamanoharan, Arthur Conmy, Lewis Smith, Nicolas Sonnerat, Vikrant Varma, Janos Kramar, Anca Dragan, Rohin Shah, and Neel Nanda. 2024. Gemma scope: Open sparse autoencoders everywhere all at once on gemma 2. In *Proceedings of the 7th BlackboxNLP Workshop: Analyzing and Interpreting Neural Networks for NLP*, pages 278–300, Miami, Florida, US.

Meta AI. 2024a. Llama 3.2 1b. `https://huggingface.co/meta-llama/Llama-3.2-1B`. Accessed: 2025-07-29.

Meta AI. 2024b. Llama 3.2 brings connect 2024 vision to edge and mobile devices. `https://ai.meta.com/blog/llama-3-2-connect-2024-vision-edge-mobile-devices/`. Accessed: 2025-05-08.

Soumen Kumar Mondal, Sayambhu Sen, Abhishek Singhania, and Preethi Jyothi. 2025. Language-specific neurons do not facilitate cross-lingual transfer. In *The Sixth Workshop on Insights from Negative Results in NLP*, pages 46–62, Albuquerque, New Mexico.

Niklas Muennighoff, Thomas Wang, Lintang Sutawika, Adam Roberts, Stella Biderman, Teven Le Scao, M Saiful Bari, Sheng Shen, Zheng Xin Yong, Hailey Schoelkopf, Xiangru Tang, Dragomir Radev, Alham Fikri Aji, Khalid Almubarak, Samuel Albanie, Zaid Alyafeai, Albert Webson, Edward Raff, and Colin Raffel. 2023. Crosslingual generalization through multitask finetuning. In *Proceedings of the 61st Annual Meeting of the Association for Computational Linguistics (Volume 1: Long Papers)*, pages 15991–16111, Toronto, Canada.

NLLB Team, Marta R. Costa-jussà, James Cross, Onur Çelebi, Maha Elbayad, Kenneth Heafield, Kevin Heffernan, Elahe Kalbassi, Janice Lam, Daniel Licht, Jean Maillard, Anna Sun, Skyler Wang, Guillaume Wenzek, Al Youngblood, Bapi Akula, Loic Barrault, Gabriel Mejia Gonzalez, Prangthip Hansanti, and 20 others. 2024. Scaling neural machine translation to 200 languages. *Nature*, 630(8018):841–846.

nostalgebraist. 2020. Interpreting gpt: The logit lens. LessWrong post.

OpenAI. 2025. Introducing gpt-4.1 in the api. `https://openai.com/index/gpt-4-1/`. Accessed: 2025-05-06.

Gonçalo Santos Paulo, Alex Troy Mallen, Caden Juang, and Nora Belrose. 2025. Automatically interpreting millions of features in large language models. In *Forty-second International Conference on Machine Learning*.

Senthooran Rajamanoharan, Tom Lieberum, Nicolas Sonnerat, Arthur Conmy, Vikrant Varma, János Kramár, and Neel Nanda. 2024. Jumping ahead: Improving reconstruction fidelity with jumprelu sparse autoencoders. *Preprint*, arXiv:2407.14435.

Daniel Scalena, Gabriele Sarti, and Malvina Nissim. 2024. Multi-property steering of large language models with dynamic activation composition. In *Proceedings of the 7th BlackboxNLP Workshop: Analyzing and Interpreting Neural Networks for NLP*, pages 577–603, Miami, Florida, US.

Noam Shazeer. 2020. Glu variants improve transformer. *Preprint*, arXiv:2002.05202.

Samuel Soo, Wesley Teng, Chandrasekaran Balaganesh, Tan Guoxian, and Ming YAN. 2025. Interpretable steering of large language models with feature guided activation additions. In *ICLR 2025 Workshop on Building Trust in Language Models and Applications*.

Xiaoqing Sun, Joshua Engels, and Max Tegmark. 2025. High frequency latents are features, not bugs. In *Sparsity in LLMs (SLLM): Deep Dive into Mixture of Experts, Quantization, Hardware, and Inference*.

Tianyi Tang, Wenyang Luo, Haoyang Huang, Dongdong Zhang, Xiaolei Wang, Xin Zhao, Furu Wei, and Ji-Rong Wen. 2024. Language-specific neurons: The key to multilingual capabilities in large language models. In *Proceedings of the 62nd Annual Meeting of the Association for Computational Linguistics (Volume 1: Long Papers)*, pages 5701–5715, Bangkok, Thailand.

Gemma Team, Morgane Riviere, Shreya Pathak, Pier Giuseppe Sessa, Cassidy Hardin, Surya Bhupatiraju, Léonard Hussenot, Thomas Mesnard, Bobak Shahriari, Alexandre Ramé, Johan Ferret, Peter Liu, Pouya Tafti, Abe Friesen, Michelle Casbon, Sabela Ramos, Ravin Kumar, Charline Le Lan, Sammy Jerome, and 179 others. 2024. Gemma 2: Improving open language models at a practical size. *Preprint*, arXiv:2408.00118.

Adly Templeton, Tom Conerly, Jonathan Marcus, Jack Lindsey, Trenton Bricken, Brian Chen, Adam Pearce, Craig Citro, Emmanuel Ameisen, Andy Jones, Hoagy Cunningham, Nicholas L Turner, Callum McDougall, Monte MacDiarmid, C. Daniel Freeman, Theodore R. Sumers, Edward Rees, Joshua Batson, Adam Jermyn, and 3 others. 2024. Scaling monosemanticity: Extracting interpretable features from claude 3 sonnet. *Transformer Circuits Thread*.

Martin Thoma. 2018. The wili benchmark dataset for written language identification. *Preprint*, arXiv:1801.07779.

Alexander Matt Turner, Lisa Thiergart, Gavin Leech, David Udell, Juan J. Vazquez, Ulisse Mini, and Monte MacDiarmid. 2024. Steering language models with activation engineering. *Preprint*, arXiv:2308.10248.

Ashish Vaswani, Noam Shazeer, Niki Parmar, Jakob Uszkoreit, Llion Jones, Aidan N Gomez, Ł ukasz Kaiser, and Illia Polosukhin. 2017. Attention is all you need. In *Advances in Neural Information Processing Systems*, volume 30. Curran Associates, Inc.

Weixuan Wang, Barry Haddow, Minghao Wu, Wei Peng, and Alexandra Birch. 2024. Sharing matters: Analysing neurons across languages and tasks in llms. *Preprint*, arXiv:2406.09265.

Maurice Weber, Daniel Y. Fu, Quentin Anthony, Yonatan Oren, Shane Adams, Anton Alexandrov, Xiaozhong Lyu, Huu Nguyen, Xiaozhe Yao, Virginia Adams, Ben Athiwaratkun, Rahul Chalamala, Kezhen Chen, Max Ryabinin, Tri Dao, Percy Liang, Christopher Ré, Irina Rish, and Ce Zhang. 2024. Redpajama: an open dataset for training large language models. In *Advances in Neural Information Processing Systems*, volume 37, pages 116462–116492. Curran Associates, Inc.

Yinfei Yang, Yuan Zhang, Chris Tar, and Jason Baldridge. 2019. PAWS-X: A cross-lingual adversarial dataset for paraphrase identification. In *Proceedings of the 2019 Conference on Empirical Methods in Natural Language Processing and the 9th International Joint Conference on Natural Language Processing (EMNLP-IJCNLP)*, pages 3687–3692, Hong Kong, China.

# A Appendix

## A.1 Feed-Forward Network

The feed-forward network (FFN) module of LLMs with a Transformer architecture (Vaswani et al., 2017) can be described as follows:

$$\boldsymbol{h} = \sigma(\boldsymbol{W}_1\boldsymbol{x} + \boldsymbol{b}_1)\,,$$

$$\text{FFN}(\boldsymbol{x}) = \boldsymbol{W}_2\boldsymbol{h} + \boldsymbol{b}_2\,,$$

where $\boldsymbol{h}$ denotes the intermediate activations; $\boldsymbol{x} \in \mathbb{R}^d$ is the pre-FFN representation from the preceding attention sub-layer; $\sigma$ is an activation function, such as $\text{ReLU}(x) = \max(0, x)$ by Glorot et al. (2011); $\boldsymbol{W}_1 \in \mathbb{R}^{4d\times d}$ and $\boldsymbol{W}_2 \in \mathbb{R}^{d\times 4d}$ are weight matrices; and $\boldsymbol{b}_1 \in \mathbb{R}^{4d}$ and $\boldsymbol{b}_2 \in \mathbb{R}^{d}$ are bias terms. More recently, Llama 3.2 uses Gated Linear Unit (GLU) by Shazeer (2020) as its activation function to improve performance, and the FFN module operation is defined as follows:

$$\text{FFN}(\boldsymbol{x}) = \boldsymbol{W}_2\left(\boldsymbol{h} \otimes (\boldsymbol{V}\boldsymbol{x} + \boldsymbol{b}_v)\right) + \boldsymbol{b}_2\,,$$

where $\otimes$ denotes the element-wise product; $\boldsymbol{V} \in \mathbb{R}^{4d\times d}$ is a weight matrix; and $\boldsymbol{b}_v \in \mathbb{R}^{4d}$ is a bias term. In both cases, a neuron corresponds to a linear transformation applied to a single row of $\boldsymbol{W}_1$, followed by a non-linear activation that yields a component $\boldsymbol{h}_i$ of the intermediate representation. Hence, an FFN module at one layer has $4d$ neurons. Tang et al. (2024) consider a neuron to be activated when its value exceeds zero and to be deactivated when its value is set to zero.

## A.2 Finding Language-Specific Neurons

To identify language-specific neurons in the intermediate activations of the FFN (see Appendix A.1), we adopt the *language activation probability entropy (LAPE)* method proposed by Tang et al. (2024). Specifically, we provide an LLM with multilingual corpora and compute the activation probability of neuron $j$ at layer $i$ of an LLM for each language $k$ as follows:

$$p^k_{i,j} = \mathbb{E}\left(\mathbb{I}(\boldsymbol{h}^i_j > 0) \mid \text{language } k\text{'s corpus}\right)\,,$$

where $\mathbb{E}$ is the expectation computed over the tokens in language $k$'s corpus; $\mathbb{I}$ is the indicator function; and $\boldsymbol{h}^i_j$ is the value associated with neuron $j$ at layer $i$. The subsequent operations follow those described in Section 3.1. Moreover, Tang et al. (2024) imposed a predefined minimum threshold at the 95th percentile[6] of all activation probabilities on $p^k_{i,j}$ to exclude negligible activation probabilities and used the remaining values to determine which languages the neurons are specific to. Specifically, they selected only the bottom 1% of neurons, sorted by LAPE scores, as language-specific neurons. This means that even if neurons are shared across multiple languages, they are still considered specific to those languages as long as their LAPE scores are low.

## A.3 Steering with Language-Specific Neurons

Inspired by Turner et al. (2024), and to align with the approach used for steering with language-specific features, we scale the activation contribution of language-specific neurons as follows:

$$\boldsymbol{x}_{\text{new}} = \boldsymbol{x} + \alpha \boldsymbol{h}^{\max}_j \boldsymbol{d}^j\,,$$

where $\alpha \in \mathbb{R}$ is a scaling factor, $\boldsymbol{h}^{\max}_j$ denotes the maximum activation value of neuron $j$ observed in the multilingual corpora, and $\boldsymbol{d}^j = \boldsymbol{W}_2[:, j] \in \mathbb{R}^d$ is the steering vector associated with neuron $j$.

## A.4 LID with Language-Specific Neurons

To classify languages using neurons for a text input with FFN output $\boldsymbol{X} = (\boldsymbol{x}_1, \ldots, \boldsymbol{x}_t)$ consisting of $t$ tokens, we compute the score for each language $k$ as follows:

$$s_k = \sum_{i\in\mathcal{L}} \sum_{\boldsymbol{x}\in\boldsymbol{X}} \sum_{j\in\mathcal{H}^i_k} \mathbb{I}(\boldsymbol{h}^i_{\boldsymbol{x},j} > 0)\,,$$

where $\mathcal{L}$ is the set of all layers; $\mathcal{H}^i_k$ is the set of neuron indices associated with language $k$ at layer $i$; $\boldsymbol{h}^i_{\boldsymbol{x},j}$ is the activation value of neuron $j$ at layer $i$ for input $\boldsymbol{x}$; and $\mathbb{I}(\boldsymbol{h}^i_{\boldsymbol{x},j} > 0)$ is the indicator function to ensure that only active neurons are considered. Then, we predict the language by selecting the index with the highest score in the vector $\boldsymbol{s} = (s_1, \ldots, s_l)$ over $l$ languages.

## A.5 TopK SAEs Configuration

We use TopK SAEs by EleutherAI (2024) with $k = 32$ and an expansion factor of 64, such that $\boldsymbol{z} \in \mathbb{R}^{64d}$. These SAEs are trained on the RedPajama-v2 (Weber et al., 2024) multilingual corpus. RedPajama-v2 primarily consists of documents in English, German, French, Spanish, and Italian. Specifically, EleutherAI (2024) use the `Sample-10B` subset of this dataset (see Table 4 for the language distribution) and train their SAEs on the FFN outputs with $d = 2048$ across all layers. Furthermore, the specific configuration of the SAEs and their training process are described below.

[6] The 95th percentile of all activation probabilities reported by Tang et al. (2024) for Llama 2 70B is 0.505. However, for SAEs, we find it to be 0.0007 due to their sparsity. Thus, this threshold is not suitable for SAEs, as it excludes almost nothing. To overcome this, we apply two filters to exclude features with low activation probabilities for particular languages.

#### A.5.1 SAE Architecture

The SAEs are configured with the following core parameters:

- **Expansion Factor**: The expansion factor is set to 64, determining the number of SAE features relative to the input dimension. Given the FFN output dimension of $d = 2048$, each SAE learns $64 \times 2048 = 131,072$ features.
- **Activation Function**: A TopK activation function is employed with $k = 32$, meaning that only the 32 highest-activated features are retained as non-zero for any given input.
- **Decoder Normalization**: Normalization is applied to the decoder weights, enforcing a unit norm.

#### A.5.2 Training Configuration

The training procedure is governed by the following settings:

- **Hookpoints**: SAEs are trained on the outputs of the FFN blocks from `layers.0.mlp` to `layers.15.mlp`, resulting in one SAE per layer (16 total).
- **Batching**: A batch size of 4 sequences is used, and gradients are accumulated across 2 steps, yielding an effective batch size of 8.
- **Learning Rate**: The learning rate is automatically determined based on the number of SAE features. A warm-up period of 1000 steps is applied.
- **Dead Feature Threshold**: Features that remain inactive (i.e., not firing) for 10,000,000 consecutive tokens are considered "dead".
- **Context Length**: The SAEs are trained on sequences with a context length of 2048 tokens.
- **Initialization**: A fixed random seed of 0 is used to ensure reproducibility, initializing one SAE per hookpoint.

### A.6 SAE Feature Interpretability

SAE features can be interpreted either manually by humans or automatically by LLMs (Huben et al., 2024; Bricken et al., 2023). Manual interpretation involves analyzing a feature's activating contexts (i.e., sentences) and generating an explanation, which is then evaluated using a predefined rubric. In contrast, automatic interpretation involves presenting these contexts along with predefined instructions to LLMs, which generate an explanation that is also scored by LLMs. This automatic method was inspired by Bills et al. (2023), who used LLMs to interpret neurons, and was further improved by Paulo et al. (2025) to interpret SAE features.

### A.7 Automated Interpretability Process

We use an automated interpretability pipeline by Paulo et al. (2025) to efficiently explain and evaluate language-specific features in Llama 3.2 1B's SAEs. The process involves three key steps:

- **Activation Collection**: We gather activations from diverse multilingual corpora (XNLI, PAWS-X, and FLORES+) spanning 15 languages, caching contexts of 256 tokens for efficient processing.
- **Interpretation Generation**: For each language-specific feature, we prompt GPT-4.1 (OpenAI, 2025) with 40 top-activating examples, each centered on the activating token with 32 tokens of surrounding context.
- **Evaluation Methods**: We assess interpretation quality through detection scoring and fuzzing scoring, each using 100 examples (50 stratified by activation deciles and 50 non-activating). In detection scoring, the model determines whether any token in the full sentence activates the feature based on its explanation. In fuzzing scoring, the model performs the same evaluation but only over randomly selected highlighted words within the sentence.

This automated pipeline enabled us to interpret 540 language-specific features for approximately \$50: \$5 for generating explanations and \$45 for scoring them.

### A.8 Language-Shared Features

Using *SAE-LAPE*, we identified 503 language-shared features from a total of 2 million across layers. As discussed in Section 5.1, features shared across 2–8 languages (roughly half of those analyzed) display a pattern similar to that of language-specific features, as illustrated in Figure 3. In contrast, features shared across 9–15 languages show a different trend, being concentrated in the early to middle layers (see Figure 4).

We also observe the intersection of language-shared features across languages. Figure 10 shows the feature intersection counts, while Figure 11 presents the Jaccard similarity (i.e., intersection over union) of the corresponding features. Remarkably, although the number of language-shared features varies across languages, the Jaccard similarity can indicate the relatedness between them to a certain extent, since languages with a similarity score close to one tend to be more closely related, as they share a greater number of concepts associated with languages.

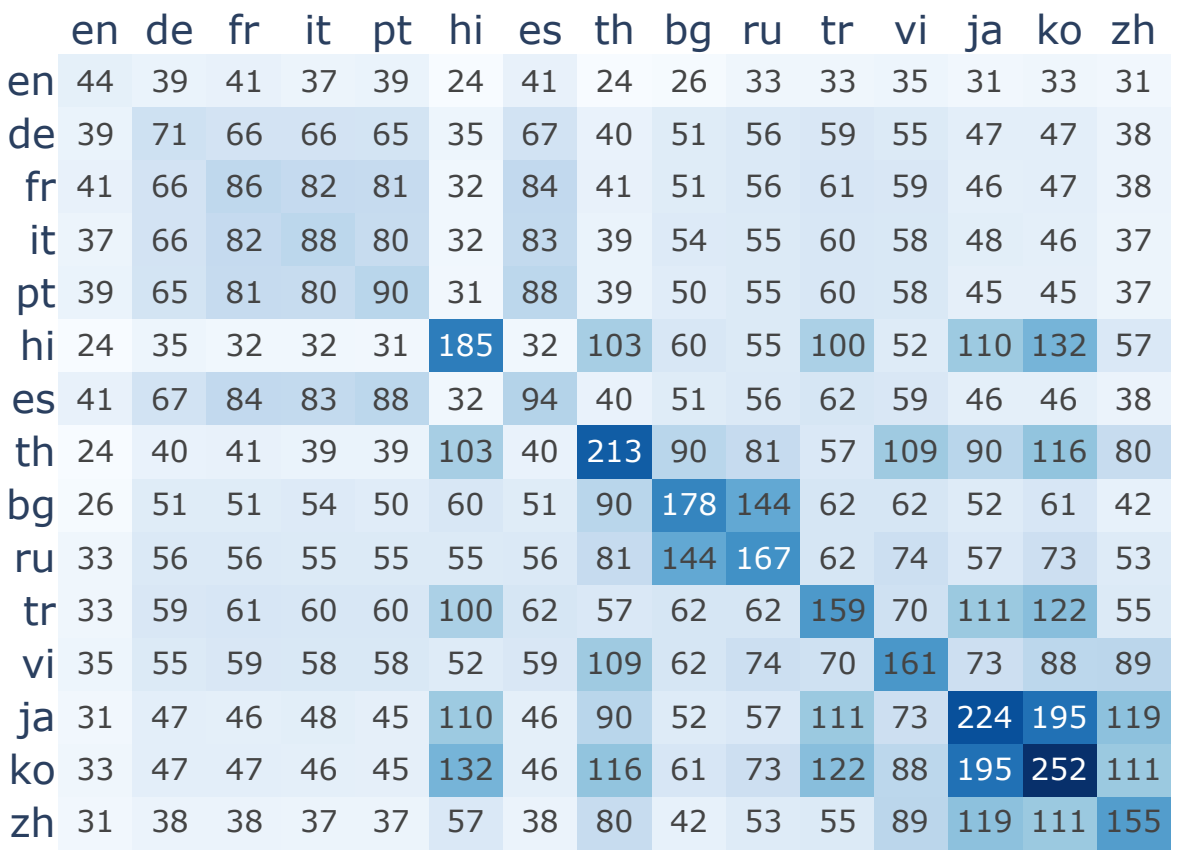

Figure 10: Feature intersection counts of the identified language-shared features in Llama 3.2 1B across layers for each language.

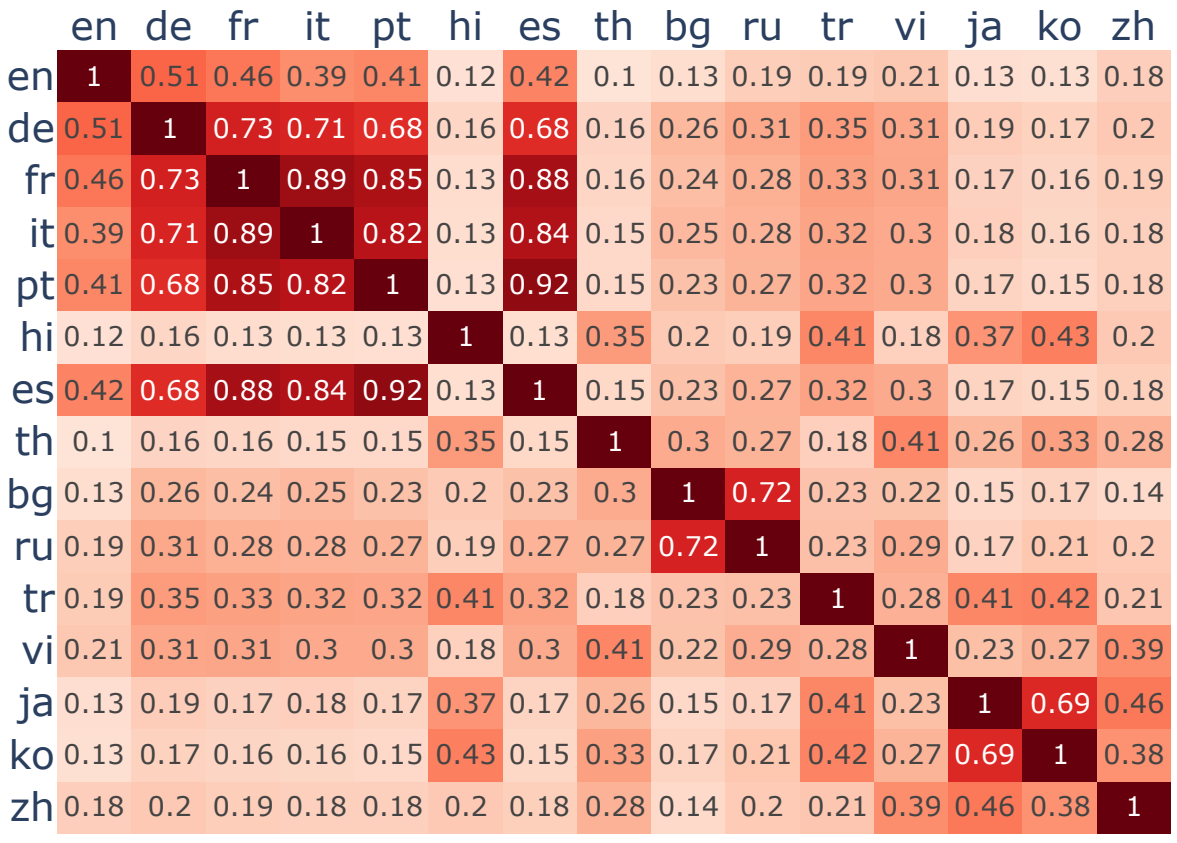

Figure 11: Jaccard similarity of identified language-shared features in Llama 3.2 1B across layers for each language.

Notably, English is most closely related to German, which aligns with their common origin in the Germanic language family. Interestingly, English also exhibits moderate similarity scores with Romance languages such as French, Italian, Portuguese, and Spanish, likely due to significant Latin influence on English vocabulary (Bryson, 2015). Within the Romance group, we observe particularly high similarity scores with one another, indicating a large number of shared concepts and a strong degree of mutual lexical and grammatical overlap. A similar trend is found in the Slavic language group, where Russian and Bulgarian show a high similarity score. Finally, East Asian languages such as Japanese, Korean, and Chinese exhibit moderate similarity scores with one another, which may reflect historical and cultural influences, such as the borrowing of Chinese characters and vocabulary, contributing to a degree of conceptual overlap among these languages (Bryson, 2015).

Lastly, we conduct a PPL change experiment similar to the one described in Section 5.2.1, this time using language-shared features. Specifically, we perform steering using features shared among all 15 languages analyzed in this study, with $\alpha = \pm 0.2, \pm 0.3, \pm 0.4$. As shown in Table 1, these language-shared features consistently affect PPL evenly across languages when suppressed, leading to a greater increase in PPL. Interestingly, they also exhibit a similar effect when activated, in contrast to the behavior of language-specific features, which typically have minimal impact on their corresponding language, as shown in Figure 6 in Section 5.

| Language | Normal | Suppressed | | | Activated | | |
|---|---|---|---|---|---|---|---|
| | | $\alpha$=-0.2 | $\alpha$=-0.3 | $\alpha$=-0.4 | $\alpha$=0.2 | $\alpha$=0.3 | $\alpha$=0.4 |
| en | 40.09 | 40.98 | 42.15 | 44.06 | 40.97 | 42.35 | 44.98 |
| de | 23.99 | 24.87 | 26.00 | 27.86 | 24.85 | 26.26 | 29.14 |
| fr | 18.37 | 18.71 | 19.26 | 20.21 | 19.04 | 19.93 | 21.55 |
| it | 20.55 | 21.18 | 22.01 | 23.43 | 21.21 | 22.24 | 24.24 |
| pt | 28.32 | 29.08 | 30.15 | 32.00 | 29.23 | 30.64 | 33.34 |
| hi | 9.92 | 10.19 | 10.57 | 11.23 | 10.31 | 10.94 | 12.31 |
| es | 25.24 | 25.74 | 26.53 | 27.91 | 26.16 | 27.50 | 30.09 |
| th | 16.14 | 16.78 | 17.71 | 19.32 | 16.99 | 18.42 | 21.56 |
| bg | 21.35 | 22.84 | 24.41 | 26.92 | 21.75 | 23.03 | 25.69 |
| ru | 20.80 | 21.79 | 22.86 | 24.60 | 21.24 | 22.25 | 24.42 |
| tr | 61.50 | 65.41 | 69.91 | 77.42 | 64.02 | 69.25 | 80.50 |
| vi | 44.93 | 46.14 | 48.00 | 51.74 | 46.72 | 49.40 | 54.76 |
| ja | 32.67 | 34.62 | 36.64 | 39.87 | 33.53 | 35.86 | 41.36 |
| ko | 40.27 | 42.60 | 45.05 | 48.93 | 41.88 | 45.51 | 54.11 |
| zh | 49.43 | 53.29 | 56.72 | 61.77 | 49.21 | 51.20 | 56.18 |

Table 1: PPL changes in Llama 3.2 1B across languages with features shared among all 15 languages in this study under various scaling factors $\alpha$.

## A.9 Top Tokens Promoted by Language-Specific Features

To identify which tokens are emphasized by language-specific features, we adopt a method inspired by the logit lens technique proposed by nostalgebraist (2020). Originally, the logit lens was introduced as a simple, although partial, tool for interpreting the intermediate residual stream. It works by multiplying the intermediate activations $\boldsymbol{h}^i \in \mathbb{R}^d$ at layer $i$ by the unembedding matrix $\boldsymbol{W}_U \in \mathbb{R}^{|\mathcal{V}|\times d}$, where $|\mathcal{V}|$ is the vocabulary size. The operation is defined as follows:

$$\boldsymbol{l}^i = \text{softmax}(\boldsymbol{W}_U \boldsymbol{h}^i)\,.$$

When this operation is applied, the resulting token probability distributions $\boldsymbol{l}^i$ tend to be intuitively meaningful. Notably, these distributions often resemble the model's final output distribution and begin to align with it progressively over the course of the network's layers, frequently approaching it well before the final layer. This behavior helps track how the model gradually refines its predictions throughout the forward pass.

However, this method requires a text input so that the distribution resulting from the intermediate residual stream is meaningful. Applying this method directly to language-specific features would require identifying the activating inputs for each of these features, which is time-consuming and makes it difficult to isolate the effect of language-specific features themselves. Therefore, as described by Arad et al. (2025); Bloom and Lin (2024), to simplify our analysis, we directly multiply the features with the token unembedding matrix as follows:

$$\boldsymbol{l}^i = \text{softmax}(\boldsymbol{W}_U \boldsymbol{d}^{i,s})\,,$$

where $\boldsymbol{d}^{i,s}$ is the direction of a language-specific feature at layer $i$. Essentially, this is equivalent to applying the logit lens to a single token, but without the need to determine the corresponding token.

This method is possible due to the way the intermediate residual stream is computed:

$$\boldsymbol{h}^i = \boldsymbol{x}^i + \text{FFN}^i(\boldsymbol{x}^i)\,,$$

$$\boldsymbol{h}^i = \boldsymbol{x}^i + \hat{\text{FFN}}^i(\boldsymbol{x}^i) + \boldsymbol{e}\,,$$

$$\boldsymbol{h}^i = \boldsymbol{x}^i + \sum\nolimits_{j\in\mathcal{T}_K} \boldsymbol{z}^i_j \boldsymbol{d}^{i,j} + \boldsymbol{b}^i_{\text{dec}} + \boldsymbol{e}\,,$$

where $\boldsymbol{h}^i \in \mathbb{R}^d$ is the intermediate activations at layer $i$; $\boldsymbol{x}^i \in \mathbb{R}^d$ is the pre-FFN representation at layer $i$; $\text{FFN}^i(\boldsymbol{x}^i)$ is the FFN output at layer $i$; $\hat{\text{FFN}}^i(\boldsymbol{x}^i)$ is the FFN output reconstruction by an SAE at layer $i$; $\boldsymbol{e} \in \mathbb{R}^d$ is the reconstruction error; $\mathcal{T}_K \subseteq \{0, \ldots, n-1\}$ is the set of indices of the top-$k$ values; $\boldsymbol{z}^i_j$ is the activation of feature $j$ at layer $i$; $\boldsymbol{d}^{i,j} = \boldsymbol{W}^i_{\text{dec}}[:, j] \in \mathbb{R}^d$ is the direction of feature $j$ at layer $i$; and $\boldsymbol{b}^i_{\text{dec}}$ is the bias vector for an SAE at layer $i$. Then, for a language-specific feature $\boldsymbol{d}^{i,s}$, we can decompose the token probability distribution $\boldsymbol{l}^i$ into:

$$\boldsymbol{l}^i = \text{softmax}(\boldsymbol{W}_U \boldsymbol{h}^i)\,,$$

$$\boldsymbol{l}^i = \text{softmax}(\boldsymbol{W}_U(\boldsymbol{x}^i + \boldsymbol{z}^i_s \boldsymbol{d}^{i,s} + \boldsymbol{b}^i_{\text{dec}} + \boldsymbol{e}))\,,$$

$$\boldsymbol{l}^i \propto \exp(\boldsymbol{W}_U \boldsymbol{z}^i_s \boldsymbol{d}^{i,s})\,,$$

$$\boldsymbol{l}^i \propto \text{softmax}(\boldsymbol{W}_U \boldsymbol{d}^{i,s}) \text{ when } \boldsymbol{z}^i_s = 1\,.$$

Here, $\boldsymbol{W}_U \boldsymbol{x}^i$ can be regarded as a constant value for a particular token, $\boldsymbol{b}^i_{\text{dec}}$ is the fixed bias vector, and $\boldsymbol{e}$ is assumed to be small. Thus, based on this proportionality, $\text{softmax}(\boldsymbol{W}_U \boldsymbol{d}^{i,s})$ can be interpreted as the contribution of a language-specific feature $\boldsymbol{d}^{i,s}$ to the token probability distribution $\boldsymbol{l}^i$.

By using this method, we found that features from shallower layers are associated with generic or mixed-language tokens, while deeper-layer features increasingly promote coherent, language-specific vocabulary and complex morphological patterns. This pattern implies that deeper layers of the model capture increasingly language-specific linguistic concepts. See Figures 35–43 for the top 10 tokens promoted by particular language-specific features.

### A.10 Language-Specific Feature Interpretations

This section provides a detailed summary of language-specific feature interpretations across fifteen languages and highlights emerging patterns. While many feature interpretations identified in earlier layers persist in later ones, we focus on aspects that stand out in deeper layers. It is also worth noting that Llama 3.2 1B does not include complete token coverage for all languages in its tokenizer, which particularly affects morphologically rich or non-Latin script languages such as Korean, Japanese, and Chinese.

#### A.10.1 Spanish-Specific Features

A discernible pattern emerges in the Spanish interpretations, suggesting hierarchical processing of linguistic information within the model. Early layers (layers 2, 3, and 5) tend to focus on more granular, morphological aspects of the language, such as morphemes (e.g., suffixes, affixes, and endings), word stems, and basic function words, emphasizing the constituent parts of words and their immediate grammatical roles (e.g., gender, number, and verb conjugation). As we move to the deeper layers (layers 11 and 13), the model appears to integrate these low-level concepts into a broader contextual and structural understanding. The focus shifts toward how words and morphemes combine to form phrases and contribute to sentence-level meaning. There is increased sensitivity to a wider range of function words that govern grammatical relationships (e.g., conjunctions, prepositions, pronouns), verb forms, phrase structures, and higher-order elements like punctuation and proper nouns. These deeper layers seem to capture the "connective tissue" of the language, which is essential for coherence, fluency, and the overall structure of sentences.

### A.10.2 English-Specific Features

A more interesting pattern shows up in English, also suggesting a hierarchical pattern of linguistic processing. Early layers (layers 0 and 2) predominantly focus on low-level, foundational elements such as punctuation, function words, morphemes, and common short phrases, which are crucial for basic syntactic structure and token delimitation. Progressing to mid-layers (layers 4 and 8), the interpretations indicate a shift toward processing larger, multi-word phrases and clauses as cohesive semantic units that convey specific actions, states, relationships, and even context-dependent idiomatic expressions, thereby starting to capture more complex meaning. Finally, interpretations from the deeper layers (layer 15) suggest the model comprehends higher-level linguistic phenomena. This includes identifying structural elements for enumerating information in lists, recognizing concepts specific to different text styles (formal or conversational), understanding discourse structures, and identifying pragmatic elements such as spoken language patterns and dialogue flow. This overall progression demonstrates an increasing sophistication from detecting basic building blocks of language to understanding nuanced semantic content and complex discourse-level characteristics.

### A.10.3 French-Specific Features

The French interpretations reveal a consistent pattern of linguistic processing, progressing from foundational morphological elements in earlier layers to more complex syntactic and discourse-level structures in deeper layers. Initially, features in these earlier and middle layers (layers 5, 8, and 9) focus on identifying core word components such as morphemes, inflectional and derivational affixes, verb endings, and basic function words, which are crucial for word formation and elementary grammatical marking like tense or gender. As information propagates to the deeper layers (layers 11–14), the interpretations indicate an increasing capacity to recognize more abstract linguistic phenomena. This includes the identification of a broader range of function words (prepositions, conjunctions, pronouns, discourse markers), an understanding of phrase boundaries, syntactic relationships, sentence structuring, inter-clausal connections, and even the flow of conversation, suggesting a hierarchical abstraction of linguistic knowledge within the model.

### A.10.4 German-Specific Features

The German interpretations also reveal a hierarchical processing pattern. Initial layers (layers 2 and 5) primarily focus on decomposing German words into fundamental morphological units, such as morphemes, word stems, and common inflectional endings. As information progresses to intermediate layers (layers 8–9), the model begins to integrate these morphological components into broader grammatical contexts, identifying function words, common collocations, and basic syntactic structures like phrase boundaries. Finally, in the deeper layers (layers 10, 11, 12, 14, and 15), the interpretations demonstrate a more sophisticated understanding, capturing complex syntactic dependencies, multi-word expressions, phrasal patterns, and even specific grammatical constructs like verb-final clauses and subordinate clause markers, indicating a progression toward comprehending sentence-level structure and meaning.

### A.10.5 Italian-Specific Features

The Italian interpretations demonstrate a progression from earlier to deeper layers. Initially, in the earlier layers (layer 1), the focus is on identifying broad morphological components such as basic morphemes, affixes (prefixes and suffixes), roots, stems, and some function words, often related to word boundaries and fundamental grammatical constructs. As the analysis moves to early mid-layers (layers 2–4), there is a refinement, with features capturing more specific linguistic elements like named entities, comparatives, diminutives, and sometimes broadening the context to common patterns in Romance or other European languages. Progressing further, late-mid layers (layers 6–12) exhibit increased sophistication, recognizing the agglutinative and inflectional nature of Italian morphology and specializing in highly frequent, specific affixes, occasionally providing concrete examples of com-mon Italian suffixes. Finally, in the deeper layers (layers 13–14), interpretations often solidify the link between these morphological concepts and their syntactic roles, detailed grammatical categories, and the overall structure of morphologically rich languages, indicating a hierarchical build-up of linguistic understanding from foundational units to more complex grammatical functions.

### A.10.6 Korean-Specific Features

The Korean interpretations show a progression from general structural elements to nuanced, context-specific knowledge across model layers. The earlier layers (layers 0 and 4) predominantly focus on foundational Korean linguistic units, such as morphemes, basic grammatical particles (e.g., for subject or object marking), and core semantic building blocks essential for elementary sentence structure and meaning. As the model progresses to deeper layers, the features demonstrate increasing specialization and abstraction. This includes the identification of more complex grammatical nuances like specific verb endings related to tense or politeness (emerging around layer 9) and honorifics (noted in layer 10), an understanding of abstract thematic content such as historical events like those involving King Sejong (in layer 10 and recurring in layer 12) or specific domains such as travel and communication (appearing by layer 13 and expanding in layer 15), and a refined recognition of proper nouns and specialized terminology (with initial indications around layer 7 and more developed instances in layer 10, becoming broadly evident by layer 15). Thus, there is a trajectory from a general, structural understanding of the language toward more nuanced, context-aware, and specialized knowledge representation in the model's later layers.

### A.10.7 Chinese-Specific Features

The Chinese interpretations demonstrate a progression from foundational to complex linguistic understanding. Early active layers (e.g., layer 2) primarily identify basic semantic tokens and morphemes. As the model deepens, mid-layers (layers 6–10) begin to distinguish larger phrasal units like noun and verb phrases, idioms, and also show an emerging specialization between features capturing semantic content and those focused on grammatical structure. In the deeper layers (layers 11–15), this specialization intensifies, with features becoming more granular, identifying specific entities like proper nouns, handling context-specific language (e.g., scientific or news domains), and robustly processing diverse elements such as punctuation, complex idioms, and structural connectors in parallel, indicating a sophisticated, multi-faceted comprehension of the language.

### A.10.8 Russian-Specific Features

The Russian interpretations reveal a hierarchical pattern of feature abstraction, progressing from the identification of basic morphological units and general grammatical functions in earlier layers (e.g., layers 4, 6 focusing on morphemes, prefixes, and connectors) toward more granular and specific roles in the mid-layers (e.g., layers 7–12 detailing stems, suffixes, case, tense, and aspect). The deeper layers (layers 13–15) continue this morphological emphasis but increasingly integrate these features to interpret core semantic content, relationships, and their function in constructing meaning within complex sentences and specific contexts (e.g., in academic or formal text). However, we observed a notable deviation in feature 56594 in layer 10 (see Figure 61), which focuses on recognizing higher-level semantic categories like named entities before later layers synthesize morphological understanding for nuanced comprehension.

### A.10.9 Japanese-Specific Features

The Japanese interpretations reveal a progression in linguistic abstraction. Earlier layers (e.g., layer 0) primarily identify foundational elements such as specific grammatical particles and basic semantic categories like nouns or key morphemes. As information flows to intermediate layers (layers 8–11), the interpretations demonstrate an increasing ability to recognize more complex Japanese structures, including noun and verb phrases, alongside a stronger focus on function words and initial indicators of contextual understanding. The deeper layers (layers 12–15) showcase highly refined and specialized feature interpretations. These capture nuanced grammatical functions, including specific particles and politeness markers, complex semantic relationships, domain-specific terminology (e.g., legal, technical), and a sophisticated grasp of Japanese morphology and its agglutinative nature. This evolution signifies a hierarchical processing of linguistic information, advancing from discrete components toward integrated, context-aware representations of language. However, we observed a deviation for feature 32154 in layer 8 (see Figure 60). Though predominantly responsive to Japanese linguistic patterns and ex-amples, it also shows higher activation for specific Bulgarian morphemes, hinting that some features, while most relevant to the primary language of analysis, can also capture distinct patterns in other languages.

#### A.10.10 Vietnamese-Specific Features

The Vietnamese interpretations across layers reveal support for a hierarchical processing pattern for Vietnamese. Earlier layers (e.g., layers 0, 3, 4) predominantly focus on decomposing text into fundamental linguistic units, such as morphemes, syllables, function words, and elementary phrases, thereby establishing a foundational understanding. As information propagates to subsequent, mid-level layers (e.g., layers 7, 10, 12), features begin to assemble these components into more complex structures, identifying compound nouns, nuanced semantic roles, and key content elements. Consequently, deeper layers (e.g., layers 13–15) demonstrate the most abstract comprehension, interpreting larger semantic units like full phrases and collocations, recognizing tokens within specific discourse or pragmatic contexts (e.g., formal, conversational), and integrating diverse linguistic cues for a holistic understanding of the input. However, feature 49501 in layer 5 (see Figure 62), while mostly processing Vietnamese, also strongly detects elements from Romanian and Eastern European place names, suggesting it captures some cross-lingual patterns. Similarly, feature 61383 in layer 9 (see Figure 63) for Vietnamese also shows high activation for Cambodian place names, indicating it can specialize in recognizing frequently mentioned foreign entities.

#### A.10.11 Bulgarian-Specific Features

The Bulgarian interpretations across its layers exhibit a hierarchical progression in complexity and abstraction. Initially, in earlier layers (e.g., layers 0–1), the focus is on foundational elements such as single Cyrillic letters, basic positional cues, and common suffixes tied to elementary word formation and fundamental grammatical features like gender and number. As processing moves to the middle layers (layers 3–9), the analysis deepens into more comprehensive morphological decomposition, identifying a wider array of morphemes like stems, prefixes, and diverse suffixes, alongside function words such as pronouns and conjunctions, and their roles in establishing core grammatical relationships, occasionally recognizing patterns shared with other Slavic languages. Finally, in the deeper layers (layers 10–15), the interpretations demonstrate an understanding of more complex syntactic roles and semantic nuances; features here often describe how these identified morphemes and function words contribute to constructing larger linguistic units, such as clauses, complex verb phrases, expressing modality, forming idiomatic expressions, and recognizing specific grammatical constructions like comparatives or superlatives, thereby indicating a grasp of higher-level sentence structure and meaning.

#### A.10.12 Portuguese-Specific Features

The Portuguese interpretations across the model layers exhibit a hierarchical pattern of linguistic abstraction. The earlier layers (layers 0–5) primarily focus on identifying fundamental building blocks, such as specific high-frequency function words (e.g., the preposition "no" in layer 0) and basic morphological units like general function words, suffixes, and morphemes. Progressing to the middle layers (layers 7–12), the model demonstrates an enhanced capability to discern more specific morphological components (e.g., roots, stems, and affixes) and their roles in word formation (verb conjugations, noun/adjective forms), and begins to link these to semantic entities (e.g., "Brazil" in layer 7). Finally, the deeper layers (layers 13–15) showcase a sophisticated synthesis, where interpretations reflect an understanding of how these accumulated morphological elements and function words, often in conjunction with punctuation, contribute to complex grammatical structures, sentence cohesion, nuanced meanings such as modality or conditional relationships, and the overall construction of coherent text. This trajectory suggests a process that starts with basic language elements and moves toward understanding how they work together to create complex meaning.

#### A.10.13 Turkish-Specific Features

The Turkish interpretations exhibit a hierarchical progression across its layers. In early layers (layers 0–7), the model begins with a foundational focus on deconstructing words into their core components, with features dedicated to identifying fundamental morphemes, suffixes, and roots essential for the language's agglutinative structure. As processing moves to the middle layers (layers 8–12), a pattern of specialization emerges, with features becoming more attuned to specific grammatical categories like verb morphology and distinct semantic concepts such as named entities. Finally, in the deeper layers (layers 13–15), the model achieves a higher level of abstraction, developing features that interpret entire syntactic units, such as noun and verb phrases, and recognize language within specific semantic domains (e.g., administrative, legal, travel, and service-related contexts), all while continuously leveraging the foundational morphological features. An exception occurs with feature 97688 in layer 10 (see Figure 59). Although this feature activates mainly on examples within the Turkish corpus, it shows higher activations in identifying verb endings in Romance languages.

#### A.10.14 Thai-Specific Features

The Thai interpretations reveal a hierarchical pattern of linguistic processing, progressing from foundational to abstract representations. In the earlier layers (layers 0–4), the model predominantly identifies fundamental morphological and syntactic units such as morphemes, syllables, and function words, focusing on the basic grammatical scaffolding of the Thai language. As we move to mid-layers (layers 5–9), these foundational elements are increasingly associated with concrete semantic categories and specific named entities, such as the literal concept of "Thailand." In the deeper layers (layers 10–15), the interpretations demonstrate a more sophisticated and abstract understanding, capturing complex semantic relationships like causality and temporality, as well as nuanced pragmatic functions. This progression shows a trajectory from deconstructing language into its structural components to synthesizing meaning and understanding its use in complex, often cross-lingual, contexts. Notably, feature 8775 in layer 12 (see Figure 64) describes phenomena that seem exclusive to other languages, such as politeness markers in Japanese and Korean, even though the feature itself is also highly activated within the Thai corpus.

#### A.10.15 Hindi-Specific Features

The Hindi interpretations exhibit a hierarchical progression in linguistic comprehension. Initially (layers 0–4), the model identifies fundamental, high-frequency morphemes such as postpositions, auxiliary verb components, and inflectional vowel diacritics, which form the basic syntactic building blocks of Hindi. As information propagates to the middle layers (layers 5–10), these components are assembled into more specific grammatical constructions, such as distinct verb conjugations and pronouns. Toward the end of this stage, particularly in layer 10, the model shows the first signs of semantic abstraction, moving beyond pure syntax with features that activate on specific geopolitical entities, including the country Pakistan, and related historical concepts. Finally, in the deepest layers (layers 11–15), this ability to interpret context becomes more sophisticated and widespread. The model builds upon its initial entity recognition to more consistently identify a broader range of named entities, locations, and organizations, completing a trajectory from recognizing grammatical form to comprehending contextual meaning.

### A.11 Task Experiment

Although language-specific features influence PPL, our experiments on the XWinograd task (Muennighoff et al., 2023) suggest they do not reliably impact downstream task performance. We conduct two experiments to evaluate this. In the first experiment (Table 2), suppressing language-specific features across all layers may significantly disrupt performance across languages. In contrast, the second experiment (Table 3) involves suppressing these features only in the middle layers, where their influence is presumed to be less critical, and results in negligible performance changes. These findings align with prior work showing that while language-specific neurons may affect PPL, they do not necessarily degrade task performance (Mondal et al., 2025). This distinction may explain why researchers differentiate between language-specific and task-specific neurons (e.g., Tang et al., 2024; Kojima et al., 2024; Leng and Xiong, 2025), with the latter playing a more crucial role in enabling effective multilingual task execution. This also suggests that task-specific features are likely to exist in LLMs.

| Task Language | Baseline | en | fr | ja | pt | ru | zh |
|---|---|---|---|---|---|---|---|
| en | 0.8129 | 0.5196 | 0.8116 | 0.5166 | 0.5118 | 0.5768 | 0.5923 |
| fr | 0.6506 | 0.4940 | 0.6506 | 0.4699 | 0.4217 | 0.5181 | 0.5181 |
| ja | 0.5996 | 0.5328 | 0.6111 | 0.5172 | 0.5203 | 0.5255 | 0.5276 |
| pt | 0.6616 | 0.5171 | 0.6616 | 0.5019 | 0.5171 | 0.5133 | 0.5057 |
| ru | 0.5968 | 0.5397 | 0.5810 | 0.5587 | 0.5429 | 0.5556 | 0.5143 |
| zh | 0.6389 | 0.4861 | 0.6468 | 0.4861 | 0.4802 | 0.5655 | 0.5298 |

Table 2: Accuracy on the XWinograd task under language-specific feature interventions applied across all layers of Llama 3.2 1B with a scaling factor of $\alpha = -0.2$. Each column corresponds to an intervention targeting a specific language.

| Task Language | Baseline | en | fr | ja | pt | ru | zh |
|---|---|---|---|---|---|---|---|
| en | 0.8129 | 0.8077 | 0.8095 | 0.8120 | 0.7931 | 0.8168 | 0.8090 |
| fr | 0.6506 | 0.6627 | 0.6024 | 0.6506 | 0.6747 | 0.6506 | 0.6386 |
| ja | 0.5996 | 0.6017 | 0.5923 | 0.6121 | 0.5912 | 0.5933 | 0.5850 |
| pt | 0.6616 | 0.6654 | 0.6692 | 0.6616 | 0.6046 | 0.6692 | 0.6654 |
| ru | 0.5968 | 0.6032 | 0.6063 | 0.6095 | 0.6317 | 0.5968 | 0.6413 |
| zh | 0.6389 | 0.6567 | 0.6290 | 0.6429 | 0.6429 | 0.6548 | 0.6369 |

Table 3: Accuracy on the XWinograd task under language-specific feature interventions applied from layers 3–13 of Llama 3.2 1B with a scaling factor of $\alpha = -0.2$. Each column corresponds to an intervention targeting a specific language.

### A.12 Relationship between Language-Specific Features and SAE Training Data

The existence of language-specific features could be related to the frequency of specific concepts (in this case, related to languages) in the SAE training data, which influences the dedicated learned features for those concepts, with the more frequent ones likely to have dedicated features, as described by Templeton et al. (2024). RedPajama-v2 is not uniformly distributed, with English being the dominant language, as shown in Table 4. Interestingly, all major languages in RedPajama-v2 (en, de, fr, es, and it) have fewer language-specific features than the other languages (see Figure 20). This suggests that training SAEs on uniformly distributed multilingual corpora can help them learn language-specific concepts more evenly.

| Language | Row Count | Percentage |
|---|---|---|
| en | 4,730,801 | 74.02% |
| de | 531,629 | 8.32% |
| fr | 433,604 | 6.78% |
| es | 459,208 | 7.18% |
| it | 236,397 | 3.70% |

Table 4: Language distribution and row counts in the RedPajama-Data-V2 `Sample-10B` subset

### A.13 Text Generation Assessment Rubric

To assess steered generated text, we developed a rubric targeting two key criteria: (1) language occurrence and (2) language coherence. This rubric is designed to be applied manually by humans or automatically with an LLM, and aims to ensure consistency and transparency in evaluating whether and how language-specific features affect generated outputs. In this study, we use GPT-4.1 (OpenAI, 2025) to automatically evaluate the generated text according to this rubric.

#### A.13.1 Language Occurrence

This criterion captures whether a meaningful language change has occurred in the generated text. A change is considered to have occurred if the output contains at least one *morpheme* (the smallest meaningful unit of language) from a different language than English.

- **Score 0 (Unchange)**: The text remains fully in English and contains no morphemes from another language. Minor orthographic changes that do not introduce new meaning or reflect actual cross-linguistic variation are *not counted* as a language change. For example:
  - `apple` → `åpple` (still English, only orthographic variation)
  - `naive` → `naïve` (still English usage)
- **Score 1 (Change)**: At least one morpheme from a different language is introduced in the output, even if the change occurs only in part of the sentence. For example:
  - `apple` → `manzana` (Spanish)
  - `He is a teacher` → `Er ist ein Lehrer` (German)

#### A.13.2 Language Coherence

This criterion evaluates the overall intelligibility and fluency of the generated text when steered toward multilingual expression. It is assessed along a three-point ordinal scale:

- **Incoherent**: The text is unintelligible and does not form meaningful sentences. It may consist of repeated characters, morphemes, or nonsensical word sequences. For example:
  - `blblblblbl èèèè $$$$$`
  - `rerere banana moon qwerty`
- **Partially Coherent**: The general meaning of the text can be inferred, but parts may require guessing or interpretation. This includes major grammatical or syntactic errors, repetitive yet intelligible structures, or repetitions at the end of the text. For example:
  - `He go to escuela because want aprender`
  - `The cat is sleep sleep the bed the bed`

- **Coherent**: The text is readable and understandable, with only minor issues (e.g., minor grammatical or syntactic errors and mid-sentence code-switching that does not obscure meaning). Code-switching is acceptable as long as overall sentence-level coherence is maintained. For example:
  - `She goes to the bibliothèque every evening to study`
  - `I like to eat sushi y también tacos`
  - `Er ist ein Lehrer und he teaches German very well`

This rubric allows researchers to both quantify the presence of cross-linguistic influence and evaluate the intelligibility of the generated text, supporting a robust evaluation of multilingual steering methods. Note that this rubric does not assess the correctness of the outputs. For example, the generated text may contain logical errors (as illustrated in Figures 72–74) or may include inaccurate information.

### A.14 Misclassification Analysis of SAE-based LID

For the misclassified example, we analyze the following German sentence from WiLI-2018: "Olga Alexandrowna Girja (russisch Ольга Александровна Гиря; * 4. Juni 1991 in Langepas) ist eine russische Schachspielerin und seit 2009 Großmeister der Frauen (WGM)." For complete details on feature activations, refer to Figures 80–84. The lower layers (0–3) do recognize some German elements, such as "isch", "ist", and "ß". However, a critical shift occurs in the middle layers, where Russian features begin to dominate. By layer 6, Russian feature 121507 activates on Cyrillic script elements, such as "ль", "га", " Александ", and " Г", and its activation intensifies in higher layers. The decisive factor appears to be feature 56594 in layer 10, which activates on proper nouns and named entities across both scripts, including the transliterated Russian name in Latin script ("ga", " Alexand", "row", "na", "Gir", "ja") and the Cyrillic characters ( "О", "ль", "га", "Александ", "ров", "на", "Г", "ир", "я"). While German features persist in later layers, they are ultimately overshadowed by the Cyrillic script and Russian name, which trigger more consistent activations in the deeper layers. Notably, the SAEs contain 13 German-specific features compared to 26 Russian-specific features, as shown in Figure 20, which may further amplify this imbalance. Taken together, this behavior suggests that the current approach may be biased toward languages with a larger number of features, especially in mixed-language inputs.

### A.15 Additional Results

For completeness, this section presents the results related to language-specific neurons and language-specific features for Llama 3.2 1B, Gemma 2 2B, and Gemma 2 9B.

#### A.15.1 Language-Specific Neuron Distributions

Figures 12–14 show the distribution of language-specific neurons across layers for Llama 3.2 1B, Gemma 2 2B, and Gemma 2 9B.

#### A.15.2 Language-Specific Neuron Counts

Figures 15–17 show the language-specific neuron counts across languages for Llama 3.2 1B, Gemma 2 2B, and Gemma 2 9B.

#### A.15.3 Language-Specific Feature Distributions

Figures 18–19 show the distribution of language-specific features across layers for Gemma 2 2B and Gemma 2 9B.

#### A.15.4 Language-Specific Feature Counts

Figures 20–22 show the language-specific feature counts across languages for Llama 3.2 1B, Gemma 2 2B, and Gemma 2 9B.

#### A.15.5 Language-Specific Feature LAPE Scores

Figures 23–25 show that the majority of the identified language-specific features in Llama 3.2 1B, Gemma 2 2B, and Gemma 2 9B have low LAPE scores.

#### A.15.6 Perplexity Change

Figures 26–28 illustrate the change in perplexity when steering with language-specific features and language-specific neurons in Llama 3.2 1B across different scaling factors $\alpha$. The corresponding results for Gemma 2 2B are shown in Figures 29–32. Overall, the patterns are consistent across models, as explained in Section 5.2.1.

#### A.15.7 LID with Language-Specific Features

Figure 33 and Table 5 present the SAE-based, neuron-based, and fastText LID performance on WiLI-2018 across 15 languages for Gemma 2 2B.

Figure 34 and Table 6 present the corresponding results for Gemma 2 9B. Overall, the results exhibit patterns largely consistent with those observed for Llama 3.2 1B in Section 5.3.

However, performance degrades for several languages. We hypothesize that this is due to a bias toward languages with a larger number of language-specific features, as discussed in Section 5.3. This bias is more pronounced in JumpReLU SAEs than in TopK SAEs, since JumpReLU SAEs do not constrain the number of activations per token. As a result, languages with more language-specific features may activate a disproportionately large number of features, thereby amplifying the bias, even when the activations are small. In contrast, TopK SAEs explicitly limit the number of activations per token to the top-$k$ values, which may partially mitigate this effect. This behavior is particularly evident for English, which has the fewest language-specific features among the 15 languages, as shown in Figure 21 for Gemma 2 2B and Figure 22 for Gemma 2 9B.

Moreover, these results may be influenced by the multilingual capabilities of Gemma 2 2B, Gemma 2 9B, and their corresponding SAEs, which are predominantly trained on English data (Team et al., 2024; Lieberum et al., 2024). For Turkish, both Gemma 2 2B and Gemma 2 9B exhibit poor performance, suggesting that the models and their SAEs may have limited multilingual capabilities for this language. In contrast, Gemma 2 2B exhibits poor performance on Italian, whereas Gemma 2 9B performs well on this language, suggesting that Gemma 2 9B may have stronger multilingual capabilities for Italian than Gemma 2 2B due to its larger parameter count.

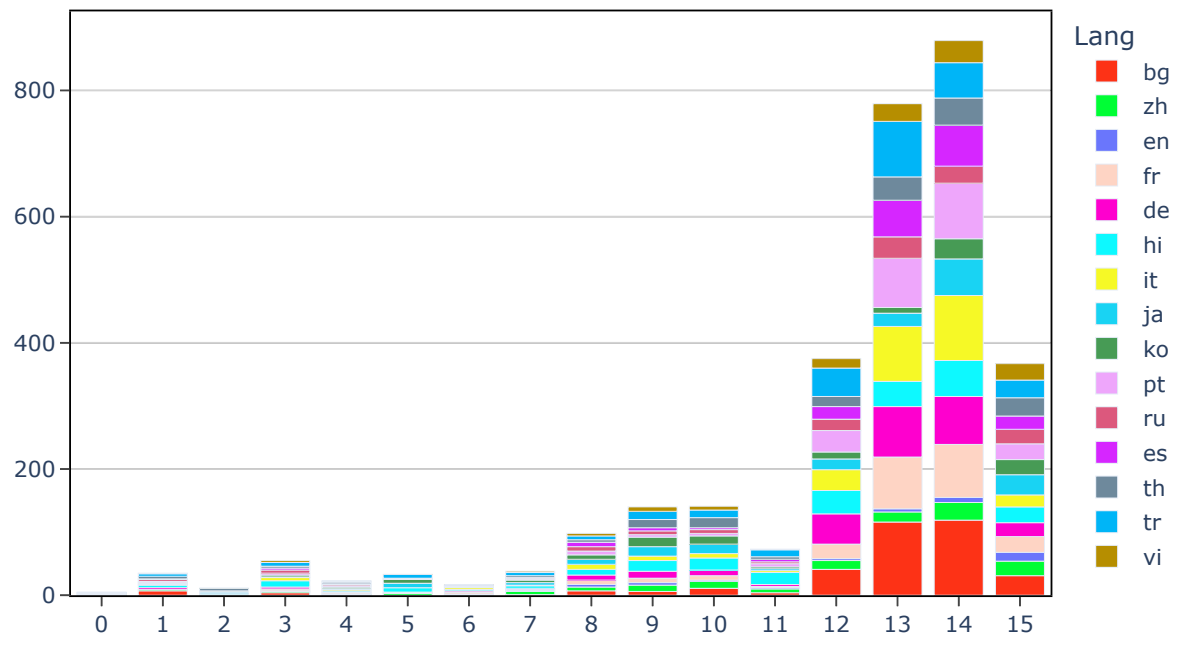

Figure 12: Distribution of language-specific neurons identified by *LAPE* across layers of Llama 3.2 1B.

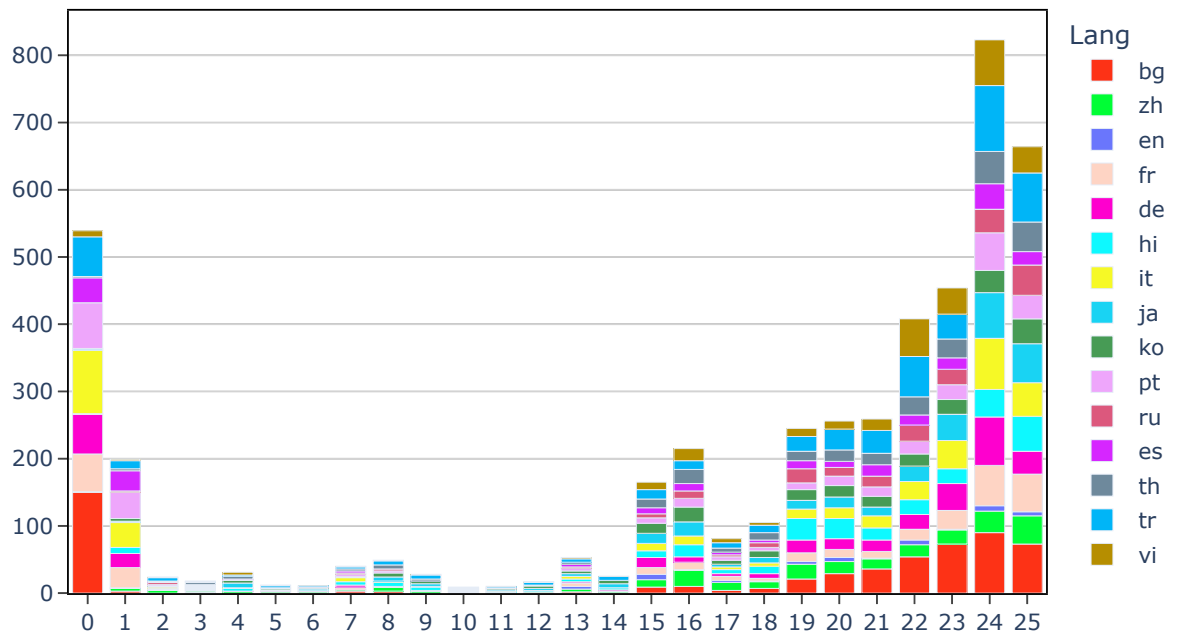

Figure 13: Distribution of language-specific neurons identified by *LAPE* across layers of Gemma 2 2B.

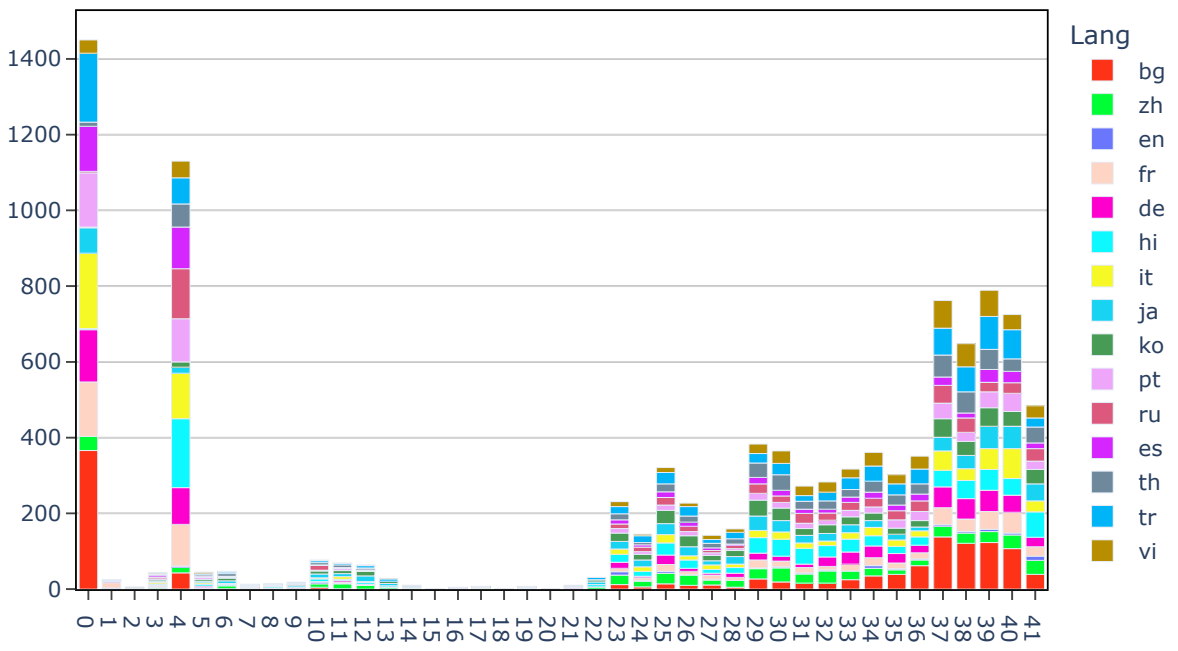

Figure 14: Distribution of language-specific neurons identified by *LAPE* across layers of Gemma 2 9B.

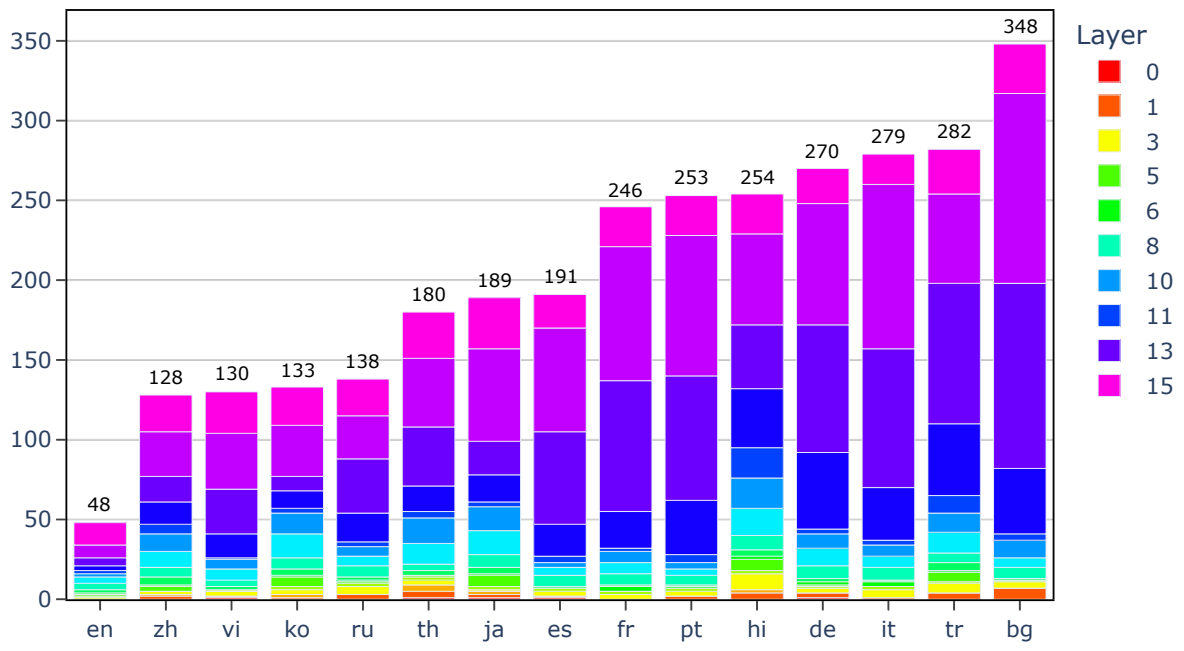

Figure 15: Language-specific neuron counts in Llama 3.2 1B across languages.

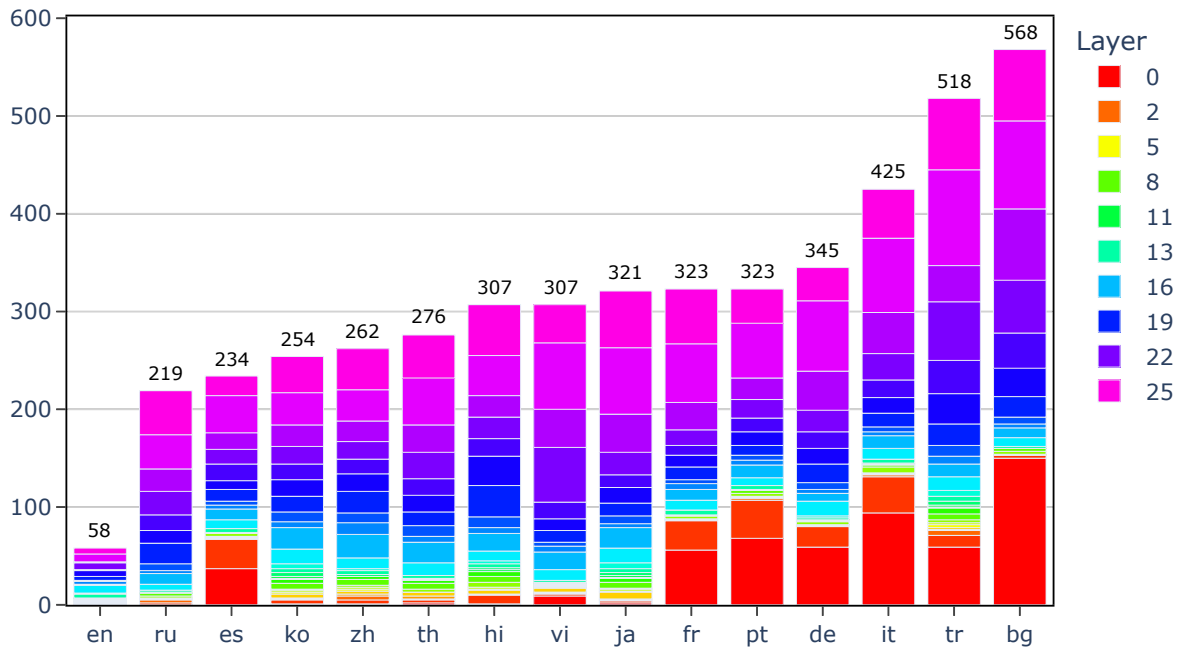

Figure 16: Language-specific neuron counts in Gemma 2 2B across languages.

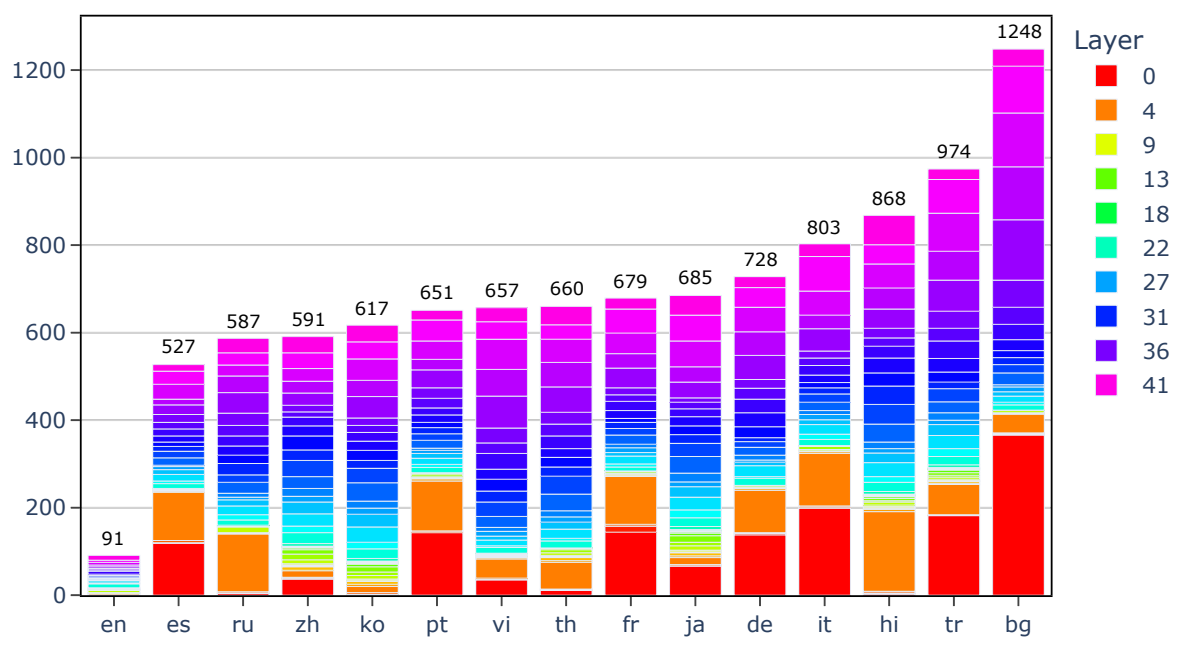

Figure 17: Language-specific neuron counts in Gemma 2 9B across languages.

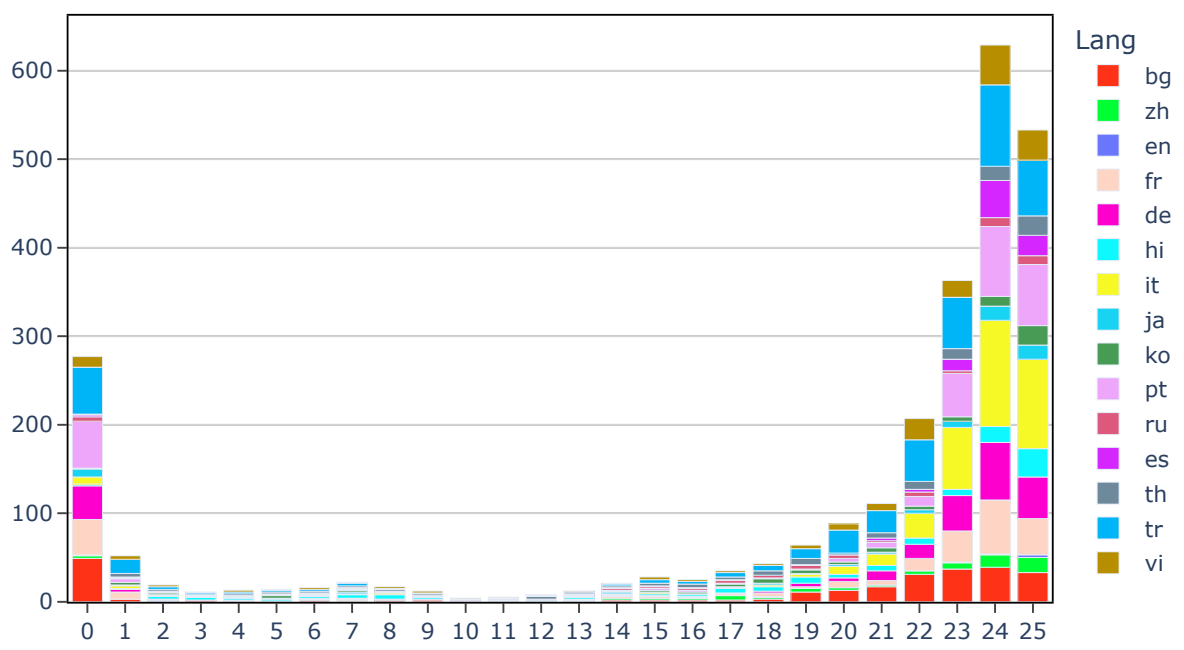

Figure 18: Distribution of language-specific features identified by *SAE-LAPE* across layers of Gemma 2 2B.

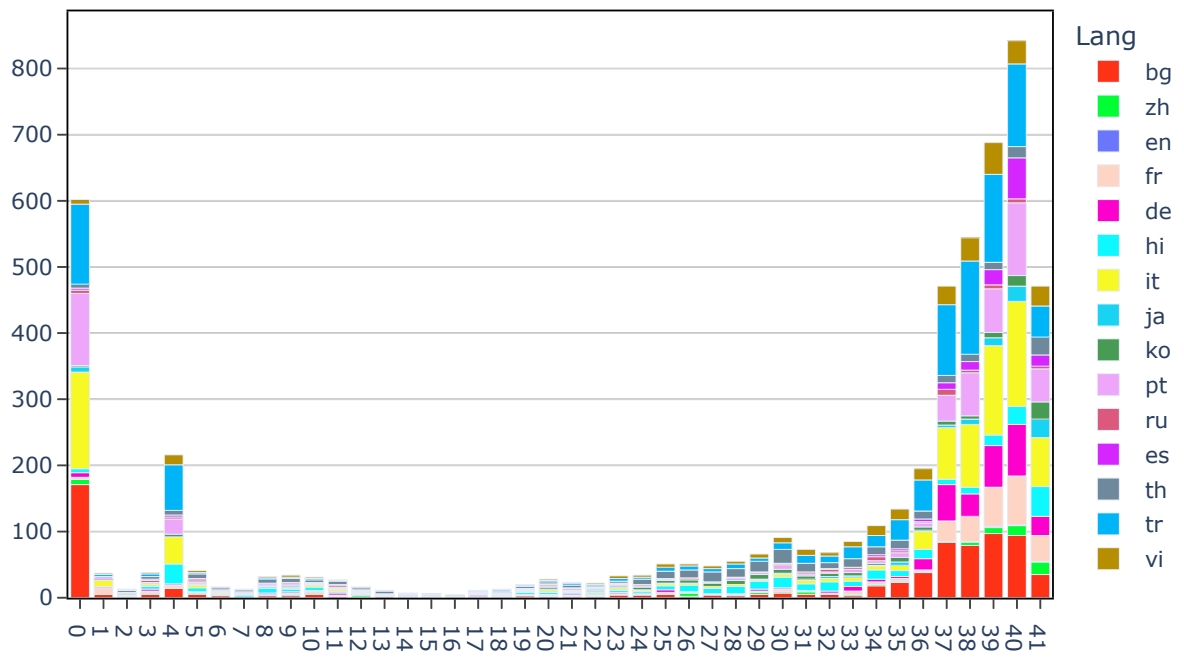

Figure 19: Distribution of language-specific features identified by *SAE-LAPE* across layers of Gemma 2 9B.

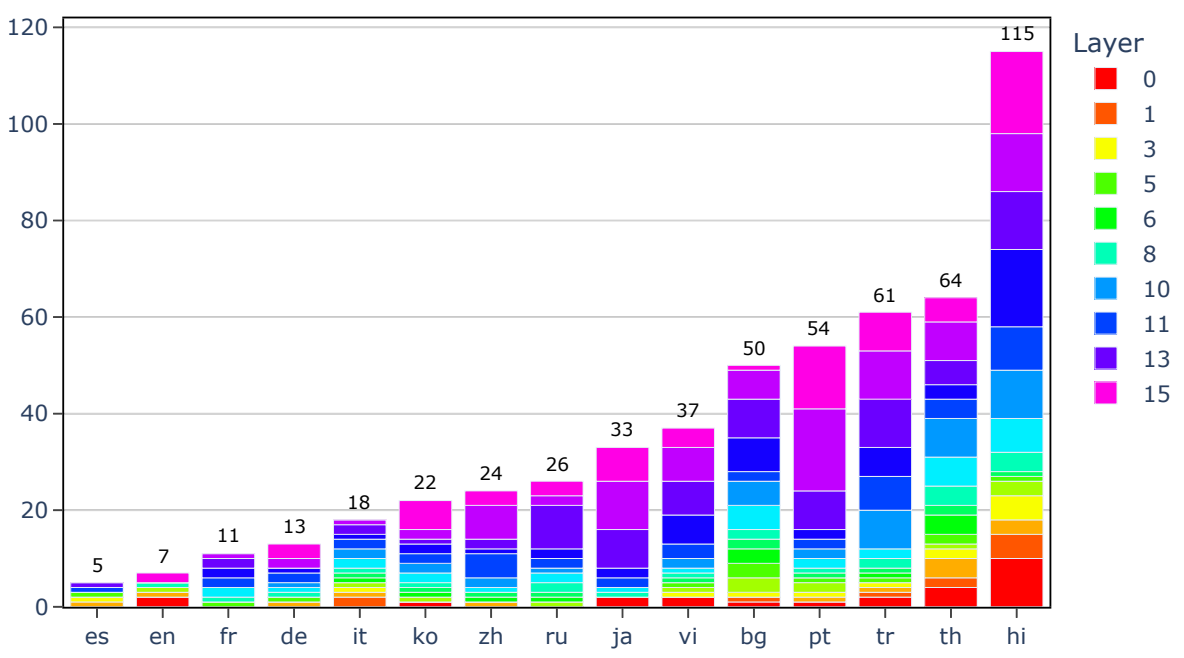

Figure 20: Language-specific feature counts in Llama 3.2 1B across languages.

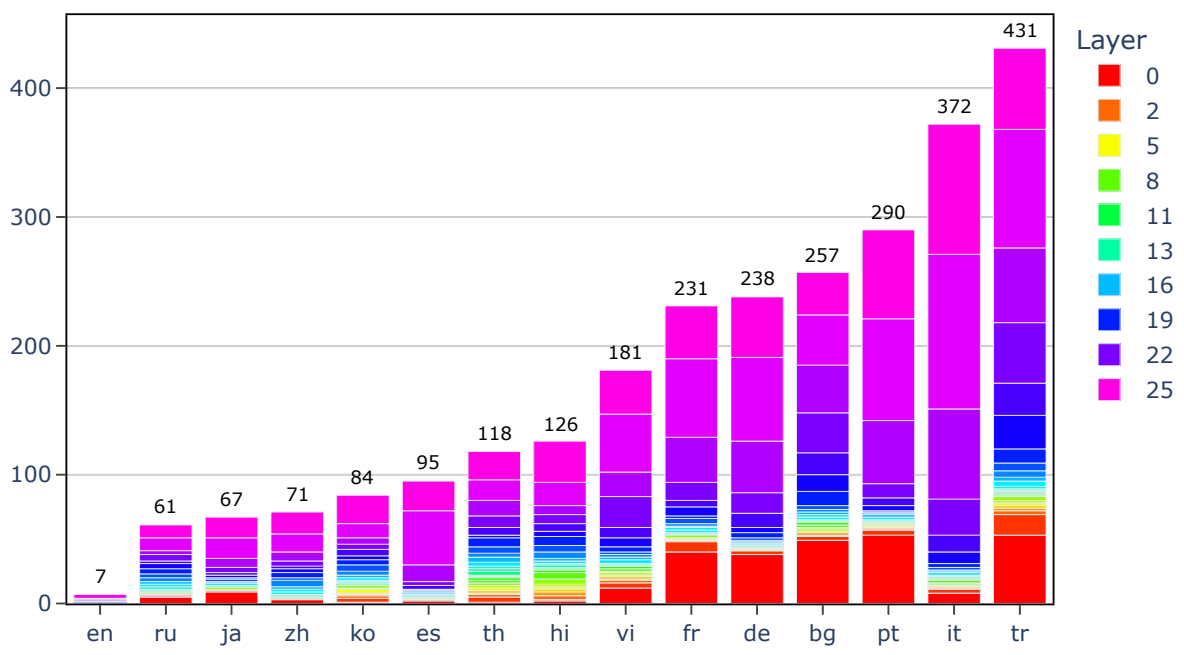

Figure 21: Language-specific feature counts in Gemma 2 2B across languages.

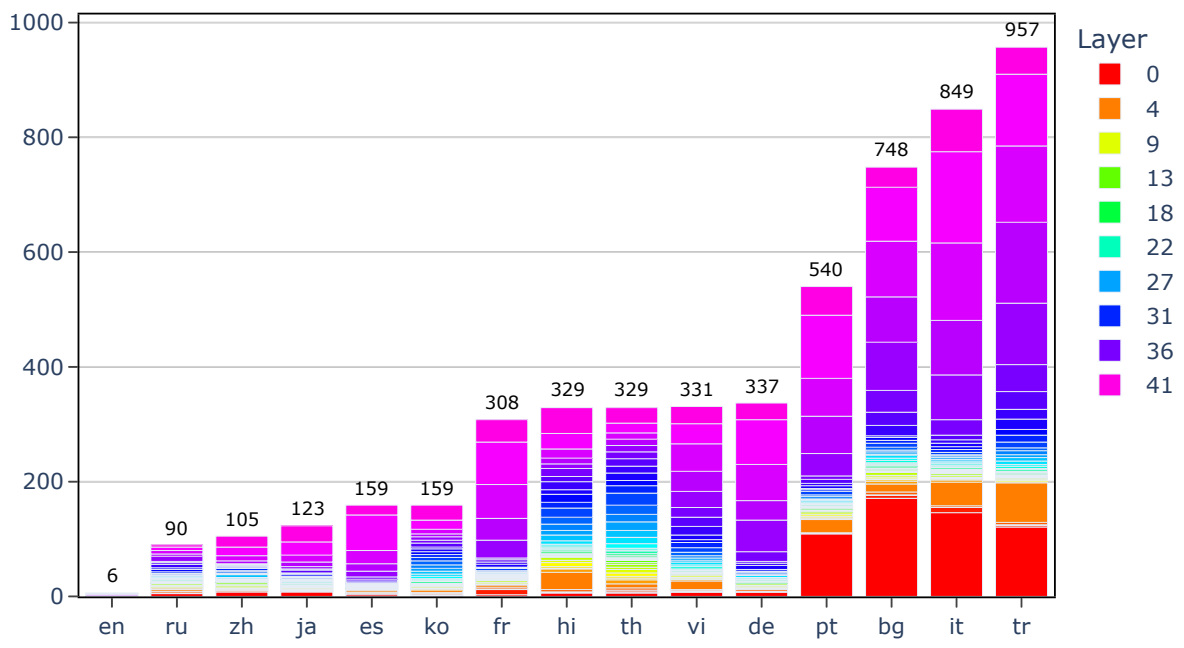

Figure 22: Language-specific feature counts in Gemma 2 9B across languages.

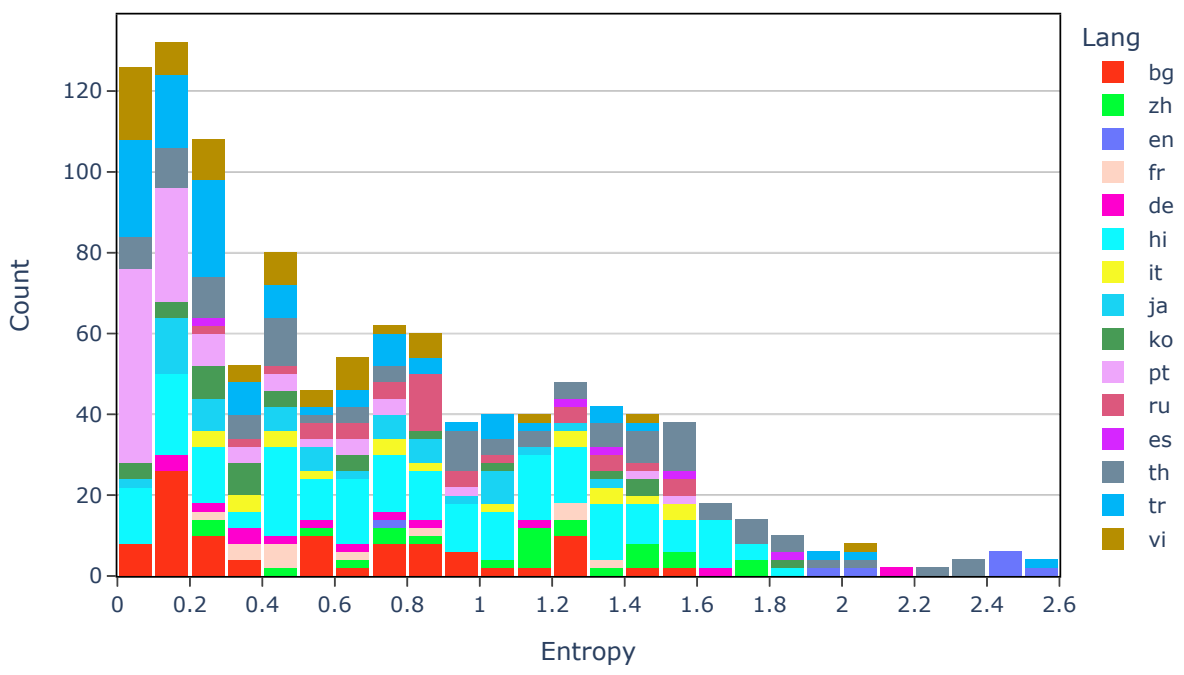

Figure 23: Distribution of language-specific features identified by *SAE-LAPE* in Llama 3.2 1B across LAPE scores.

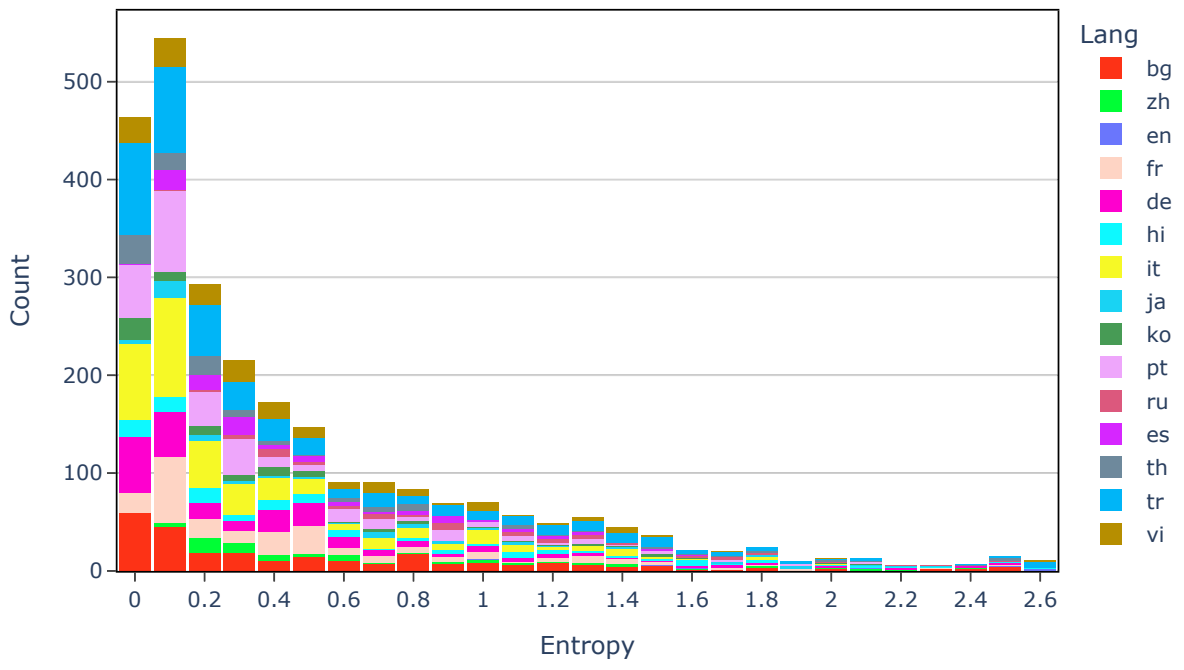

Figure 24: Distribution of language-specific features identified by *SAE-LAPE* in Gemma 2 2B across LAPE scores.

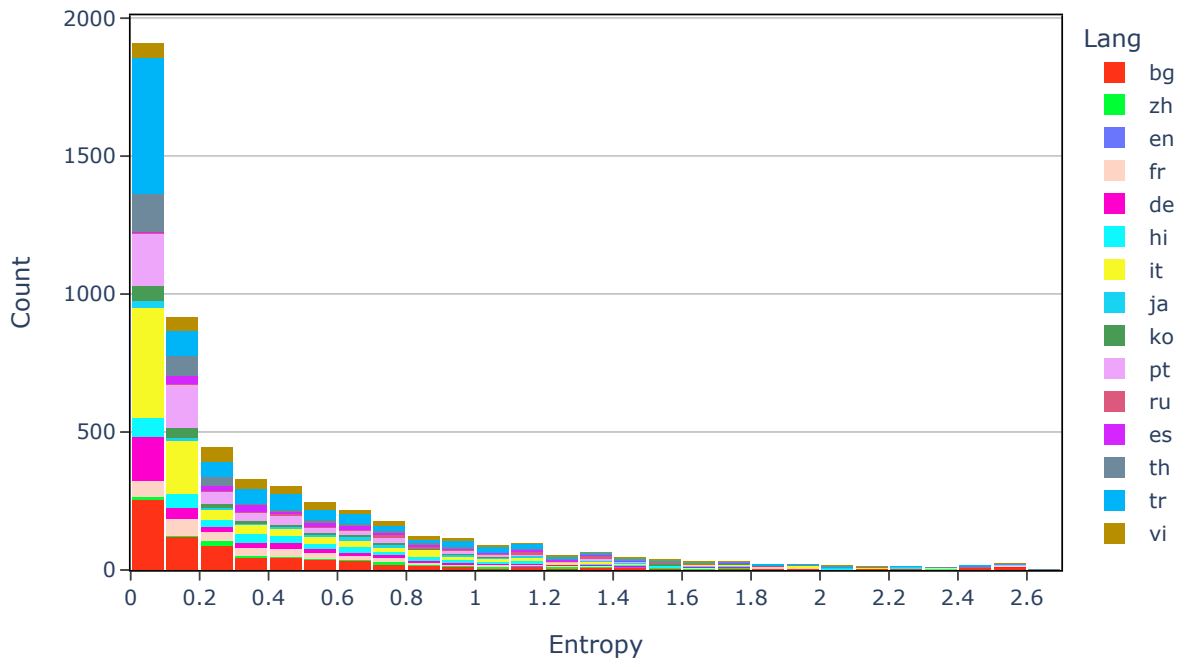

Figure 25: Distribution of language-specific features identified by *SAE-LAPE* in Gemma 2 9B across LAPE scores.

| | en | de | fr | it | pt | hi | es | th | bg | ru | tr | vi | ja | ko | zh |
|---|---|---|---|---|---|---|---|---|---|---|---|---|---|---|---|
| en | 0.8 | 0.3 | 0.2 | 0.2 | 0.4 | -0.1 | 0.4 | -0.2 | 0.0 | 0.0 | 0.2 | 0.8 | -0.2 | -0.1 | -0.1 |
| de | 0.8 | 2.1 | 0.4 | 0.6 | 0.9 | 0.2 | 0.8 | 0.1 | 0.7 | 0.5 | 2.2 | 1.5 | 0.9 | 0.9 | 1.0 |
| fr | 1.0 | 1.1 | 2.5 | 1.4 | 2.0 | 0.5 | 1.3 | 0.9 | 1.4 | 1.0 | 4.2 | 2.8 | 1.8 | 2.0 | 2.1 |
| it | 1.1 | 0.6 | 0.5 | 65.2 | 1.4 | 0.3 | 0.6 | 0.6 | 3.6 | 0.9 | 3.7 | 2.1 | 1.9 | 1.7 | 2.5 |
| pt | 4.5 | 3.4 | 3.2 | 4.0 | 484.0 | 0.9 | 5.1 | 2.0 | 4.7 | 2.6 | 11.0 | 8.8 | 4.1 | 5.5 | 8.3 |
| hi | 5.1 | 6.4 | 4.1 | 4.8 | 6.4 | 20.8 | 5.0 | 9.4 | 13.9 | 6.6 | 27.4 | 15.0 | 15.3 | 17.8 | 20.4 |
| es | 0.5 | 0.6 | 0.5 | 0.7 | 6.8 | 0.3 | 18.5 | 0.4 | 0.4 | 1.1 | 3.8 | 1.8 | 0.5 | 1.2 | 2.6 |
| th | 3.2 | 2.3 | 1.4 | 1.6 | 2.0 | 1.2 | 1.8 | 100.8 | 3.1 | 2.2 | 10.7 | 7.5 | 6.7 | 7.2 | 6.9 |
| bg | 0.9 | 1.6 | 1.0 | 1.5 | 1.8 | 0.4 | 1.2 | 1.0 | 20.0 | 0.9 | 4.1 | 1.8 | 2.1 | 2.6 | 2.8 |
| ru | 0.7 | 0.4 | 0.3 | 0.4 | 0.3 | 0.5 | 0.4 | 0.5 | 1.5 | 1.0 | 2.1 | 1.4 | 1.7 | 2.2 | 1.7 |
| tr | 2.7 | 4.1 | 2.3 | 3.0 | 5.0 | 1.7 | 3.9 | 3.3 | 3.8 | 3.0 | $10^3$ | 9.1 | 7.0 | 9.8 | 8.8 |
| vi | 1.0 | 1.1 | 1.0 | 1.3 | 2.1 | 0.7 | 1.2 | 1.3 | 1.9 | 1.1 | 5.8 | 9.0 | 1.4 | 2.9 | 2.9 |
| ja | 1.0 | 0.9 | 0.7 | 0.9 | 0.6 | 0.4 | 0.6 | 1.4 | 1.8 | 0.7 | 4.0 | 2.5 | 8.7 | 2.4 | 4.4 |
| ko | 1.0 | 0.5 | 0.5 | 0.3 | 1.1 | 0.4 | 0.7 | 0.7 | 0.5 | 1.0 | 3.7 | 1.5 | 2.4 | 103.1 | 1.6 |
| zh | 0.9 | 0.5 | 0.2 | 0.4 | 0.6 | 0.3 | 0.4 | 1.6 | 0.9 | 0.7 | 1.7 | 1.0 | 1.2 | 2.2 | 3.2 |

| | en | de | fr | it | pt | hi | es | th | bg | ru | tr | vi | ja | ko | zh |
|---|---|---|---|---|---|---|---|---|---|---|---|---|---|---|---|
| en | -0.1 | 0.3 | 0.2 | 0.1 | 0.2 | 0.2 | 0.1 | 0.5 | 0.3 | 0.3 | 1.9 | 1.2 | 0.6 | 0.8 | 0.6 |
| de | 0.6 | -0.5 | 0.5 | 0.7 | 0.9 | 0.2 | 0.6 | 1.1 | 1.7 | 0.7 | 5.2 | 2.2 | 0.9 | 1.5 | 1.0 |
| fr | 0.6 | 2.7 | -0.5 | 11.2 | 13.7 | 0.8 | 6.4 | 1.8 | 3.3 | 1.3 | 12.2 | 4.9 | 1.8 | 2.3 | 1.7 |
| it | 0.7 | 4.2 | 201.8 | -0.5 | 94.2 | 0.7 | 113.8 | 2.3 | 5.9 | 3.3 | 20.7 | 5.9 | 1.5 | 5.6 | 0.5 |
| pt | 6.3 | 6.7 | 88.8 | 140.9 | -0.6 | 2.9 | 322.5 | 7.8 | 15.8 | 16.0 | 128.8 | 18.8 | 6.5 | 8.2 | 6.2 |
| hi | 7.3 | 7.3 | 4.1 | 5.2 | 7.2 | 2.9 | 6.1 | 44.1 | 39.4 | 10.9 | 286.9 | 22.2 | 19.5 | 63.6 | 21.6 |
| es | 0.5 | 2.0 | 32.0 | 32.5 | 7.6 | 0.3 | -0.7 | 0.7 | 4.1 | 1.6 | 6.3 | 7.2 | 1.1 | 0.6 | -0.7 |
| th | 2.2 | 1.9 | 1.4 | 1.6 | 2.8 | 1.2 | 1.9 | 0.9 | 3.3 | 2.0 | 9.1 | 9.2 | 2.3 | 6.3 | 5.0 |
| bg | 1.4 | 2.4 | 1.4 | 2.3 | 3.7 | 0.8 | 2.0 | 1.9 | -1.8 | 7.2 | 13.0 | 2.7 | 1.9 | 1.9 | 1.8 |
| ru | 0.7 | 1.5 | 0.8 | 1.1 | 1.8 | 0.8 | 1.1 | 2.0 | 4.6 | 0.4 | 6.1 | 2.3 | 1.4 | 2.7 | 1.8 |
| tr | 3.2 | 5.2 | 2.1 | 4.2 | 5.2 | 9.7 | 2.8 | 7.2 | 25.6 | 4.4 | -2.1 | 10.5 | 8.8 | 15.7 | 4.5 |
| vi | 1.2 | 1.6 | 0.8 | 1.2 | 1.8 | 1.0 | 1.3 | 12.4 | 3.3 | 1.2 | 7.4 | -2.0 | 3.2 | 3.2 | 3.3 |
| ja | 2.4 | 1.8 | 1.2 | 1.1 | 2.7 | 1.4 | 1.7 | 4.2 | 1.7 | 1.7 | 6.9 | 3.9 | -0.6 | 10.8 | 9.6 |
| ko | 1.4 | 1.6 | 0.7 | 1.4 | 1.0 | 1.3 | 1.0 | 7.4 | 3.2 | 0.8 | 6.3 | 2.7 | 22.4 | 1.8 | 4.4 |
| zh | 0.9 | 1.4 | 0.9 | 1.0 | 1.3 | 0.8 | 1.1 | 3.4 | 2.1 | 1.0 | 6.2 | 4.0 | 4.2 | 5.7 | 0.3 |

| | en | de | fr | it | pt | hi | es | th | bg | ru | tr | vi | ja | ko | zh |
|---|---|---|---|---|---|---|---|---|---|---|---|---|---|---|---|
| en | 0.1 | 0.4 | 0.0 | 0.2 | 0.3 | 0.3 | -0.1 | 1.0 | 1.7 | 0.9 | 0.8 | 2.9 | 1.2 | 3.7 | 2.0 |
| de | 0.4 | 1.8 | 0.2 | 0.4 | 0.8 | 0.2 | 0.4 | 0.5 | 0.9 | 0.5 | 2.4 | 1.3 | 0.6 | 1.4 | 1.1 |
| fr | 0.9 | 1.1 | 1.6 | 2.2 | 3.6 | 0.6 | 2.9 | 1.3 | 2.7 | 1.1 | 4.8 | 4.2 | 1.5 | 3.2 | 1.7 |
| it | 0.5 | 0.4 | 0.4 | 1.8 | 1.4 | 0.2 | 1.0 | 0.6 | 1.6 | 0.3 | 1.8 | 1.5 | 0.6 | 1.5 | 0.9 |
| pt | 0.6 | 0.5 | 0.4 | 0.7 | 4.0 | 0.2 | 1.3 | 0.6 | 1.1 | 0.3 | 2.3 | 2.2 | 0.5 | 1.5 | 0.8 |
| hi | 3.5 | 4.2 | 2.7 | 3.2 | 5.7 | 0.5 | 3.7 | 1.7 | 3.7 | 2.7 | 20.3 | 11.7 | 5.8 | 4.3 | 7.7 |
| es | 0.5 | 0.4 | 0.3 | 0.3 | 0.7 | 0.2 | 1.7 | 0.5 | 1.2 | 0.2 | 2.0 | 2.3 | 0.5 | 1.5 | 0.8 |
| th | 4.0 | 3.9 | 2.5 | 3.0 | 5.3 | 0.4 | 3.4 | 1.6 | 17.7 | 1.5 | 17.8 | 8.4 | 3.3 | 2.7 | 4.0 |
| bg | 4.2 | 3.3 | 2.4 | 3.5 | 5.0 | 0.5 | 3.7 | 1.6 | 5.3 | 1.5 | 17.3 | 8.7 | 3.5 | 3.2 | 5.2 |
| ru | 4.1 | 3.3 | 2.3 | 2.8 | 4.8 | 0.5 | 3.3 | 1.4 | 4.0 | 1.5 | 17.9 | 8.8 | 3.3 | 2.9 | 4.7 |
| tr | 0.3 | 1.2 | 0.7 | 0.7 | 1.6 | 0.3 | 0.5 | 1.0 | 0.8 | 0.6 | 11.0 | 3.7 | 3.0 | 2.9 | 3.2 |
| vi | 0.5 | 0.8 | 0.8 | 0.5 | 1.2 | 0.3 | 0.5 | 1.5 | 0.9 | 0.6 | 5.0 | 5.0 | 2.9 | 2.3 | 3.1 |
| ja | 5.3 | 4.7 | 2.7 | 4.0 | 5.9 | 0.5 | 3.5 | 1.6 | 3.3 | 2.0 | 18.9 | 9.6 | 7.8 | 4.3 | 7.1 |
| ko | 4.2 | 4.0 | 2.6 | 3.2 | 5.5 | 0.4 | 3.6 | 1.5 | 4.1 | 2.6 | 19.0 | 8.3 | 4.9 | 3.3 | 3.7 |
| zh | 5.1 | 4.3 | 2.8 | 3.3 | 5.8 | 0.3 | 4.6 | 1.3 | 4.5 | 1.7 | 18.0 | 10.7 | 1.8 | 2.6 | 3.0 |

| | en | de | fr | it | pt | hi | es | th | bg | ru | tr | vi | ja | ko | zh |
|---|---|---|---|---|---|---|---|---|---|---|---|---|---|---|---|
| en | 0.3 | 2.6 | 2.3 | 2.8 | 3.6 | 1.1 | 3.5 | 3.3 | 3.8 | 3.7 | 8.8 | 4.6 | 5.5 | 8.3 | 8.3 |
| de | 0.5 | -0.3 | 0.9 | 1.1 | 1.1 | 0.1 | 1.7 | 0.6 | 0.2 | 0.8 | 2.9 | 1.8 | 1.4 | 1.3 | 0.8 |
| fr | 0.7 | 1.6 | -0.1 | 2.7 | 4.5 | 0.6 | 5.1 | 2.1 | 1.9 | 2.2 | 7.8 | 6.2 | 2.7 | 4.0 | 3.2 |
| it | 0.4 | 0.7 | 0.8 | -0.2 | 0.7 | 0.2 | 1.7 | 0.6 | 0.2 | 1.2 | 3.0 | 1.3 | 1.5 | 1.4 | 1.4 |
| pt | 0.3 | 0.7 | 0.6 | 0.5 | -1.1 | 0.3 | 2.1 | 0.9 | 0.6 | 1.1 | 2.9 | 1.5 | 1.6 | 2.1 | 1.5 |
| hi | 4.4 | 4.7 | 3.6 | 4.9 | 7.3 | 0.3 | 6.8 | 2.7 | 5.5 | 3.6 | 15.3 | 8.5 | 5.9 | 4.5 | 13.1 |
| es | 0.2 | 0.5 | 0.4 | 0.4 | 0.2 | 0.2 | -0.4 | 0.6 | 0.3 | 0.6 | 1.6 | 0.4 | 1.2 | 1.5 | 1.2 |
| th | 4.4 | 3.3 | 3.4 | 4.3 | 5.9 | 0.2 | 6.0 | 0.3 | 1.9 | 1.7 | 12.6 | 2.6 | 1.8 | 1.4 | 5.3 |
| bg | 3.9 | 3.4 | 3.6 | 3.4 | 5.6 | 0.3 | 6.7 | 0.9 | -0.6 | 1.1 | 13.4 | 4.3 | 3.8 | 2.7 | 7.8 |
| ru | 3.9 | 3.2 | 2.9 | 3.8 | 5.2 | 0.2 | 5.1 | 0.8 | 1.3 | 0.4 | 12.1 | 2.6 | 3.7 | 2.2 | 6.9 |
| tr | 1.1 | 1.2 | 0.9 | 1.1 | 1.5 | 0.2 | 2.0 | 1.0 | 1.4 | 1.4 | -2.9 | 1.5 | 3.4 | 1.8 | 7.7 |
| vi | 0.8 | 0.7 | 0.5 | 0.9 | 1.1 | 0.2 | 1.4 | 1.7 | 1.5 | 1.2 | 0.7 | -3.3 | 2.7 | 1.0 | 10.0 |
| ja | 5.2 | 4.4 | 4.7 | 5.2 | 8.0 | 0.9 | 8.4 | 4.9 | 3.8 | 3.5 | 15.1 | 13.2 | -1.6 | 6.0 | 9.9 |
| ko | 4.4 | 3.9 | 3.4 | 4.6 | 6.7 | 0.4 | 6.3 | 1.3 | 3.8 | 2.5 | 13.8 | 5.8 | 2.6 | 0.7 | 10.4 |
| zh | 5.2 | 4.2 | 4.3 | 5.6 | 8.0 | 0.3 | 8.0 | 1.6 | 2.3 | 1.3 | 14.1 | 7.9 | 1.4 | 1.8 | 1.2 |

Figure 26: PPL changes in Llama 3.2 1B. **Top**: Steering using **language-specific features** with $\alpha = -0.1$ (left) and $\alpha = 0.1$ (right). **Bottom**: Steering using **language-specific neurons** with $\alpha = -0.1$ (left) and $\alpha = 0.1$ (right).

| | en | de | fr | it | pt | hi | es | th | bg | ru | tr | vi | ja | ko | zh |
|---|---|---|---|---|---|---|---|---|---|---|---|---|---|---|---|
| en | 4.8 | 2.6 | 1.7 | 1.9 | 3.1 | 0.1 | 2.8 | 0.2 | 0.9 | 1.1 | 6.8 | 7.6 | 0.2 | 1.6 | 1.1 |
| de | 7.3 | $10^3$ | 4.2 | 5.9 | 7.9 | 2.1 | 6.6 | 3.5 | 8.4 | 5.4 | 26.2 | 13.8 | 8.3 | 9.8 | 9.5 |
| fr | 9.3 | 20.4 | $10^3$ | 25.3 | 35.9 | 5.9 | 19.5 | 14.5 | 22.4 | 14.6 | 66.8 | 38.3 | 21.1 | 29.3 | 23.8 |
| it | 10.8 | 10.0 | 7.0 | $10^4$ | 17.2 | 3.9 | 7.5 | 7.9 | 44.8 | 9.3 | 123.7 | 20.0 | 368.3 | 17.2 | 18.1 |
| pt | 78.3 | 84.1 | 115.6 | 147.0 | $10^4$ | 48.4 | 142.5 | 378.5 | 214.2 | 93.4 | 364.1 | 776.0 | 142.4 | 303.9 | 173.2 |
| hi | 161.5 | 795.4 | 361.5 | 467.8 | 527.8 | $10^8$ | 432.1 | $10^5$ | $10^5$ | $10^4$ | $10^3$ | $10^3$ | $10^3$ | $10^4$ | $10^3$ |
| es | 2.8 | 6.3 | 3.6 | 5.5 | 315.4 | 3.0 | $10^3$ | 6.2 | 5.8 | 60.2 | 56.6 | 10.4 | 7.4 | 14.5 | 16.2 |
| th | 39.0 | 46.6 | 26.4 | 32.5 | 41.9 | 87.1 | 34.8 | $10^7$ | 457.4 | 51.2 | 236.2 | 405.2 | 229.2 | 469.7 | 161.6 |
| bg | 13.6 | 23.3 | 13.6 | 56.0 | 23.0 | 8.8 | 19.5 | 21.2 | 744.7 | 30.3 | 101.1 | 59.7 | 30.6 | 59.0 | 43.9 |
| ru | 7.2 | 8.0 | 5.1 | 6.7 | 9.0 | 6.4 | 7.4 | 9.3 | $10^3$ | 31.2 | 31.3 | 18.9 | 21.1 | 43.4 | 19.9 |
| tr | 42.4 | 86.8 | 43.1 | 56.3 | 98.5 | 60.5 | 72.3 | 263.8 | 120.1 | 89.9 | $10^6$ | 316.2 | 182.2 | $10^3$ | 200.4 |
| vi | 11.3 | 14.2 | 10.1 | 13.8 | 21.6 | 7.9 | 13.4 | 79.1 | 26.4 | 13.4 | 75.7 | $10^5$ | 22.2 | 67.6 | 38.3 |
| ja | 13.4 | 13.2 | 8.8 | 11.9 | 13.0 | 6.5 | 10.4 | 34.6 | 21.9 | 11.3 | 57.8 | 30.1 | $10^3$ | 52.0 | 77.4 |
| ko | 13.6 | 9.5 | 6.8 | 7.3 | 25.7 | 6.2 | 12.3 | 38.1 | 18.4 | 14.2 | 50.6 | 34.2 | 31.1 | $10^6$ | 25.6 |
| zh | 8.5 | 7.7 | 4.2 | 6.0 | 7.9 | 4.7 | 6.5 | 51.9 | 73.3 | 8.4 | 27.9 | 26.7 | 218.5 | 39.2 | 413.5 |

| | en | de | fr | it | pt | hi | es | th | bg | ru | tr | vi | ja | ko | zh |
|---|---|---|---|---|---|---|---|---|---|---|---|---|---|---|---|
| en | 2.2 | 2.6 | 1.7 | 1.6 | 2.4 | 1.0 | 2.0 | 3.2 | 2.3 | 2.5 | 13.4 | 11.6 | 3.7 | 6.1 | 4.6 |
| de | 7.1 | 1.2 | 67.5 | 107.7 | 63.5 | 17.6 | 40.6 | 22.4 | 68.4 | 60.4 | 144.8 | 48.2 | 18.8 | 30.3 | 15.4 |
| fr | 11.0 | 577.4 | 1.5 | $10^3$ | $10^3$ | $10^3$ | $10^3$ | $10^3$ | $10^4$ | $10^3$ | $10^3$ | $10^4$ | 90.6 | 173.3 | 76.1 |
| it | 12.6 | $10^4$ | $10^3$ | 3.0 | $10^3$ | $10^3$ | $10^3$ | $10^4$ | $10^4$ | $10^5$ | $10^5$ | $10^5$ | 102.7 | $10^3$ | 39.7 |
| pt | 118.4 | $10^4$ | $10^3$ | $10^3$ | 29.9 | $10^4$ | $10^3$ | $10^4$ | $10^5$ | $10^5$ | $10^5$ | $10^5$ | $10^3$ | $10^4$ | 472.3 |
| hi | $10^3$ | $10^5$ | $10^5$ | $10^5$ | $10^5$ | 104.1 | $10^5$ | $10^6$ | $10^6$ | $10^5$ | $10^6$ | $10^5$ | $10^5$ | $10^5$ | $10^5$ |
| es | 9.0 | $10^4$ | $10^3$ | $10^3$ | 264.1 | $10^4$ | 0.6 | $10^3$ | $10^4$ | $10^4$ | $10^4$ | $10^5$ | 651.3 | $10^3$ | 43.1 |
| th | 39.9 | 492.6 | 341.6 | 798.8 | $10^3$ | $10^3$ | 307.9 | 21.6 | $10^4$ | $10^4$ | $10^4$ | $10^4$ | $10^3$ | $10^4$ | $10^3$ |
| bg | 21.0 | $10^3$ | $10^3$ | 223.1 | $10^3$ | $10^4$ | $10^3$ | $10^5$ | 11.1 | 480.9 | $10^5$ | $10^4$ | $10^4$ | $10^4$ | 204.6 |
| ru | 7.2 | 76.4 | 22.7 | 116.7 | 118.4 | 449.2 | 50.1 | 285.4 | 73.2 | 5.3 | $10^3$ | 164.5 | 139.2 | 260.5 | 39.2 |
| tr | 69.7 | $10^3$ | $10^3$ | $10^3$ | $10^3$ | $10^5$ | $10^3$ | $10^5$ | $10^5$ | $10^4$ | 120.1 | $10^4$ | $10^5$ | $10^4$ | $10^4$ |
| vi | 15.2 | 140.2 | 51.8 | 169.5 | 310.6 | $10^3$ | 131.8 | $10^4$ | $10^3$ | 279.5 | $10^3$ | 11.8 | $10^3$ | $10^3$ | 281.1 |
| ja | 21.9 | 22.1 | 15.4 | 15.8 | 37.2 | $10^3$ | 21.1 | $10^3$ | 164.6 | 56.1 | 142.0 | 147.3 | 20.2 | $10^4$ | 664.1 |
| ko | 13.7 | 29.6 | 10.4 | 36.4 | 22.4 | $10^3$ | 14.9 | $10^4$ | 271.6 | 58.8 | $10^4$ | $10^3$ | $10^4$ | 18.5 | $10^3$ |
| zh | 13.7 | 49.5 | 22.5 | 43.3 | 59.8 | 642.0 | 41.8 | $10^3$ | 678.6 | 255.6 | $10^3$ | $10^3$ | 236.4 | $10^3$ | 10.5 |

| | en | de | fr | it | pt | hi | es | th | bg | ru | tr | vi | ja | ko | zh |
|---|---|---|---|---|---|---|---|---|---|---|---|---|---|---|---|
| en | 2.3 | 7.9 | 3.7 | 5.2 | 7.4 | 19.4 | 4.8 | 21.4 | 18.3 | 9.1 | $10^3$ | 76.9 | 306.4 | 48.9 | 32.3 |
| de | 4.2 | 153.2 | 3.6 | 12.5 | 13.6 | 1.8 | 7.5 | 6.5 | 7.5 | 4.7 | 25.5 | 11.8 | 151.4 | 17.3 | 15.0 |
| fr | 9.2 | 17.5 | $10^3$ | 85.0 | 157.4 | 7.3 | 95.9 | 29.7 | 44.0 | 20.7 | 103.9 | 60.8 | 50.3 | 50.2 | 43.9 |
| it | 5.1 | 4.6 | 5.1 | 868.5 | 16.6 | 2.2 | 11.8 | 16.1 | 14.3 | 4.9 | 28.5 | 26.0 | 66.4 | 22.3 | 28.1 |
| pt | 5.1 | 5.5 | 4.5 | 6.7 | 203.2 | 2.4 | 18.1 | 9.8 | 11.1 | 4.6 | 29.9 | 22.1 | 55.0 | 19.6 | 17.8 |
| hi | 69.2 | 191.3 | 105.8 | 146.0 | 193.1 | $10^3$ | 128.2 | $10^3$ | $10^4$ | 221.8 | $10^3$ | $10^3$ | 90.8 | 96.1 | 115.4 |
| es | 4.5 | 4.2 | 3.3 | 3.4 | 6.4 | 1.5 | 13.9 | 7.9 | 10.9 | 3.2 | 29.0 | 19.3 | 67.2 | 21.5 | 16.4 |
| th | 69.2 | 152.0 | 87.3 | 104.0 | 157.9 | $10^4$ | 95.1 | $10^3$ | $10^6$ | $10^5$ | 858.6 | $10^3$ | 140.0 | 264.7 | 86.2 |
| bg | 61.4 | 115.7 | 85.3 | 258.2 | 147.1 | 96.5 | 103.4 | 659.8 | $10^6$ | $10^3$ | 719.9 | $10^3$ | 60.7 | 79.3 | 101.0 |
| ru | 61.8 | 115.4 | 68.6 | 105.9 | 132.8 | 130.5 | 91.7 | 120.8 | $10^6$ | $10^5$ | 736.9 | 496.4 | 54.7 | 51.2 | 66.8 |
| tr | 5.2 | 17.2 | 8.4 | 13.4 | 21.1 | 3.4 | 9.2 | 20.3 | 10.3 | 9.2 | 499.1 | 44.1 | 35.6 | 35.0 | 45.0 |
| vi | 25.3 | 22.9 | 16.7 | 14.2 | 25.5 | 4.5 | 15.8 | 256.0 | 342.7 | 10.1 | 103.1 | 80.5 | 23.2 | 41.2 | 66.0 |
| ja | 98.7 | 288.0 | 110.4 | 463.7 | 254.5 | 398.4 | 126.9 | $10^4$ | $10^4$ | 869.3 | $10^3$ | $10^3$ | $10^4$ | $10^3$ | 666.3 |
| ko | 76.8 | 194.7 | 106.8 | 168.9 | 220.2 | $10^3$ | 133.9 | $10^3$ | $10^5$ | $10^3$ | $10^3$ | $10^3$ | 327.6 | $10^4$ | 77.6 |
| zh | 94.9 | 201.2 | 113.5 | 150.4 | 217.8 | 465.5 | 153.1 | 974.1 | $10^5$ | $10^3$ | $10^3$ | $10^3$ | 337.1 | 142.5 | $10^3$ |

| | en | de | fr | it | pt | hi | es | th | bg | ru | tr | vi | ja | ko | zh |
|---|---|---|---|---|---|---|---|---|---|---|---|---|---|---|---|
| en | 2.5 | $10^3$ | $10^3$ | $10^3$ | $10^3$ | $10^5$ | $10^3$ | $10^5$ | $10^4$ | $10^5$ | $10^3$ | $10^5$ | $10^5$ | $10^5$ | $10^5$ |
| de | 5.1 | 3.0 | 41.8 | 45.0 | 17.5 | 4.2 | 25.1 | 12.0 | 4.7 | 13.9 | 45.0 | 98.3 | 47.5 | 53.5 | 29.5 |
| fr | 7.9 | 24.1 | 2.8 | 102.8 | 167.1 | 328.5 | 205.7 | $10^3$ | 743.0 | $10^3$ | 130.6 | $10^4$ | $10^3$ | $10^4$ | 558.1 |
| it | 4.9 | 7.4 | 20.3 | 2.8 | 10.1 | 11.2 | 22.6 | 13.8 | 70.0 | 92.6 | 33.0 | 53.3 | 378.2 | 70.2 | 174.4 |
| pt | 4.8 | 7.9 | 6.9 | 8.4 | 1.8 | 24.7 | 62.8 | 821.7 | 96.4 | 105.6 | 43.3 | $10^3$ | 143.8 | $10^3$ | 134.2 |
| hi | 76.5 | $10^3$ | $10^3$ | $10^3$ | $10^3$ | 3.2 | $10^3$ | $10^3$ | $10^3$ | 292.9 | $10^3$ | $10^3$ | 266.3 | 891.4 | 481.4 |
| es | 3.5 | 5.3 | 3.9 | 4.0 | 4.3 | 4.8 | 1.3 | 28.0 | 28.2 | 44.0 | 22.4 | 52.4 | 59.7 | 623.9 | 88.4 |
| th | 65.9 | 145.8 | 296.9 | 398.6 | 628.8 | 3.5 | 909.8 | 5.9 | 38.1 | 46.5 | 931.9 | 124.1 | 25.2 | 22.3 | 68.5 |
| bg | 83.5 | $10^3$ | $10^4$ | 935.6 | $10^3$ | 115.0 | $10^4$ | 492.7 | 5.1 | 18.3 | $10^4$ | $10^4$ | 270.6 | 732.4 | $10^3$ |
| ru | 84.7 | $10^3$ | $10^3$ | $10^3$ | $10^3$ | 25.8 | $10^3$ | 565.0 | 11.0 | 5.5 | $10^3$ | $10^3$ | 662.2 | 850.0 | 989.8 |
| tr | 9.6 | 23.8 | 79.5 | 53.8 | 80.3 | 5.0 | 107.3 | 60.3 | 31.6 | 33.0 | 8.6 | 435.8 | 172.7 | 308.1 | 162.4 |
| vi | 6.7 | 9.8 | 12.2 | 29.2 | 42.7 | 5.5 | 38.3 | 96.0 | 25.4 | 21.6 | 28.5 | -0.2 | 45.7 | 24.8 | 242.3 |
| ja | 238.3 | $10^3$ | $10^4$ | $10^3$ | $10^3$ | 133.2 | $10^4$ | $10^3$ | $10^3$ | 989.1 | $10^3$ | $10^4$ | 5.1 | $10^3$ | 277.8 |
| ko | 75.0 | 804.3 | $10^3$ | $10^3$ | $10^3$ | 19.8 | $10^3$ | 308.0 | 180.2 | 93.7 | $10^3$ | $10^3$ | 83.3 | 13.3 | 201.4 |
| zh | 374.7 | $10^3$ | $10^3$ | $10^3$ | $10^3$ | 6.8 | $10^4$ | $10^3$ | 329.1 | 209.2 | $10^3$ | $10^4$ | 26.4 | 90.0 | 17.4 |

Figure 27: PPL changes in Llama 3.2 1B. **Top**: Steering using **language-specific features** with $\alpha = -0.3$ (left) and $\alpha = 0.3$ (right). **Bottom**: Steering using **language-specific neurons** with $\alpha = -0.3$ (left) and $\alpha = 0.3$ (right).

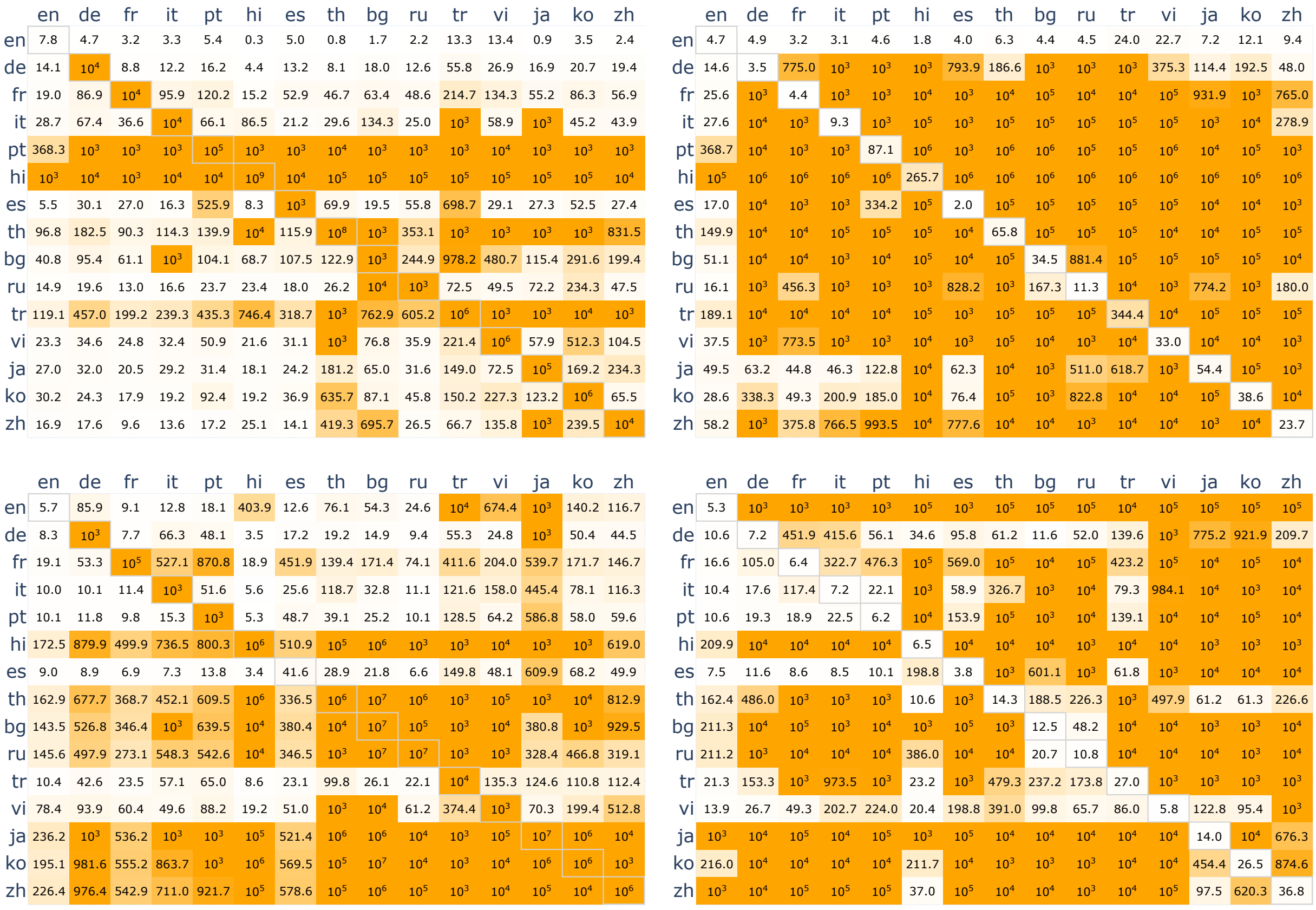

Figure 28: PPL changes in Llama 3.2 1B. **Top**: Steering using **language-specific features** with $\alpha = -0.4$ (left) and $\alpha = 0.4$ (right). **Bottom**: Steering using **language-specific neurons** with $\alpha = -0.4$ (left) and $\alpha = 0.4$ (right).

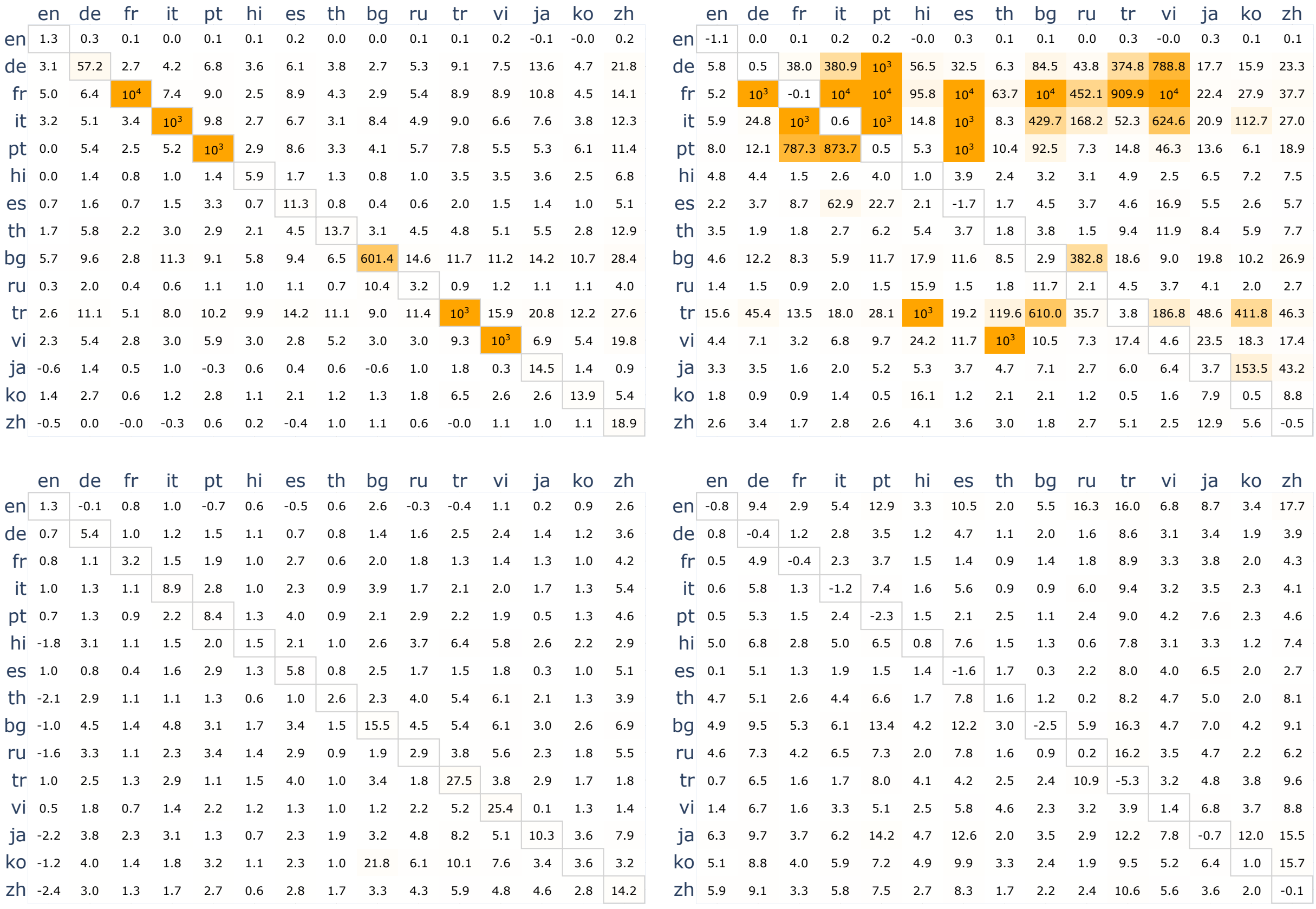

Figure 29: PPL changes in Gemma 2 2B. **Top**: Steering using **language-specific features** with $\alpha = -0.1$ (left) and $\alpha = 0.1$ (right). **Bottom**: Steering using **language-specific neurons** with $\alpha = -0.1$ (left) and $\alpha = 0.1$ (right).

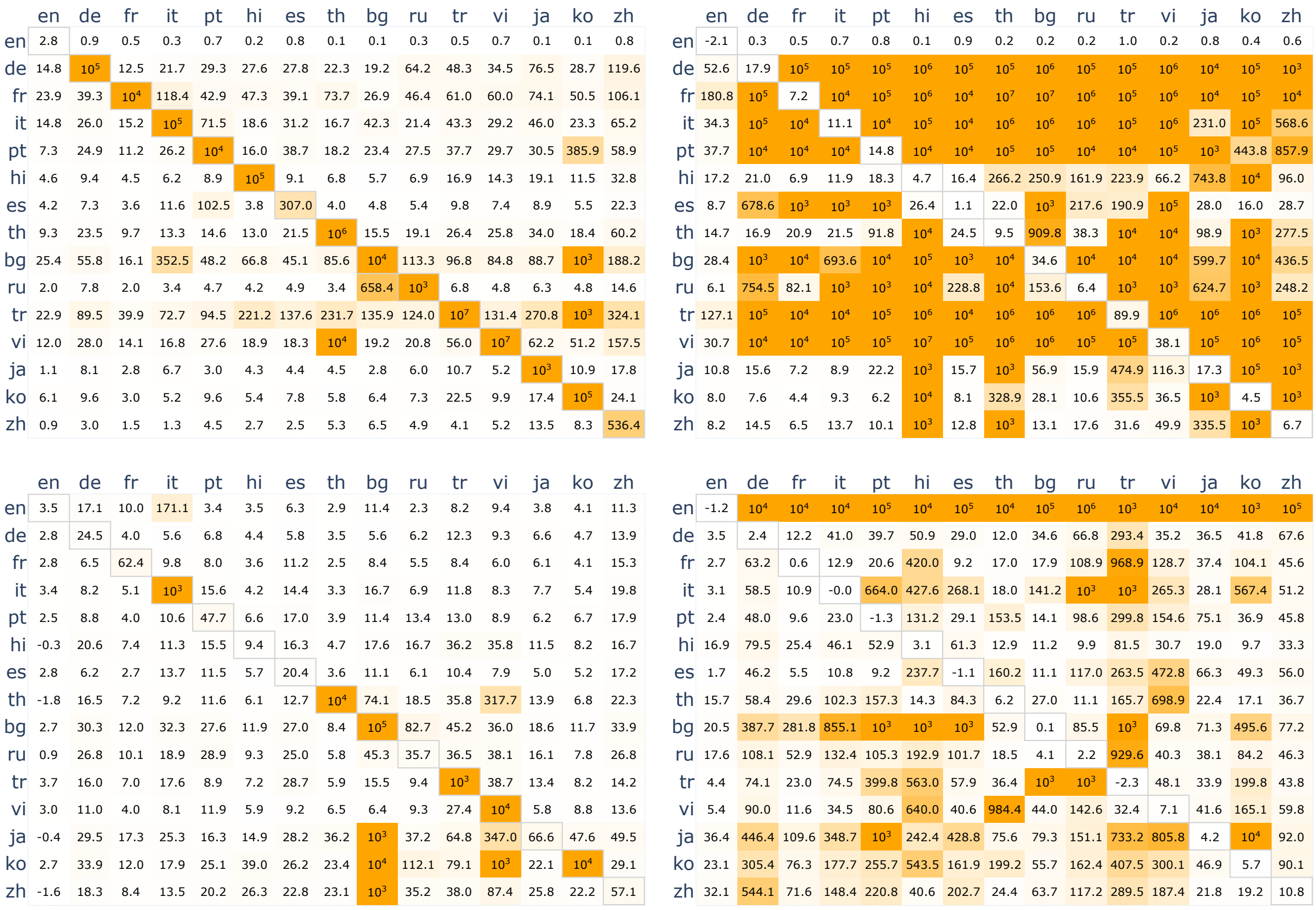

Figure 30: PPL changes in Gemma 2 2B. **Top**: Steering using **language-specific features** with $\alpha = -0.2$ (left) and $\alpha = 0.2$ (right). **Bottom**: Steering using **language-specific neurons** with $\alpha = -0.2$ (left) and $\alpha = 0.2$ (right).

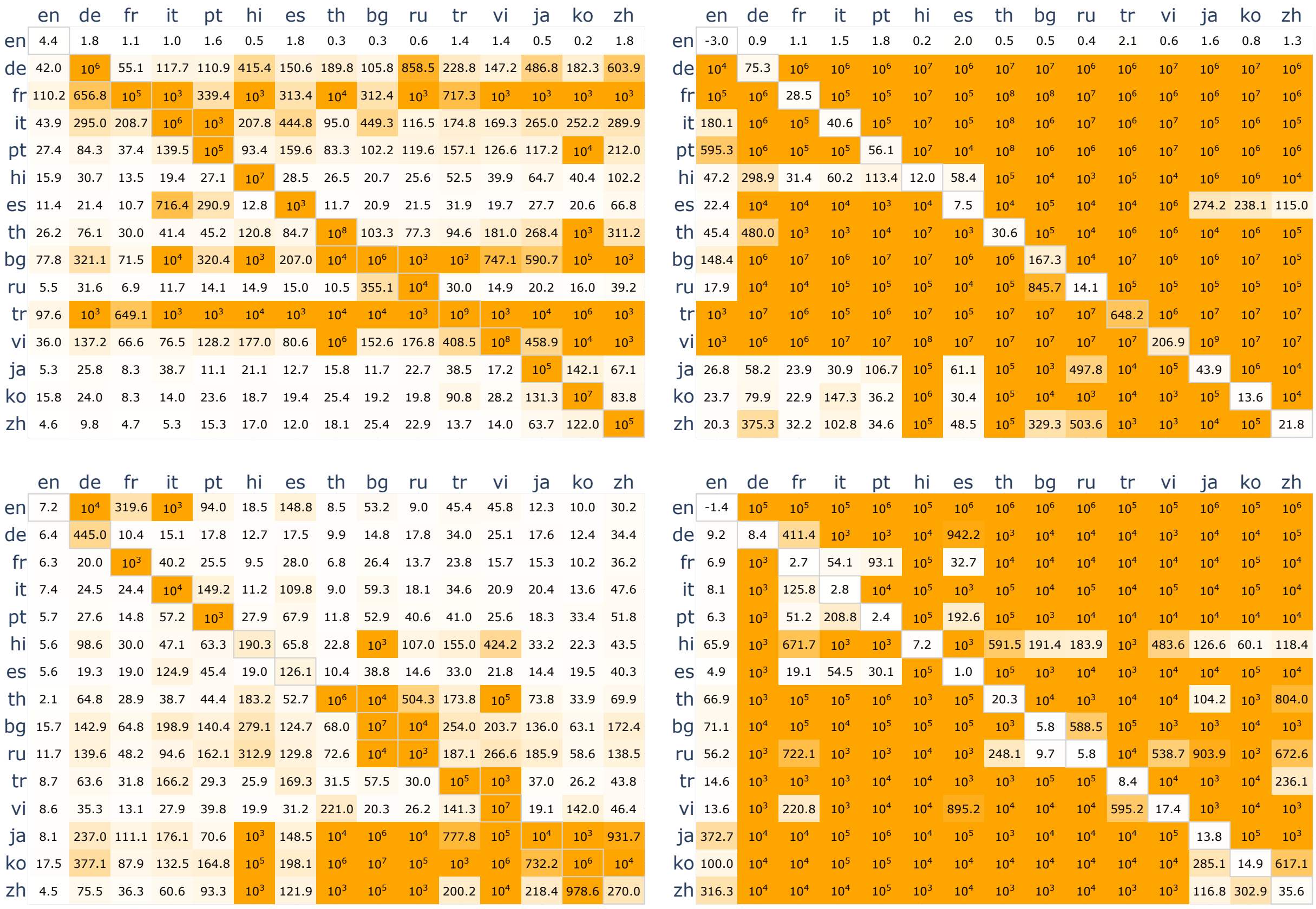

Figure 31: PPL changes in Gemma 2 2B. **Top**: Steering using **language-specific features** with $\alpha = -0.3$ (left) and $\alpha = 0.3$ (right). **Bottom**: Steering using **language-specific neurons** with $\alpha = -0.3$ (left) and $\alpha = 0.3$ (right).

| | en | de | fr | it | pt | hi | es | th | bg | ru | tr | vi | ja | ko | zh |
|---|---|---|---|---|---|---|---|---|---|---|---|---|---|---|---|
| en | 6.3 | 3.0 | 2.1 | 1.9 | 2.9 | 0.8 | 3.4 | 0.6 | 0.6 | 1.0 | 2.7 | 2.3 | 1.1 | 0.6 | 3.2 |
| de | 113.8 | $10^{8}$ | 642.8 | $10^{3}$ | 922.0 | $10^{4}$ | $10^{3}$ | $10^{4}$ | $10^{3}$ | $10^{3}$ | $10^{3}$ | $10^{3}$ | $10^{3}$ | $10^{3}$ | $10^{3}$ |
| fr | 903.2 | $10^{4}$ | $10^{7}$ | $10^{5}$ | $10^{4}$ | $10^{6}$ | $10^{4}$ | $10^{7}$ | $10^{3}$ | $10^{4}$ | $10^{4}$ | $10^{6}$ | $10^{5}$ | $10^{5}$ | $10^{5}$ |
| it | 128.0 | $10^{3}$ | $10^{3}$ | $10^{7}$ | $10^{4}$ | $10^{3}$ | $10^{3}$ | $10^{3}$ | $10^{3}$ | $10^{3}$ | $10^{3}$ | $10^{3}$ | $10^{4}$ | $10^{3}$ | $10^{3}$ |
| pt | 95.9 | 346.0 | 159.7 | $10^{3}$ | $10^{6}$ | $10^{3}$ | $10^{3}$ | 464.6 | 760.9 | 908.1 | $10^{3}$ | 584.1 | 579.4 | $10^{5}$ | 911.1 |
| hi | 39.1 | 87.8 | 36.5 | 53.8 | 72.1 | $10^{8}$ | 80.9 | 165.7 | 79.6 | 104.9 | 157.7 | 109.0 | 217.1 | 214.0 | 301.5 |
| es | 25.0 | 81.3 | 35.2 | $10^{3}$ | 392.6 | 82.7 | $10^{4}$ | 37.7 | 92.9 | 77.8 | 120.1 | 49.1 | 89.9 | 88.8 | 215.6 |
| th | 61.0 | 303.6 | 97.5 | 138.1 | 139.5 | $10^{3}$ | 402.9 | $10^{10}$ | $10^{3}$ | 829.2 | 373.6 | $10^{3}$ | $10^{4}$ | $10^{4}$ | $10^{3}$ |
| bg | 255.8 | $10^{3}$ | 633.8 | $10^{5}$ | $10^{4}$ | $10^{6}$ | $10^{3}$ | $10^{6}$ | $10^{8}$ | $10^{5}$ | $10^{5}$ | $10^{4}$ | $10^{3}$ | $10^{7}$ | $10^{4}$ |
| ru | 12.1 | 242.4 | 26.6 | 35.0 | 44.1 | 88.2 | 39.1 | 31.0 | 751.0 | $10^{4}$ | 105.1 | 45.5 | 82.0 | 136.4 | 136.9 |
| tr | 415.3 | $10^{4}$ | $10^{4}$ | $10^{5}$ | $10^{5}$ | $10^{6}$ | $10^{5}$ | $10^{7}$ | $10^{6}$ | $10^{5}$ | $10^{10}$ | $10^{5}$ | $10^{6}$ | $10^{8}$ | $10^{5}$ |
| vi | 105.9 | $10^{3}$ | 850.3 | 737.3 | $10^{3}$ | $10^{4}$ | 636.7 | $10^{8}$ | $10^{3}$ | $10^{3}$ | $10^{4}$ | $10^{10}$ | $10^{3}$ | $10^{7}$ | $10^{4}$ |
| ja | 12.7 | 86.7 | 24.7 | 298.7 | 27.7 | 269.3 | 28.3 | 70.4 | 33.7 | 112.0 | 194.4 | 44.8 | $10^{6}$ | $10^{3}$ | 300.1 |
| ko | 33.8 | 57.0 | 19.7 | 35.5 | 58.5 | 101.9 | 43.7 | 752.7 | 65.7 | 54.8 | $10^{3}$ | 175.3 | 382.4 | $10^{8}$ | $10^{3}$ |
| zh | 11.1 | 23.1 | 11.1 | 13.5 | 95.4 | 358.4 | 48.2 | 94.4 | 127.8 | 110.8 | 35.0 | 36.8 | 752.3 | $10^{3}$ | $10^{6}$ |

| | en | de | fr | it | pt | hi | es | th | bg | ru | tr | vi | ja | ko | zh |
|---|---|---|---|---|---|---|---|---|---|---|---|---|---|---|---|
| en | -3.7 | 1.8 | 1.9 | 2.5 | 3.1 | 0.4 | 3.6 | 0.8 | 0.9 | 0.8 | 3.7 | 1.2 | 2.6 | 1.3 | 2.5 |
| de | $10^{5}$ | 278.5 | $10^{6}$ | $10^{6}$ | $10^{6}$ | $10^{7}$ | $10^{6}$ | $10^{7}$ | $10^{7}$ | $10^{7}$ | $10^{6}$ | $10^{7}$ | $10^{7}$ | $10^{8}$ | $10^{7}$ |
| fr | $10^{6}$ | $10^{6}$ | 105.7 | $10^{5}$ | $10^{5}$ | $10^{7}$ | $10^{5}$ | $10^{8}$ | $10^{8}$ | $10^{7}$ | $10^{6}$ | $10^{7}$ | $10^{6}$ | $10^{7}$ | $10^{7}$ |
| it | $10^{4}$ | $10^{6}$ | $10^{5}$ | 125.8 | $10^{5}$ | $10^{7}$ | $10^{5}$ | $10^{8}$ | $10^{7}$ | $10^{7}$ | $10^{6}$ | $10^{8}$ | $10^{6}$ | $10^{7}$ | $10^{6}$ |
| pt | $10^{5}$ | $10^{6}$ | $10^{5}$ | $10^{5}$ | 184.2 | $10^{7}$ | $10^{5}$ | $10^{8}$ | $10^{7}$ | $10^{6}$ | $10^{6}$ | $10^{7}$ | $10^{7}$ | $10^{7}$ | $10^{7}$ |
| hi | 123.1 | $10^{4}$ | 283.8 | $10^{3}$ | $10^{3}$ | 25.1 | 469.5 | $10^{6}$ | $10^{6}$ | $10^{5}$ | $10^{6}$ | $10^{5}$ | $10^{7}$ | $10^{6}$ | $10^{7}$ |
| es | 49.8 | $10^{5}$ | $10^{5}$ | $10^{4}$ | $10^{4}$ | $10^{6}$ | 18.8 | $10^{6}$ | $10^{6}$ | $10^{5}$ | $10^{5}$ | $10^{7}$ | $10^{4}$ | $10^{4}$ | $10^{3}$ |
| th | 171.0 | $10^{5}$ | $10^{6}$ | $10^{6}$ | $10^{7}$ | $10^{8}$ | $10^{6}$ | 90.0 | $10^{6}$ | $10^{6}$ | $10^{7}$ | $10^{7}$ | $10^{6}$ | $10^{7}$ | $10^{6}$ |
| bg | $10^{3}$ | $10^{7}$ | $10^{8}$ | $10^{7}$ | $10^{7}$ | $10^{7}$ | $10^{7}$ | $10^{7}$ | 804.5 | $10^{5}$ | $10^{8}$ | $10^{7}$ | $10^{7}$ | $10^{7}$ | $10^{7}$ |
| ru | 52.3 | $10^{5}$ | $10^{5}$ | $10^{5}$ | $10^{5}$ | $10^{5}$ | $10^{5}$ | $10^{5}$ | $10^{3}$ | 30.2 | $10^{6}$ | $10^{5}$ | $10^{6}$ | $10^{6}$ | $10^{5}$ |
| tr | $10^{5}$ | $10^{7}$ | $10^{6}$ | $10^{6}$ | $10^{6}$ | $10^{7}$ | $10^{6}$ | $10^{7}$ | $10^{7}$ | $10^{7}$ | $10^{3}$ | $10^{7}$ | $10^{8}$ | $10^{7}$ | $10^{7}$ |
| vi | $10^{5}$ | $10^{7}$ | $10^{7}$ | $10^{7}$ | $10^{7}$ | $10^{8}$ | $10^{7}$ | $10^{8}$ | $10^{8}$ | $10^{7}$ | $10^{8}$ | 950.7 | $10^{10}$ | $10^{8}$ | $10^{8}$ |
| ja | 63.4 | 499.8 | 121.4 | 384.8 | $10^{3}$ | $10^{6}$ | $10^{3}$ | $10^{6}$ | $10^{5}$ | $10^{4}$ | $10^{5}$ | $10^{7}$ | 93.4 | $10^{6}$ | $10^{5}$ |
| ko | 57.9 | $10^{4}$ | $10^{3}$ | $10^{4}$ | $10^{3}$ | $10^{6}$ | 347.9 | $10^{6}$ | $10^{5}$ | $10^{5}$ | $10^{5}$ | $10^{5}$ | $10^{6}$ | 32.4 | $10^{5}$ |
| zh | 56.1 | $10^{4}$ | $10^{3}$ | $10^{3}$ | 865.3 | $10^{5}$ | $10^{3}$ | $10^{6}$ | $10^{4}$ | $10^{4}$ | $10^{5}$ | $10^{4}$ | $10^{5}$ | $10^{6}$ | 47.3 |

| | en | de | fr | it | pt | hi | es | th | bg | ru | tr | vi | ja | ko | zh |
|---|---|---|---|---|---|---|---|---|---|---|---|---|---|---|---|
| en | 18.5 | $10^{5}$ | $10^{5}$ | $10^{5}$ | $10^{4}$ | 121.2 | $10^{4}$ | 23.7 | 254.9 | 21.9 | 239.7 | 250.1 | 31.0 | 21.4 | 81.2 |
| de | 12.3 | $10^{3}$ | 25.1 | 37.2 | 40.6 | 42.6 | 42.4 | 29.7 | 34.9 | 47.8 | 89.0 | 68.7 | 40.0 | 34.8 | 76.3 |
| fr | 11.8 | 57.6 | $10^{4}$ | 271.1 | 208.1 | 26.0 | 104.7 | 18.6 | 79.3 | 35.6 | 56.9 | 39.1 | 31.9 | 21.4 | 75.2 |
| it | 13.5 | 60.8 | 325.5 | $10^{5}$ | $10^{3}$ | 29.8 | $10^{3}$ | 26.2 | 221.8 | 42.8 | 92.3 | 48.0 | 53.8 | 28.5 | 103.5 |
| pt | 11.8 | 75.8 | 136.6 | 674.8 | $10^{4}$ | 143.7 | 764.0 | 39.5 | 245.8 | 95.7 | 125.6 | 69.9 | 40.9 | 231.5 | 190.2 |
| hi | 18.7 | 725.5 | 116.8 | 190.0 | 249.4 | $10^{4}$ | 263.7 | $10^{3}$ | $10^{5}$ | $10^{4}$ | 845.9 | $10^{4}$ | 99.8 | 76.0 | 102.7 |
| es | 9.7 | 50.0 | 615.3 | $10^{3}$ | $10^{3}$ | 59.2 | $10^{4}$ | 30.3 | 132.9 | 31.2 | 95.8 | 56.5 | 30.9 | 67.1 | 91.6 |
| th | 11.8 | 270.4 | 110.9 | 151.6 | 149.2 | $10^{5}$ | 186.3 | $10^{7}$ | $10^{6}$ | $10^{4}$ | $10^{3}$ | $10^{6}$ | $10^{3}$ | $10^{3}$ | 225.5 |
| bg | 47.3 | 745.2 | 379.8 | $10^{3}$ | 751.6 | $10^{4}$ | 570.1 | $10^{3}$ | $10^{8}$ | $10^{6}$ | $10^{3}$ | $10^{3}$ | $10^{3}$ | 966.9 | $10^{3}$ |
| ru | 38.6 | 798.5 | 227.8 | 465.8 | 979.4 | $10^{4}$ | 752.9 | $10^{3}$ | $10^{6}$ | $10^{6}$ | 947.3 | $10^{3}$ | $10^{3}$ | $10^{3}$ | $10^{3}$ |
| tr | 17.3 | 274.1 | 236.8 | $10^{3}$ | 80.2 | 123.2 | 757.4 | 528.8 | 215.1 | 87.4 | $10^{6}$ | $10^{4}$ | 90.3 | 89.2 | 112.8 |
| vi | 19.1 | 104.0 | 38.9 | 96.8 | 134.1 | 66.3 | 98.2 | $10^{4}$ | 57.8 | 68.4 | $10^{3}$ | $10^{8}$ | 46.6 | $10^{3}$ | 135.3 |
| ja | 28.8 | $10^{3}$ | 830.3 | $10^{3}$ | 276.8 | $10^{6}$ | 614.0 | $10^{7}$ | $10^{7}$ | $10^{6}$ | $10^{4}$ | $10^{6}$ | $10^{7}$ | $10^{5}$ | $10^{6}$ |
| ko | 54.3 | $10^{4}$ | 844.7 | $10^{3}$ | $10^{3}$ | $10^{8}$ | $10^{3}$ | $10^{9}$ | $10^{9}$ | $10^{7}$ | $10^{4}$ | $10^{8}$ | $10^{7}$ | $10^{7}$ | $10^{8}$ |
| zh | 20.4 | 346.6 | 157.1 | 263.2 | 449.1 | $10^{5}$ | 836.7 | $10^{5}$ | $10^{7}$ | $10^{5}$ | $10^{3}$ | $10^{6}$ | $10^{3}$ | $10^{4}$ | $10^{4}$ |

| | en | de | fr | it | pt | hi | es | th | bg | ru | tr | vi | ja | ko | zh |
|---|---|---|---|---|---|---|---|---|---|---|---|---|---|---|---|
| en | -1.3 | $10^{5}$ | $10^{5}$ | $10^{5}$ | $10^{6}$ | $10^{5}$ | $10^{5}$ | $10^{6}$ | $10^{6}$ | $10^{6}$ | $10^{5}$ | $10^{5}$ | $10^{6}$ | $10^{5}$ | $10^{6}$ |
| de | 19.8 | 19.0 | $10^{4}$ | $10^{4}$ | $10^{4}$ | $10^{5}$ | $10^{4}$ | $10^{5}$ | $10^{5}$ | $10^{5}$ | $10^{5}$ | $10^{4}$ | $10^{6}$ | $10^{6}$ | $10^{6}$ |
| fr | 13.5 | $10^{4}$ | 6.2 | 216.5 | 432.7 | $10^{6}$ | 100.6 | $10^{6}$ | $10^{6}$ | $10^{5}$ | $10^{5}$ | $10^{5}$ | $10^{5}$ | $10^{6}$ | $10^{5}$ |
| it | 16.0 | $10^{4}$ | 936.5 | 7.3 | $10^{4}$ | $10^{6}$ | $10^{4}$ | $10^{6}$ | $10^{5}$ | $10^{6}$ | $10^{5}$ | $10^{6}$ | $10^{6}$ | $10^{6}$ | $10^{6}$ |
| pt | 12.9 | $10^{4}$ | 231.9 | $10^{3}$ | 9.1 | $10^{6}$ | 632.2 | $10^{6}$ | $10^{4}$ | $10^{5}$ | $10^{4}$ | $10^{5}$ | $10^{6}$ | $10^{6}$ | $10^{6}$ |
| hi | 257.9 | $10^{5}$ | $10^{4}$ | $10^{5}$ | $10^{5}$ | 13.6 | $10^{4}$ | $10^{4}$ | $10^{4}$ | $10^{3}$ | $10^{4}$ | $10^{3}$ | $10^{3}$ | 578.2 | 498.2 |
| es | 10.4 | $10^{4}$ | 72.1 | 333.6 | 79.0 | $10^{6}$ | 4.8 | $10^{6}$ | $10^{5}$ | $10^{5}$ | $10^{4}$ | $10^{5}$ | $10^{6}$ | $10^{6}$ | $10^{6}$ |
| th | 327.2 | $10^{5}$ | $10^{7}$ | $10^{7}$ | $10^{7}$ | $10^{5}$ | $10^{7}$ | 52.8 | $10^{5}$ | $10^{5}$ | $10^{6}$ | $10^{5}$ | $10^{3}$ | $10^{5}$ | $10^{4}$ |
| bg | 183.8 | $10^{5}$ | $10^{6}$ | $10^{6}$ | $10^{6}$ | $10^{6}$ | $10^{6}$ | $10^{4}$ | 16.2 | $10^{3}$ | $10^{6}$ | $10^{5}$ | $10^{5}$ | $10^{5}$ | $10^{4}$ |
| ru | 130.8 | $10^{4}$ | $10^{3}$ | $10^{4}$ | $10^{4}$ | $10^{4}$ | $10^{3}$ | $10^{3}$ | 19.3 | 11.6 | $10^{5}$ | $10^{3}$ | $10^{4}$ | $10^{4}$ | $10^{4}$ |
| tr | 38.3 | $10^{5}$ | $10^{4}$ | $10^{4}$ | $10^{5}$ | $10^{5}$ | $10^{4}$ | $10^{5}$ | $10^{6}$ | $10^{6}$ | 31.2 | $10^{5}$ | $10^{5}$ | $10^{5}$ | $10^{4}$ |
| vi | 26.5 | $10^{4}$ | $10^{4}$ | $10^{5}$ | $10^{5}$ | $10^{5}$ | $10^{5}$ | $10^{5}$ | $10^{5}$ | $10^{5}$ | $10^{4}$ | 34.4 | $10^{5}$ | $10^{5}$ | $10^{5}$ |
| ja | $10^{3}$ | $10^{5}$ | $10^{5}$ | $10^{6}$ | $10^{6}$ | $10^{5}$ | $10^{6}$ | $10^{5}$ | $10^{6}$ | $10^{5}$ | $10^{5}$ | $10^{7}$ | 29.1 | $10^{6}$ | $10^{4}$ |
| ko | 313.4 | $10^{6}$ | $10^{5}$ | $10^{6}$ | $10^{6}$ | $10^{5}$ | $10^{5}$ | $10^{5}$ | $10^{5}$ | $10^{5}$ | $10^{5}$ | $10^{5}$ | $10^{3}$ | 30.6 | $10^{3}$ |
| zh | $10^{3}$ | $10^{6}$ | $10^{5}$ | $10^{5}$ | $10^{6}$ | $10^{4}$ | $10^{6}$ | $10^{4}$ | $10^{4}$ | $10^{5}$ | $10^{4}$ | $10^{4}$ | 605.5 | $10^{3}$ | 79.7 |

Figure 32: PPL changes in Gemma 2 2B. **Top**: Steering using **language-specific features** with $\alpha = -0.4$ (left) and $\alpha = 0.4$ (right). **Bottom**: Steering using **language-specific neurons** with $\alpha = -0.4$ (left) and $\alpha = 0.4$ (right).

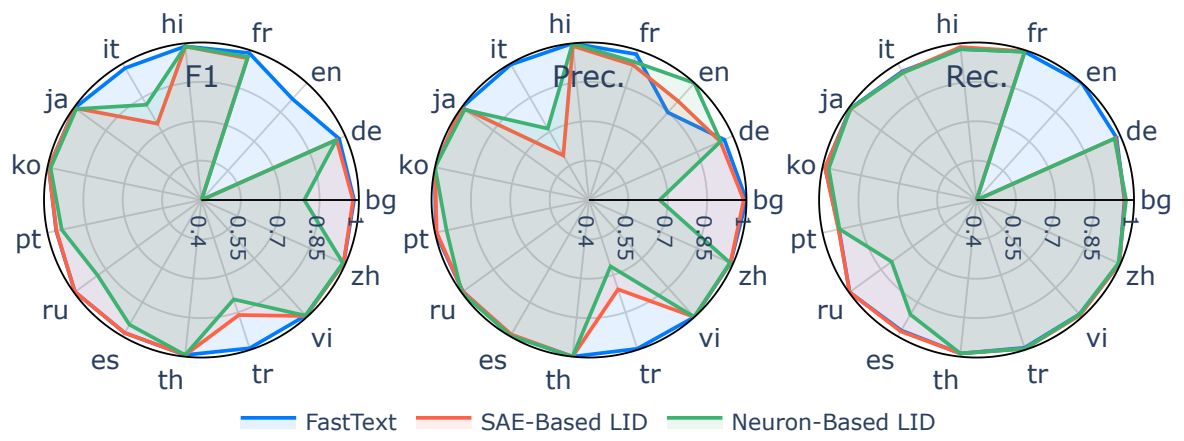

Figure 33: Results of SAE-based and neuron-based LID derived from Gemma 2 2B and fastText LID, applied to WiLI-2018 across 15 languages.

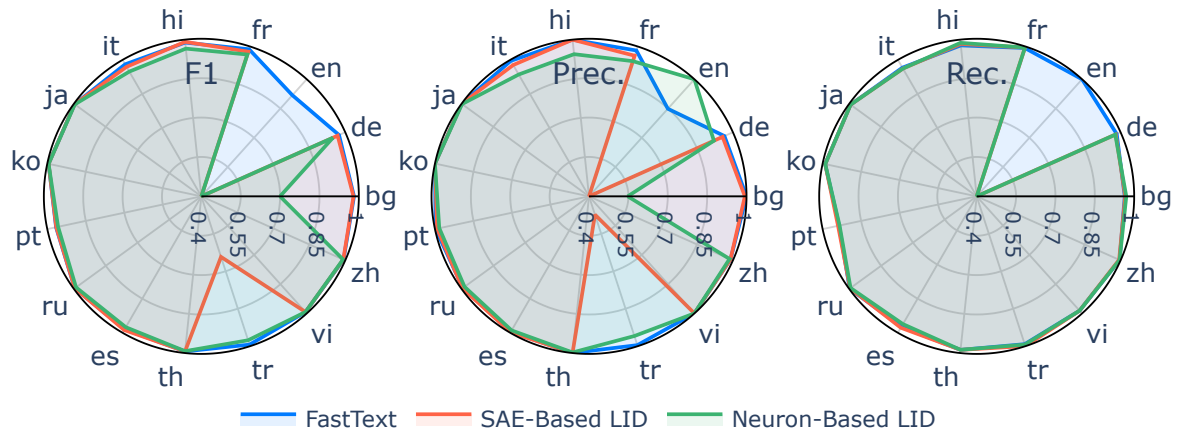

Figure 34: Results of SAE-based and neuron-based LID derived from Gemma 2 9B and fastText LID, applied to WiLI-2018 across 15 languages.

| Language | SAEs Classifier | | | fastText | | | Neurons Classifier | | |
|---|---|---|---|---|---|---|---|---|---|
| | **Prec.** | **Rec.** | **F1** | **Prec.** | **Rec.** | **F1** | **Prec.** | **Rec.** | **F1** |
| bg | 0.990 | 0.966 | 0.978 | **1.000** | 0.964 | **0.982** | 0.669 | **0.970** | 0.792 |
| de | 0.948 | 0.980 | 0.964 | **0.965** | **0.984** | **0.974** | 0.949 | 0.976 | 0.963 |
| en | 0.907 | 0.098 | 0.177 | 0.849 | **1.000** | **0.918** | **1.000** | 0.016 | 0.031 |
| fr | 0.941 | **0.996** | 0.968 | **0.984** | 0.994 | **0.989** | 0.954 | 0.994 | 0.974 |
| hi | 0.990 | **0.986** | 0.988 | **1.000** | 0.978 | **0.989** | **1.000** | 0.978 | **0.989** |
| it | 0.596 | 0.962 | 0.736 | **0.996** | **0.966** | **0.981** | 0.714 | 0.960 | 0.819 |
| ja | **1.000** | 0.992 | 0.996 | **1.000** | **0.994** | **0.997** | 0.986 | **0.994** | 0.990 |
| ko | **1.000** | **0.990** | **0.995** | **1.000** | **0.990** | **0.995** | **1.000** | 0.976 | 0.988 |
| pt | 0.994 | **0.938** | 0.965 | **0.998** | 0.936 | **0.966** | 0.955 | 0.934 | 0.944 |
| ru | 0.994 | **0.998** | **0.996** | 0.994 | **0.998** | **0.996** | **1.000** | 0.800 | 0.889 |
| es | 0.992 | **0.978** | **0.985** | 0.996 | 0.974 | **0.985** | **0.998** | 0.904 | 0.949 |
| th | **1.000** | **0.988** | **0.994** | **1.000** | **0.988** | **0.994** | **1.000** | **0.988** | **0.994** |
| tr | 0.758 | **0.996** | 0.861 | **0.996** | 0.992 | **0.994** | 0.666 | **0.996** | 0.798 |
| vi | **1.000** | **0.990** | **0.995** | **1.000** | 0.986 | 0.993 | **1.000** | 0.986 | 0.993 |
| zh | **0.990** | **0.996** | **0.993** | **0.990** | **0.996** | **0.993** | 0.988 | 0.992 | 0.990 |
| **Macro Avg** | 0.940 | 0.924 | 0.906 | **0.984** | **0.983** | **0.983** | 0.925 | 0.898 | 0.873 |
| **Accuracy** | 0.924 | | | **0.983** | | | 0.898 | | |

Table 5: Results of SAE-based and neuron-based LID derived from Gemma 2 2B and fastText LID, applied to WiLI-2018 across 15 languages. **Bold** text indicates the performance is equal to or greater than the counterpart.

| Language | SAEs Classifier | | | fastText | | | Neurons Classifier | | |
|---|---|---|---|---|---|---|---|---|---|
| | **Prec.** | **Rec.** | **F1** | **Prec.** | **Rec.** | **F1** | **Prec.** | **Rec.** | **F1** |
| bg | 0.994 | 0.966 | 0.980 | **1.000** | 0.964 | **0.982** | 0.547 | **0.970** | 0.700 |
| de | 0.957 | 0.980 | 0.968 | **0.965** | **0.984** | **0.974** | 0.921 | 0.978 | 0.949 |
| en | 0.000 | 0.000 | 0.000 | 0.849 | **1.000** | **0.918** | **1.000** | 0.096 | 0.175 |
| fr | 0.963 | **0.996** | 0.979 | **0.984** | 0.994 | **0.989** | 0.941 | **0.996** | 0.968 |
| hi | **1.000** | 0.984 | **0.992** | **1.000** | 0.978 | 0.989 | 0.945 | **0.988** | 0.966 |
| it | 0.978 | 0.962 | 0.970 | **0.996** | **0.966** | **0.981** | 0.936 | 0.962 | 0.949 |
| ja | **1.000** | **0.994** | **0.997** | **1.000** | **0.994** | **0.997** | 0.998 | **0.994** | 0.996 |
| ko | **1.000** | **0.990** | **0.995** | **1.000** | **0.990** | **0.995** | **1.000** | **0.990** | **0.995** |
| pt | 0.992 | **0.940** | 0.965 | **0.998** | 0.936 | **0.966** | 0.983 | 0.934 | 0.958 |
| ru | **0.996** | 0.996 | **0.996** | 0.994 | **0.998** | **0.996** | 0.986 | 0.994 | 0.990 |
| es | **0.998** | **0.978** | **0.988** | 0.996 | 0.974 | 0.985 | 0.990 | 0.962 | 0.976 |
| th | **1.000** | **0.988** | **0.994** | **1.000** | **0.988** | **0.994** | 0.998 | **0.988** | 0.993 |
| tr | 0.473 | **1.000** | 0.643 | **0.996** | 0.992 | **0.994** | 0.959 | 0.994 | 0.976 |
| vi | **1.000** | **0.986** | **0.993** | **1.000** | **0.986** | **0.993** | **1.000** | **0.986** | **0.993** |
| zh | **0.990** | 0.994 | 0.992 | **0.990** | **0.996** | **0.993** | 0.986 | **0.996** | 0.991 |
| **Macro Avg** | 0.889 | 0.917 | 0.897 | **0.984** | **0.983** | **0.983** | 0.946 | 0.922 | 0.905 |
| **Accuracy** | 0.917 | | | **0.983** | | | 0.922 | | |

Table 6: Results of SAE-based and neuron-based LID derived from Gemma 2 9B and fastText LID, applied to WiLI-2018 across 15 languages. **Bold** text indicates the performance is equal to or greater than the counterpart.

| Task | Accuracy | Std. Error |
|---|---|---|
| **XNLI** | | |
| bg | 0.3944 | 0.0098 |
| **de** | 0.4562 | 0.0100 |
| **en** | 0.5450 | 0.0100 |
| **es** | 0.4313 | 0.0099 |
| **fr** | 0.4618 | 0.0100 |
| **hi** | 0.3964 | 0.0098 |
| ru | 0.4659 | 0.0100 |
| **th** | 0.4092 | 0.0099 |
| tr | 0.4261 | 0.0099 |
| vi | 0.4402 | 0.0100 |
| zh | 0.3402 | 0.0095 |
| **PAWS-X** | | |
| **de** | 0.5510 | 0.0111 |
| **en** | 0.6030 | 0.0109 |
| **es** | 0.5675 | 0.0111 |
| **fr** | 0.5700 | 0.0111 |
| ja | 0.5370 | 0.0112 |
| ko | 0.5385 | 0.0111 |
| zh | 0.4985 | 0.0112 |
| **XWinograd** | | |
| **en** | 0.8129 | 0.0081 |
| **fr** | 0.6506 | 0.0527 |
| ja | 0.5996 | 0.0158 |
| **pt** | 0.6616 | 0.0292 |
| ru | 0.5968 | 0.0277 |
| zh | 0.6389 | 0.0214 |

Table 7: Llama 3.2 1B evaluation results for XNLI, PAWS-X, and XWinograd tasks in the zero-shot setting with lm-evaluation-harness (Gao et al., 2024). **Green text** indicates the languages explicitly supported by Llama 3.2 1B.

| Layer | Lang | Feature ID | Top Tokens |
|---|---|---|---|
| layers.2.mlp | Chinese | 18693 | nam, da, plet, **堂**, **公**, umba, Bundes, da, uga, **文** |
| layers.6.mlp | Chinese | 126429 | ряд, ibar, essed, cur, ar, tu, cur, quer, Ecc, ined |
| layers.7.mlp | Chinese | 82671 | imum, olith, Economics, marsh, curr, obel, .settings, Lord, esh, **滿** |
| layers.9.mlp | Chinese | 41091 | rosso, acen, agn, agner, /all, ,No, **无**, opens, -No, UGHT |
| layers.10.mlp | Chinese | 86769 | aling, aber, □, **指**, □, Sud, aled, lej, □, 便 |
| layers.10.mlp | Chinese | 119196 | _COMPAT, jig, shalt, ulia, PLUS, urn, **有**, **其**, **而**, èles |
| layers.11.mlp | Chinese | 240 | amet, Erf, tings, multiplic, atori, **影**, mast, mul, barr, amon |
| layers.11.mlp | Chinese | 38263 | rani, employed, خانه, iola, iad, ueil, ROAD, instr, ox, ût |
| layers.11.mlp | Chinese | 40026 | **们**, udes, obbled, avers, umb, odes, 們, omain, AMES, chn |
| layers.11.mlp | Chinese | 65660 | therefore, draws, همچ, drawing, Pract, Draws, vždy, есть, originally, real |
| layers.11.mlp | Chinese | 73417 | wore, lacked, ebi, simmer, 户, 家伙, **消息**, tasted, атор ,حص |
| layers.12.mlp | Chinese | 104224 | quals, **中**, **的**, **繁**, WikiLeaks, □, **之一**, **的情**, LLP, awner |
| layers.13.mlp | Chinese | 113658 | app, APP, ww, raw, WS, owe, tw, tw, the, App |
| layers.13.mlp | Chinese | 118430 | Under, Seats, ivé, Am, -Am, infos, round, works, ové, Under |
| layers.14.mlp | Chinese | 12944 | unca, นำ, **第一次**, **納**, 纳, पहल, **首**, **訪**, lan, ersten |
| layers.14.mlp | Chinese | 12987 | Water, enity, ill, seq, asy, illac, ugo, abis, Vol, **雅** |
| layers.14.mlp | Chinese | 49167 | **，它**, **，将**, **，为**, **，本**, **，其**, **，从**, **，这**, **，每**, **，大**, **，或** |
| layers.14.mlp | Chinese | 66584 | **间**, **师**, ullo, urbed, ativos, 周期, hoa, atures, iders, amentos |
| layers.14.mlp | Chinese | 88535 | □, reative, **司**, □, **余**, **任**, **文**, **但**, 示, 属 |
| layers.14.mlp | Chinese | 103432 | **程度**, **斗**, 脸, chez, **眼睛**, **身上**, **令人**, **赫**, **事情**, **反应** |
| layers.14.mlp | Chinese | 112457 | □, 锦, **到**, 则, **干**, 场, 苏, **-Fi**, □, ショ |
| layers.15.mlp | Chinese | 11994 | **奪**, an, S, We, Hi, **學校**, Tr, In, **諾**, Trav |
| layers.15.mlp | Chinese | 38094 | **提交**, **注册**, 访问, **客户**, **培训**, **活动**, **登录**, **安装**, 显示, 输出 |
| layers.15.mlp | Chinese | 114546 | **民主**, **社**, **家**, **嘉**, **世**, **利**, **邦**, **耳**, **彼**, **美** |

Figure 35: Top 10 tokens promoted by multiplying Chinese-specific features by the token unembedding matrix of Llama 3.2 1B. **Bold** text indicates tokens associated with Chinese.

<table>
<tr><th>Layer</th><th>Lang</th><th>Feature ID</th><th>Top Tokens</th></tr>
<tr><td>layers.5.mlp</td><td>French</td><td>63694</td><td>**France**, **ois**, **Antoine**, **France**, **france**, 프랑스, **Quebec**, 法国, **agne**, **Québec**</td></tr>
<tr><td>layers.8.mlp</td><td>French</td><td>86519</td><td>負, **aff**, malar, -controls, **'aff**, gos, ilde, Bowling, 励, +")</td></tr>
<tr><td>layers.9.mlp</td><td>French</td><td>16701</td><td>**ulaire**, **Marseille**, **pou**, **agne**, **travail**, **pour**, **Bordeaux**, **télé**, **istique**, **éo**</td></tr>
<tr><td>layers.9.mlp</td><td>French</td><td>44983</td><td>**iste**, **pou**, **istes**, **jeunes**, **site**, **ulaire**, **quelque**, **toujours**, **ét**, **ité**</td></tr>
<tr><td>layers.11.mlp</td><td>French</td><td>25732</td><td>**pou**, **ernes**, **encore**, **après**, **ainsi**, **aussi**, **existence**, **europé**, **aup**, **quan**</td></tr>
<tr><td>layers.11.mlp</td><td>French</td><td>110940</td><td>**même**, **seulement**, **contre**, **trait**, **ulant**, **deme**, **jamais**, **dont**, **trait**, **ache**</td></tr>
<tr><td>layers.12.mlp</td><td>French</td><td>22084</td><td>**ét**, **quelque**, **dé**, **trois**, **même**, **sous**, **pour**, **é**, **ré**, **thé**</td></tr>
<tr><td>layers.12.mlp</td><td>French</td><td>41725</td><td>sobre, Wed, urb, ough, بش, tot, Bowen, ow, **corp**, pastor</td></tr>
<tr><td>layers.13.mlp</td><td>French</td><td>6455</td><td>**l**, **le**, **les**, **la**, **(l**, л, **.l**, **{l**, **_l**, **=l**</td></tr>
<tr><td>layers.13.mlp</td><td>French</td><td>77011</td><td>**c**, **il**, **ils**, **comment**, **comment**, **Comment**, **il**, **commenting**, **pas**, **j**</td></tr>
<tr><td>layers.14.mlp</td><td>French</td><td>45764</td><td>."]<br>, ._<br>, .",<br>, ."<br>, \\<br>, .”<br>, |<br>, ,…<br>, .,<br>, •</td></tr>
</table>

Figure 36: Top 10 tokens promoted by multiplying French-specific features by the token unembedding matrix of Llama 3.2 1B. **Bold** text indicates tokens associated with French.

| Layer | Lang | Feature ID | Top Tokens |
|---|---|---|---|
| layers.2.mlp | German | 56064 | **baum**, **öt**, **feld**, **ö**, **mann**, **kre**, **platz**, \xf, ости, **auf** |
| layers.5.mlp | German | 31366 | **zburg**, **utsch**, **GmbH**, **immer**, ammer, **german**, **Germany**, lage, we, **nummer** |
| layers.8.mlp | German | 79729 | grave, eros, здесь, Serv, exposures, ck, vis, requestOptions, serialVersionUID, _args |
| layers.9.mlp | German | 9083 | rings, **chers**, **bere**, -ring, wel, ンク, nation, □, ring, bere |
| layers.10.mlp | German | 44371 | lod, zf, ,:,:, **nod**, Lod, för, zial, ໍ່ໍ, Kee, pal |
| layers.11.mlp | German | 27401 | **unge**, **unters**, **über**, **fig**, **ge**, **verst**, **nach**, **über**, **groß**, **fig** |
| layers.11.mlp | German | 52681 | **durch**, **gesch**, **erw**, **unter**, **eins**, **auf**, **auss**, **gleich**, **ange**, **voll** |
| layers.12.mlp | German | 107855 | **lichkeit**, 来说, **gaben**, 大会, behand, ậy, Whilst, tas, py, .</ |
| layers.14.mlp | German | 312 | **und**, **Und**, **Und**, **und**, **och**, **og**, **oder**, **UND**, **e**, **bzw** |
| layers.14.mlp | German | 48455 | ., .<br>, **.&**, ."<br>, :., .",<br>, .",", .";<br>, .“, ." |
| layers.15.mlp | German | 17109 | Pokemon, Taco, Spider, Elvis, **Tesla**, Viking, Batman, Dodge, Apollo, Marvel |
| layers.15.mlp | German | 75403 | **ch**, f, p, **st**, m, ke, el, h, **de**, **pen** |
| layers.15.mlp | German | 130833 | , *, , ., ", MC, CD, Ã, □, PC |

Figure 37: Top 10 tokens promoted by multiplying German-specific features by the token unembedding matrix of Llama 3.2 1B. **Bold** text indicates tokens associated with German.

| Layer | Lang | Feature ID | Top Tokens |
|---|---|---|---|
| layers.1.mlp | Italian | 28636 | term, ertiary, **mag**, WAYS, 중에, absor, terms, **ità**, 的事情, -term |
| layers.1.mlp | Italian | 47594 | endorsements, storm, шиб, -di, -files, Damage, ровер, broken, presence, endorsement |
| layers.2.mlp | Italian | 33390 | **Italian**, **Italy**, **Italians**, **tratt**, **izione**, **Italian**, **zzo**, **Italy**, **Giul**, **italiana** |
| layers.3.mlp | Italian | 120200 | **Ris**, **Luigi**, **enne**, **vent**, **igli**, **agli**, **italian**, **del**, **onta**, **atto** |
| layers.4.mlp | Italian | 114161 | **Italians**, **italiana**, **italian**, **Italia**, **Italian**, **Luigi**, **Milano**, **Italian**, **ucch**, **italiani** |
| layers.6.mlp | Italian | 109638 | romo, gian, gi, FLT, @",, **emo**, uctions, ollower, *",, ıko |
| layers.7.mlp | Italian | 52987 | radix, ـه, Hall, flex, nce, Conce, Liked, hall, Empty, heim |
| layers.8.mlp | Italian | 13547 | alcon, aldi, уди, □, Sudan, atti, ate, **gli**, addir, @g |
| layers.9.mlp | Italian | 5106 | **ò**, **aggi**, **più**, **ancora**, **migli**, **egg**, **storia**, **dei**, **ansa**, **ott** |
| layers.9.mlp | Italian | 58424 | suppl, **ott**, **att**, **butt**, ilk, **aggi**, **migli**, **pubb**, impe, **ott** |
| layers.10.mlp | Italian | 38479 | **oltre**, **nell**, **tutte**, **tratt**, **anche**, **nel**, **quanto**, **rag**, **tut**, **nella** |
| layers.10.mlp | Italian | 108570 | **già**, **oltre**, **semp**, **questa**, **pubb**, **anche**, **tratt**, **altro**, **risult**, **tutte** |
| layers.11.mlp | Italian | 14702 | tuy, apid, 抬, ئام, uellement, ément, urray, ピ, tiler, utin |
| layers.11.mlp | Italian | 35463 | **milan**, trou, timings, ater, Visual, **Milan**, Visual, -case, ميل, **Pasta** |
| layers.12.mlp | Italian | 3234 | sobre, **complet**, oron, selv, destin, gore, ill, **completo**, **attend**, **semp** |
| layers.13.mlp | Italian | 30409 | ec, **vivo**, **es**, **pi**, impe, var, collo, **e**, **ecc**, -ar |
| layers.13.mlp | Italian | 44286 | **l**, **le**, **la**, {l, .l, (l, :l, л, **lo**, **les** |
| layers.14.mlp | Italian | 90522 | **di**, **del**, **di**, **-di**, **della**, **dello**, **Di**, **Di**, **_di**, **dell** |

Figure 38: Top 10 tokens promoted by multiplying Italian-specific features by the token unembedding matrix of Llama 3.2 1B. **Bold** text indicates tokens associated with Italian.

| Layer | Lang | Feature ID | Top Tokens |
|---|---|---|---|
| layers.0.mlp | Japanese | 100720 | dou, inu, inalg, Means, WCS, histor, ız, _SPI, jid, Phon |
| layers.0.mlp | Japanese | 112017 | hci, **館**, ogui, itsu, achi, eka, unilateral, oga, PEC, piler |
| layers.8.mlp | Japanese | 32154 | elda, innie, alt, roz, programma, contro, DRAM, ambi, arious, itar |
| layers.9.mlp | Japanese | 34376 | ihu, □, ativní, granularity, Morav, Suppress, upro, upert, _REFERER, grav |
| layers.11.mlp | Japanese | 29972 | azaar, 되어, **ご**, sidewalks, /org, **代**, ôte, -match, Teams, Commit |
| layers.11.mlp | Japanese | 126209 | afen, ）**は**, Generic, Leh, apos, **フ**, **略**, owo, Generic, generic |
| layers.12.mlp | Japanese | 25492 | orm, **していた**, .$., Own, om, angep, ramids, SEQ, lical, *', |
| layers.12.mlp | Japanese | 121526 | HEL, UFFER, Password, isol, uffers, udd, elda, brief, ◌̆◌̌, hypers |
| layers.13.mlp | Japanese | 44429 | viz, Nicht, Wallace, kHz, vak, lượt, Neck, ault, **性的**, Gel |
| layers.13.mlp | Japanese | 72538 | **シ**, **発**, **マ**, **では**, **リ**, **広**, **ライ**, **プ**, **サ**, **メ** |
| layers.13.mlp | Japanese | 77268 | oga, abinet, lint, **ント**, **舗**, **のだ**, **目を**, Nations, нів, **平** |
| layers.13.mlp | Japanese | 80366 | ropol, Orn, inally, ment, privacy, Ist, ist, privacy, advertiser, tong |
| layers.13.mlp | Japanese | 84606 | loth, Sawyer, Skyl, □, rees, root, opc, 制, cee, oser |
| layers.13.mlp | Japanese | 101925 | **があり**, **措**, orts, Vlad, **をした**, **から**, ulaire, **には**, utterstock, olor |
| layers.13.mlp | Japanese | 121228 | **イス**, **フィ**, **ット**, **ラ**, **イ**, **ロ**, **Ơ**, **ク**, **素**, **ばかり** |
| layers.13.mlp | Japanese | 125863 | pora, #undef, Alpine, drift, decorators, omit, pulse, RECEIVE, Actual, Zub |
| layers.14.mlp | Japanese | 17056 | кан, rum, umin, **こ**, □, **なが**, 耳, **とも**, rez, **乃** |
| layers.14.mlp | Japanese | 17850 | **かに**, □, unal, **なる**, **な**, **により**, 故, **とする**, alar, **が出** |
| layers.14.mlp | Japanese | 34692 | **たら**, **間に**, **頂**, **ざ**, **わ**, **ば**, **級**, **細**, **ね**, **節** |
| layers.14.mlp | Japanese | 43240 | And, And, OMEM, -validation, Demp, **、小**, /angular, **和**, **児**, AndUpdate |
| layers.14.mlp | Japanese | 44808 | ctors, ypc, sez, omen, TW, inters, isode, Ens, ypr, -dist |
| layers.14.mlp | Japanese | 77483 | vol, **及び**, **福**, ve, ordered, **的手**, abb, ikers, pios, τέ |
| layers.14.mlp | Japanese | 85299 | ле, usi, era, 998, uk, okoj, □, and, -life, 258 |
| layers.14.mlp | Japanese | 97905 | asured, aff, her, .cl, bose, anson, icken, screened, dent, ash |
| layers.14.mlp | Japanese | 98700 | pay, issy, □, gre, patrick, iky, 丁, □, âk, igham |
| layers.14.mlp | Japanese | 119697 | hai, -blog, ht, pomoc, 안, hd, Blog, hest, valid, col |
| layers.15.mlp | Japanese | 4131 | ghi, edic, **国際**, NI, edu, ERIC, LM, **eds**, TJ, **桑** |
| layers.15.mlp | Japanese | 12716 | redi, es, каж, ve, **的**, dom, Jeho, 至少, 把, stick |
| layers.15.mlp | Japanese | 25135 | , **しかし**, itr, **それは**, **けれど**, **そんな**, I, **俺は**, iyi, ียบ |
| layers.15.mlp | Japanese | 35007 | **像**, **意思**, **事情**, **等**, **表示**, **について**, عا, olang, **によ**, **利用** |
| layers.15.mlp | Japanese | 95097 | abase, **しており**, ạn, **しました**, атег, **ください**, Meh, usz, hindsight, аци |
| layers.15.mlp | Japanese | 95151 | **いた**, **っ**, **いて**, **っても**, **えない**, **たく**, **えば**, **いている**, **らしい**, **さんが** |
| layers.15.mlp | Japanese | 102746 | **ジャ**, **ドラ**, **スマ**, **ランド**, **ハイ**, **ブリ**, **イヤ**, **ヌ**, **ミュ**, **キュ** |

Figure 39: Top 10 tokens promoted by multiplying Japanese-specific features by the token unembedding matrix of Llama 3.2 1B. **Bold** text indicates tokens associated with Japanese.

| Layer | Lang | Feature ID | Top Tokens |
|---|---|---|---|
| layers.0.mlp | Korean | 87423 | **kim**, velt, Office, PMC, orney, ordinances, agre, %/, agony, office |
| layers.4.mlp | Korean | 99278 | **Kim**, **Seoul**, **Jae**, **Kim**, **.kr**, **Samsung**, **Korean**, **Korea**, **Shin**, **Koreans** |
| layers.6.mlp | Korean | 109962 | Highlands, ussels, iteur, tails, Ire, ativ, $sql, plant, jsp, beginnings |
| layers.7.mlp | Korean | 115377 | vain, ane, ANE, lett, att, anes, roc, 区, igne, olina |
| layers.8.mlp | Korean | 94922 | ante, iente, innie, ensed, λλη, ssh, flips, uş, Liter, itar |
| layers.9.mlp | Korean | 31611 | beck, acles, ataka, batt, -wrap, etat, pty, umn, Kear, chat |
| layers.9.mlp | Korean | 53931 | iosa, gli, chet, coni, Giov, 89, adero, ewolf, worm, edit |
| layers.10.mlp | Korean | 96918 | -Feb, partial, juan, pur, ipro, capitals, URATION, 膜, [[, experiences |
| layers.10.mlp | Korean | 125019 | 当, Mori, Petty, Yug, Touch, _che, bare, VIC, .jp, omi |
| layers.11.mlp | Korean | 64252 | viz, avour, aura, 今日, 琴, UPDATED, apl, intend, intention, ITT |
| layers.11.mlp | Korean | 102511 | r, **으로**, _FR, aar, l, war, **은**, Protocol, айд, **이라** |
| layers.12.mlp | Korean | 27775 | ptime, epochs, ascending, located, plays, inf, idual, Microsystems, 舗, layers |
| layers.12.mlp | Korean | 67845 | **Kim**, **Korea**, **Korean**, **Seoul**, **Park**, **Kim**, **Hyundai**, **kim**, **Koreans**, **Je** |
| layers.13.mlp | Korean | 78512 | plays, **바람**, play, Maar, **편**, Boost, maz, robots, antics, 帖 |
| layers.14.mlp | Korean | 87496 | **상**, **령**, **선**, **이상**, **나**, **신**, **증**, **등**, **한**, **왕** |
| layers.14.mlp | Korean | 107903 | asc, ey, anz, erc, acic, eh, ans, eyi, ekl, erv |
| layers.15.mlp | Korean | 17864 | **이야기**, **그것**, **있었다**, **없었다**, **사람들이**, **인정**, **행동**, **사람은**, **이야**, **그를** |
| layers.15.mlp | Korean | 41829 | **척**, **찰**, **리로**, **리카**, **림**, □, **어나**, **린이**, **니아**, **택** |
| layers.15.mlp | Korean | 57880 | ?, ?<br>, ?", ?”, ?"<br>, ?', ?(, ?,, ?’, ?: |
| layers.15.mlp | Korean | 60936 | **통**, **작**, **소**, **지**, **그**, **기**, **판**, **누**, 빙, **식** |
| layers.15.mlp | Korean | 114116 | LETE, Passive, **설명**, flashlight, **게임**, careful, REVIEW, ▼, 中文, 问 |
| layers.15.mlp | Korean | 131027 | ,, reckon, örper, **림**, .., **간**, **모습**, ,',, **손**, лива |

Figure 40: Top 10 tokens promoted by multiplying Korean-specific features by the token unembedding matrix of Llama 3.2 1B. **Bold** text indicates tokens associated with Korean.

| Language | Detection | | | | Fuzzing | | | |
|---|---|---|---|---|---|---|---|---|
| | Acc. | F1 | Prec. | Rec. | Acc. | F1 | Prec. | Rec. |
| Bulgarian | 0.79 | 0.77 | 0.89 | 0.73 | 0.77 | 0.76 | 0.83 | 0.77 |
| Chinese | 0.61 | 0.59 | 0.69 | 0.60 | 0.68 | 0.65 | 0.72 | 0.64 |
| English | 0.59 | 0.66 | 0.59 | 0.80 | 0.66 | 0.62 | 0.74 | 0.62 |
| French | 0.82 | 0.78 | 0.94 | 0.68 | 0.82 | 0.81 | 0.89 | 0.75 |
| German | 0.86 | 0.83 | 0.96 | 0.74 | 0.83 | 0.81 | 0.89 | 0.75 |
| Hindi | 0.78 | 0.70 | 0.96 | 0.58 | 0.78 | 0.72 | 0.92 | 0.63 |
| Italian | 0.69 | 0.58 | 0.84 | 0.53 | 0.68 | 0.62 | 0.75 | 0.61 |
| Japanese | 0.70 | 0.60 | 0.92 | 0.49 | 0.73 | 0.67 | 0.87 | 0.58 |
| Korean | 0.69 | 0.53 | 0.93 | 0.43 | 0.69 | 0.58 | 0.87 | 0.50 |
| Portuguese | 0.85 | 0.82 | 0.93 | 0.77 | 0.82 | 0.81 | 0.86 | 0.80 |
| Russian | 0.74 | 0.65 | 0.93 | 0.52 | 0.75 | 0.71 | 0.86 | 0.64 |
| Spanish | 0.74 | 0.70 | 0.85 | 0.60 | 0.70 | 0.71 | 0.73 | 0.72 |
| Thai | 0.70 | 0.67 | 0.83 | 0.63 | 0.69 | 0.68 | 0.78 | 0.70 |
| Turkish | 0.78 | 0.71 | 0.94 | 0.61 | 0.78 | 0.74 | 0.89 | 0.68 |
| Vietnamese | 0.76 | 0.68 | 0.92 | 0.59 | 0.77 | 0.70 | 0.90 | 0.63 |
| **Average** | **0.74** | **0.68** | **0.87** | **0.62** | **0.74** | **0.71** | **0.83** | **0.67** |

Table 8: Average interpretation scores of language-specific features in Llama 3.2 1B for detection and fuzzing across languages.

| Layer | Lang | Feature ID | Top Tokens |
|---|---|---|---|
| layers.4.mlp | Russian | 48343 | dra, **Ha**, **Pavel**, **standart**, turned, **Ha**, **Vladimir**, amounted, **Vlad**, **Ukr** |
| layers.6.mlp | Russian | 121507 | acz, port, đương, zik, swer, clist, eed, lish, Niet, activity |
| layers.7.mlp | Russian | 110894 | ört, gun, sectional, pix, ø, (typeof, Schools, bond, Havana, Vendor |
| layers.8.mlp | Russian | 69357 | Ell, Qing, 말, ля, िख, Lou, ोख, Antar, tail, sire |
| layers.8.mlp | Russian | 109317 | isse, Maid, MainActivity, allowed, .low, AIN, **-м**, -shell, 】, weed |
| layers.9.mlp | Russian | 30251 | itious, ijo, sexual, @, **&**, ri, anky, Tec, „ @n |
| layers.9.mlp | Russian | 59767 | Corona, Forward, Forward, Fits, _CORE, urette, Loader, Arg, øj, Demo |
| layers.10.mlp | Russian | 56594 | **Sergey**, **Moscow**, **kh**, **Dmitry**, **Russian**, **Mikhail**, **.ru**, **Sergei**, **Dmit**, **enko** |
| layers.11.mlp | Russian | 8452 | **п**, **в**, **Р**, **дв**, **с**, **к**, **П**, **ок**, **с**, **Г** |
| layers.11.mlp | Russian | 74675 | Major, Major, ав, Band, major, bands, -band, major, **д**, Bands |
| layers.12.mlp | Russian | 31110 | flat, iones, ereco, ushing, Flat, Ledger, **бы**, amat, 兽, iona |
| layers.12.mlp | Russian | 48228 | undy, ickle, **ела**, Buen, acidad, apo, idos, **вед**, ISH, **веч** |
| layers.13.mlp | Russian | 29244 | **группы**, **статьи**, oki, **войны**, **osoby**, **устройства**, **культуры**, **таки**, **ины**, **карти** |
| layers.13.mlp | Russian | 40052 | hol, hol, hower, 等, **чи**, aul, iked, Hol, FEMA, **дня** |
| layers.13.mlp | Russian | 50197 | eer, **yii**, MAD, mad, **об**, ыс, **ц**, **коп**, Americas, 叶 |
| layers.13.mlp | Russian | 68491 | **akh**, **ах**, **лях**, **ках**, **ших**, **kh**, **ных**, **овых**, **анных**, **щих** |
| layers.13.mlp | Russian | 91930 | mark, me, yours, ithe, **компании**, :f, plast, test, him, autom |
| layers.13.mlp | Russian | 97322 | **г**, dda, essional, □, **ось**, ittest, endif, **прав**, jure, osci |
| layers.13.mlp | Russian | 103291 | inside, within, v, inside, Within, within, Inside, Inside, Within, during |
| layers.13.mlp | Russian | 115738 | **еп**, **евые**, **ев**, **еры**, echa, eyn, ebi, imenti, **ева**, 矢 |
| layers.13.mlp | Russian | 120158 | ango, Invisible, USD, uppies, ctest, ANGO, Ghost, s, Provider, cooper |
| layers.14.mlp | Russian | 44860 | **возмож**, **сог**, **ое**, **выб**, **больш**, **ён**, **ож**, **внутрен**, **енные**, **соврем** |
| layers.14.mlp | Russian | 71946 | **име**, **возв**, **вов**, wash, **мужчин**, **уже**, locked, **вещ**, **когда**, **предпоч** |
| layers.15.mlp | Russian | 56964 | **лении**, **енными**, **ственных**, **ственного**, **лением**, **нему**, **анием**, **ательных**, **ческих**, **енности** |
| layers.15.mlp | Russian | 65387 | rh, fl, **помощи**, Santana, ral, north, mid, Mellon, **действительно**, **ывает** |
| layers.15.mlp | Russian | 119968 | atum, alling, affle, **аться**, **ать**, iat, **айте**, ats, af, **аю** |

Figure 41: Top 10 tokens promoted by multiplying Russian-specific features by the token unembedding matrix of Llama 3.2 1B for layers 14–15. **Bold** text indicates tokens associated with Russian.

| Layer | Lang | Feature ID | Top Tokens |
|---|---|---|---|
| layers.2.mlp | Spanish | 17631 | **Ñ**, **Herrera**, اخ, **Españ**, **arde**, **íg**, **itas**, **Este**, **ína**, Greatest |
| layers.3.mlp | Spanish | 4518 | **normalize**, **normalization**, **iales**, **constructors**, **ices**, **nacional**, **cales**, **Nuevo**, **Puerto**, **Penal** |
| layers.5.mlp | Spanish | 32141 | **Gal**, **gal**, едь, uced, Tx, **ga**, **Tor**, **tor**, ocol, fram |
| layers.11.mlp | Spanish | 94207 | ya, **desarroll**, **mayor**, **cual**, **hoy**, **más**, **suger**, **menos**, **que**, **ci** |
| layers.13.mlp | Spanish | 53088 | **es**, **Es**, **Es**, **_es**, **fue**, **era**, **puede**, **-es**, **ha**, **ES** |

Figure 42: Top 10 tokens promoted by multiplying Spanish-specific features by the token unembedding matrix of Llama 3.2 1B. **Bold** text indicates tokens associated with Spanish.

| Layer | Lang | Feature ID | Top Tokens |
|---|---|---|---|
| layers.0.mlp | Vietnamese | 11601 | **utura**, 立, idelberg, ота, ту, яб, ём, ighth, 雅, .heap |
| layers.0.mlp | Vietnamese | 102922 | **benh**, **tat**, **tapi**, **untas**, beck, ulate, ']:, condition, Conditions, Covered |
| layers.3.mlp | Vietnamese | 14622 | ilot, анов, iste, elligence, ista, angl, ekk, istes, dit, esta |
| layers.4.mlp | Vietnamese | 110587 | nes, bt, bit, ō, simulation, optimizer, advanced, .logical, uga, cell |
| layers.5.mlp | Vietnamese | 49501 | Romanian, Romania, arium, Orioles, 楚, ovna, fortunes, escape, alat, ernes |
| layers.7.mlp | Vietnamese | 28785 | 千, actor, iere, watch, regor, alian, heel, bä, μή, watch |
| layers.8.mlp | Vietnamese | 79242 | stro, autob, benign, IDGE, Mandatory, istance, stin, pon, Jaw, 諸 |
| layers.9.mlp | Vietnamese | 61383 | enga, ella, angle, ěj, Zuk, obel, expert, apro, cassette, issor |
| layers.10.mlp | Vietnamese | 42004 | áš, 族自治, outh, stan, 酸, atti, aste, Gujar, boiler, считается |
| layers.10.mlp | Vietnamese | 42167 | cuent, Ú, Sez, yn, strument, ène, ossal, ifferent, Contemporary, Cancelled |
| layers.11.mlp | Vietnamese | 4065 | **ho**, **Ho**, **Ho**, anness, anguage, osy, urd, acie, oppins, цез |
| layers.11.mlp | Vietnamese | 58535 | **anh**, **ph**, **nh**, **tinh**, **Vietnamese**, **Nguyen**, **Thanh**, **Ho**, **Viet**, **sinh** |
| layers.11.mlp | Vietnamese | 66074 | **Mon**, **Ph**, **Y**, **yt**, dle, antine, C, 论, loit, Cone |
| layers.12.mlp | Vietnamese | 44940 | μά, труда, Teh, conto, ulled, ccd, thresh, ไหม, ialized, marketing |
| layers.12.mlp | Vietnamese | 56617 | geries, multiples, унк, čka, анка, 沿, iership, nič, yanında, Cleveland |
| layers.12.mlp | Vietnamese | 63803 | **Than**, **Van**, **VAN**, **Van**, **hung**, **-Tr**, **Tran**, **Thi**, **Nguyen**, **THAN** |
| layers.12.mlp | Vietnamese | 92851 | emp, нова, umb, mega, seb, edy, ENCY, **cé**, िह, ora |
| layers.12.mlp | Vietnamese | 100653 | œ, 鐘, eg, Gift, ovna, esthes, loop, 刻, olean, sex |
| layers.12.mlp | Vietnamese | 124904 | iali, promin, Dest, lier, Round, ROID, Dipl, mittel, ुव, 剩 |
| layers.13.mlp | Vietnamese | 7310 | Paste, ivity, Paste, ities, adamente, last, hol, **ĩnh**, umpt, 金 |
| layers.13.mlp | Vietnamese | 19743 | ори, loi, ада, coli, athing, qed, aign, 法, icates, rored |
| layers.13.mlp | Vietnamese | 65213 | -i, adden, UFFER, кры, oles, Olymp, Mix, linked, RATION, RING |
| layers.13.mlp | Vietnamese | 75072 | êt, affer, ition, nov, iger, Bris, egra, trib, ril, ови |
| layers.13.mlp | Vietnamese | 77971 | qt, оба, formation, athlete, atorial, connect, RPC, ň, **dae**, cong |
| layers.13.mlp | Vietnamese | 98734 | **IV**, 행, **Giang**, **Ho**, **Hoa**, **Hut**, **Sq**, **VO**, **Vu**, **Iv** |
| layers.13.mlp | Vietnamese | 110071 | uts, _locator, weld, oard, Presenter, nen, locator, soever, OX, awan |
| layers.14.mlp | Vietnamese | 54719 | **ẩu**, **ôn**, **àm**, **ưa**, **ần**, **ăm**, **ộn**, **ờ**, **ướng**, **ấp** |
| layers.14.mlp | Vietnamese | 59851 | hoot, ala, сол, _MODULES, ALA, olas, cio, **thiểu**, plode, po |
| layers.14.mlp | Vietnamese | 71834 | **Gi**, **Hi**, **Th**, **Tr**, **Hi**, **T**, intervening, **Bi**, **Hy**, **B** |
| layers.14.mlp | Vietnamese | 73696 | **cho**, urs, **nguyên**, unique, vers, uchs, iaz, mixing, ica, **tập** |
| layers.14.mlp | Vietnamese | 111142 | **than**, **-than**, **Than**, **than**, **THAN**, **Than**, **niż**, **än**, **_than**, **než** |
| layers.14.mlp | Vietnamese | 121534 | **Cu**, **Vu**, trans, **cu**, **vn**, ected, **duc**, **Tu**, **Vo**, □ |
| layers.14.mlp | Vietnamese | 130249 | lein, je, chie, etable, ambre, erp, евич, emon, nie, je |
| layers.15.mlp | Vietnamese | 489 | arda, variation, appearance, Grad, enders, 359, adio, Scot, imize, knowledge |
| layers.15.mlp | Vietnamese | 56098 | **tầm**, **lợi**, **biện**, **chứng**, **ấm**, **thần**, **đỡ**, **xấu**, **chỗ**, **thảo** |
| layers.15.mlp | Vietnamese | 92076 | **Quảng**, **À**, **Hồ**, **Ninh**, □, **Urb**, **Đài**, **Nguyễn**, **Khu**, **Ô** |
| layers.15.mlp | Vietnamese | 114319 | **VN**, **Ngh**, **Thanh**, **Ninh**, **VN**, **Kinh**, **Trang**, **Hoa**, **Trung**, **Nha** |

Figure 43: Top 10 tokens promoted by multiplying Vietnamese-specific features by the token unembedding matrix of Llama 3.2 1B. **Bold** text indicates tokens associated with Vietnamese.

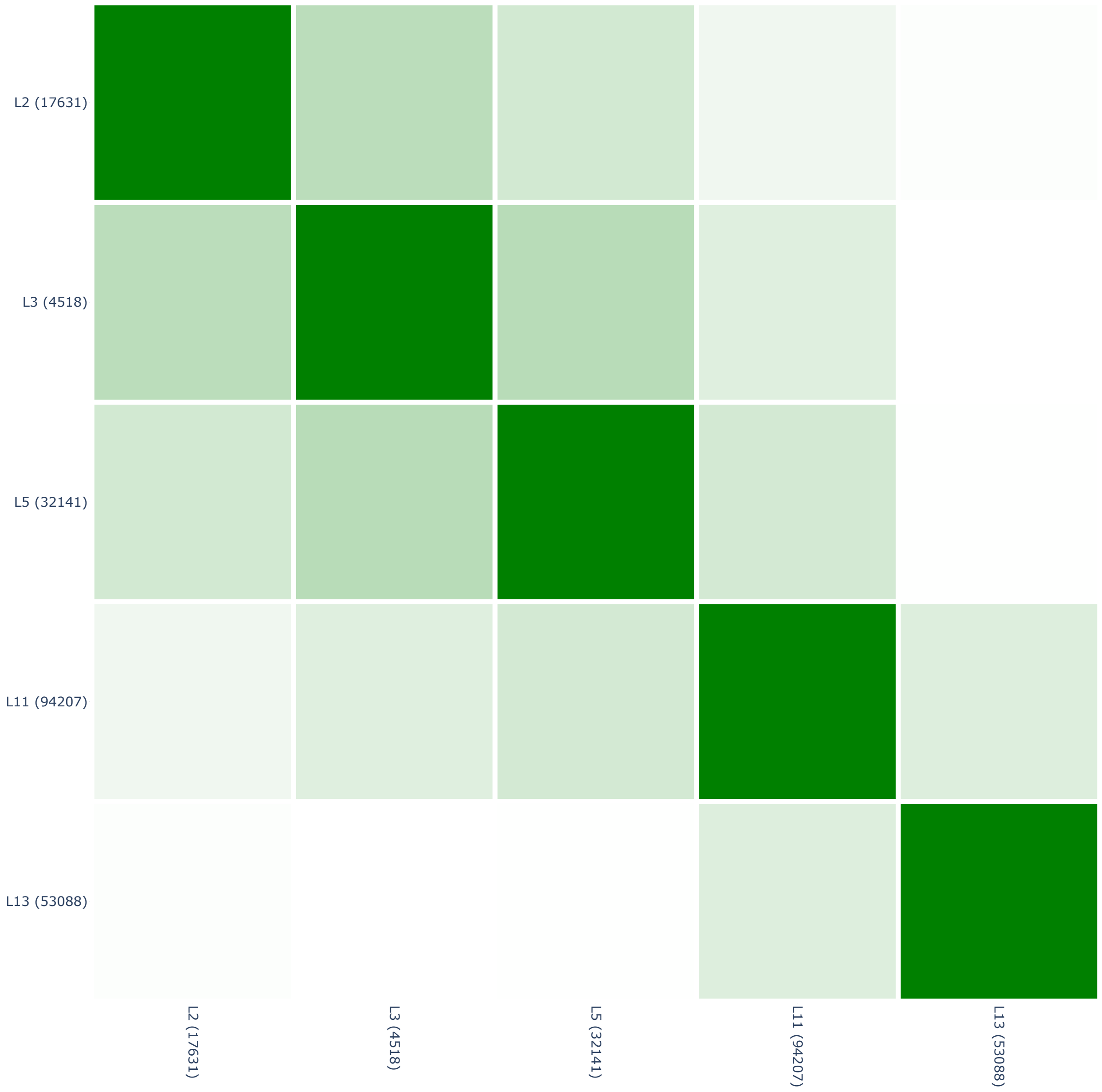

Figure 44: Intersection over union of activating tokens in Llama 3.2 1B for Spanish-specific features across layers.

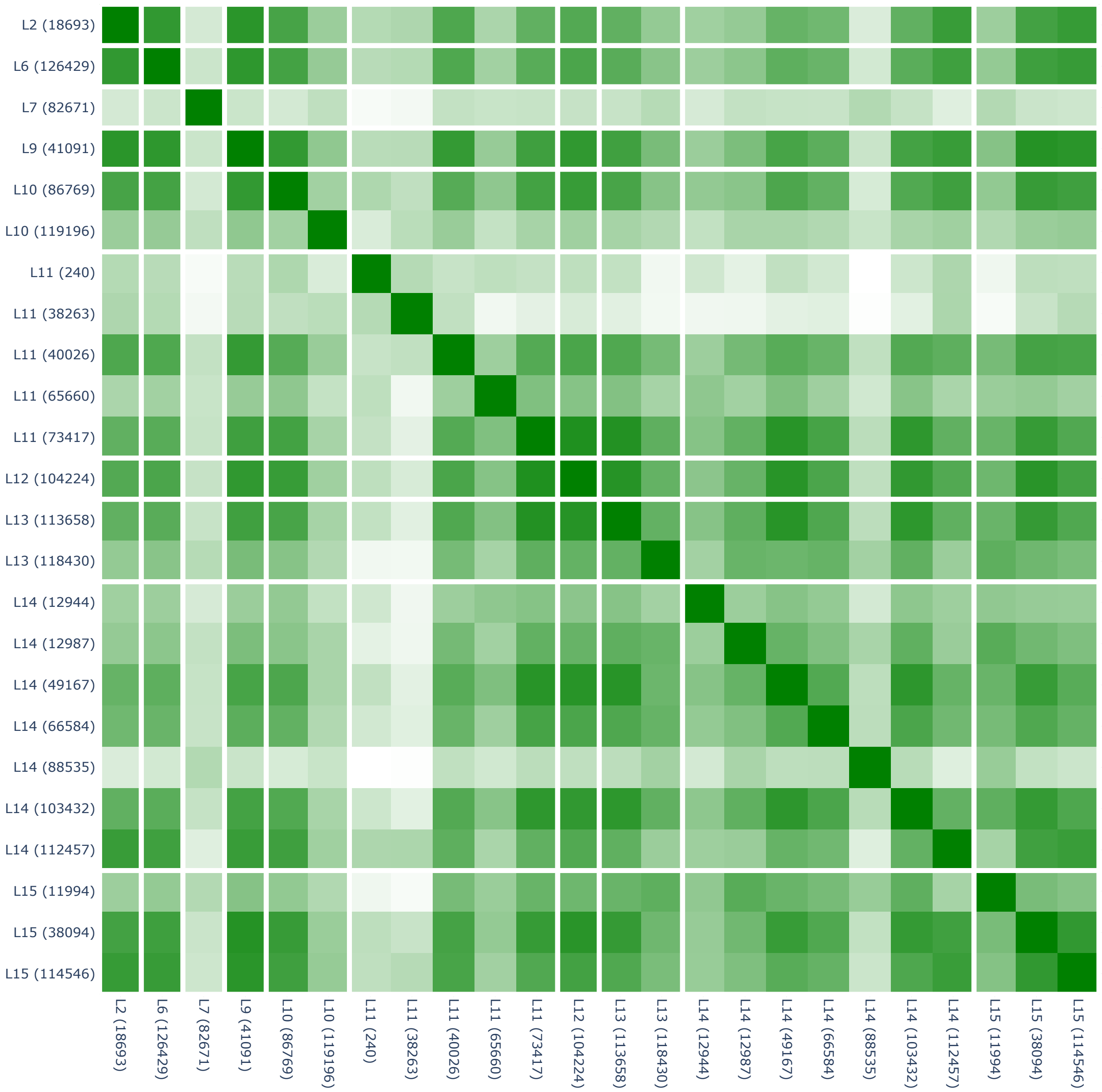

Figure 45: Intersection over union of activating tokens in Llama 3.2 1B for Chinese-specific features across layers.

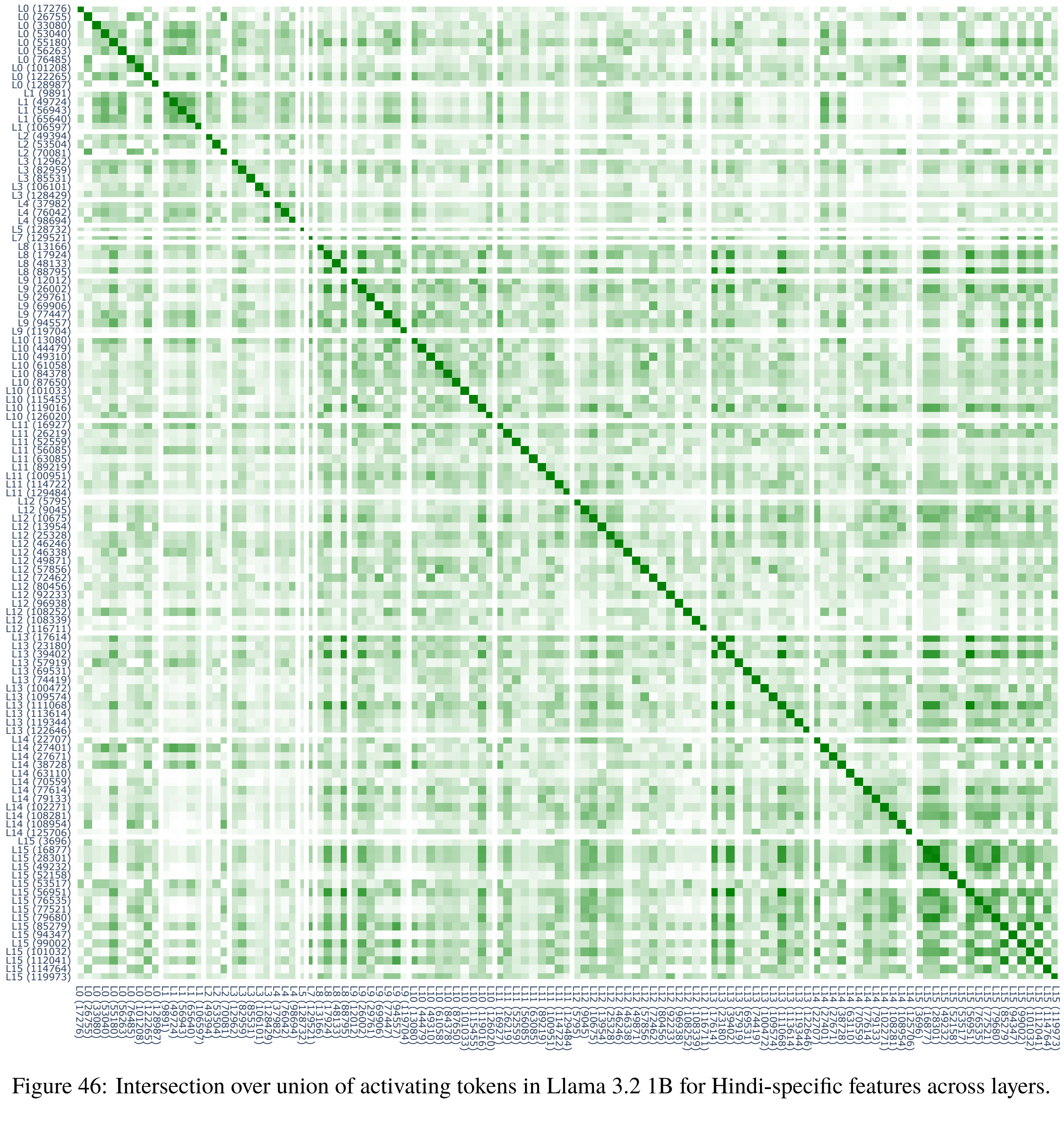

Figure 46: Intersection over union of activating tokens in Llama 3.2 1B for Hindi-specific features across layers.

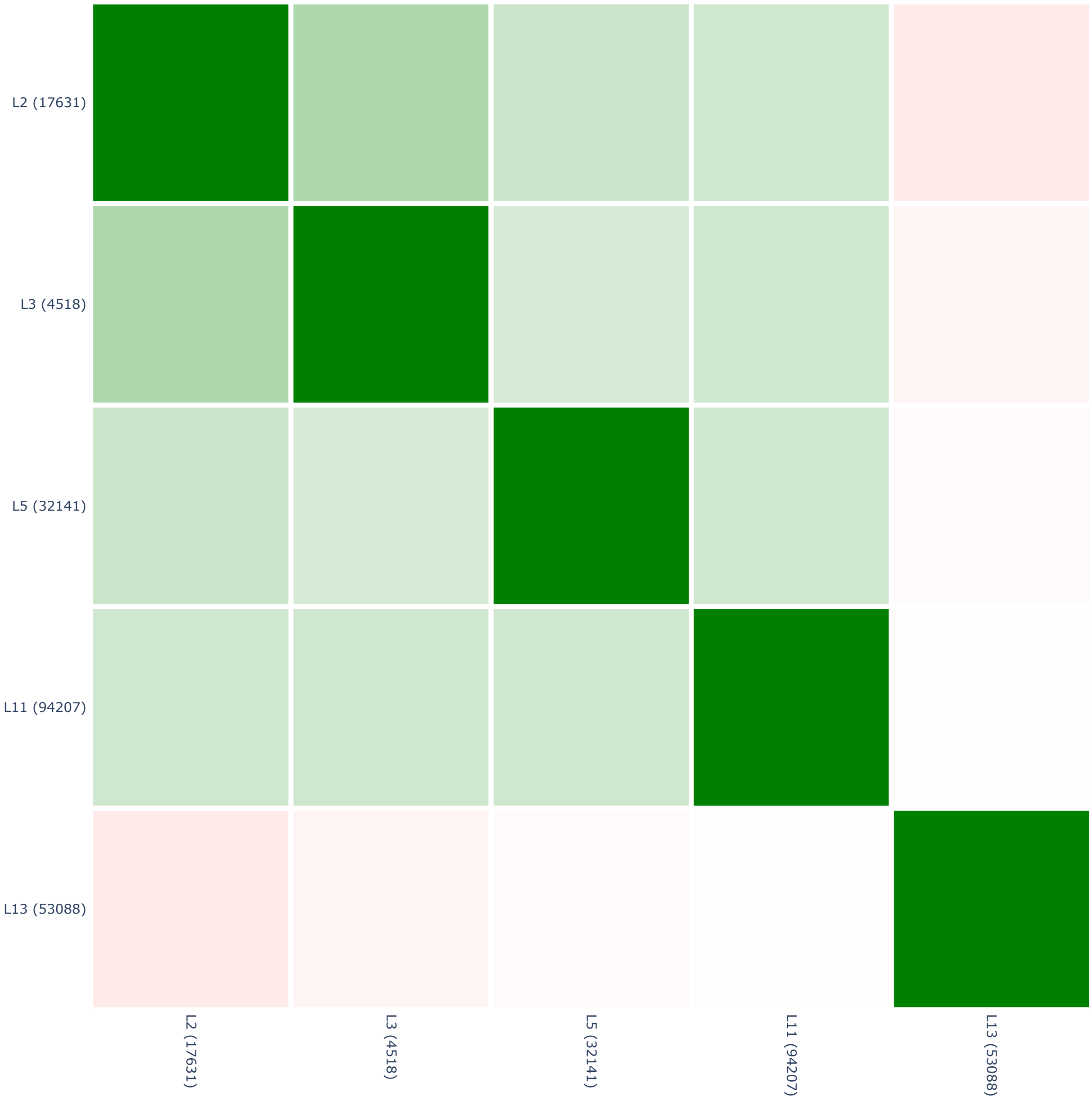

Figure 47: Pearson correlation of activating tokens in Llama 3.2 1B for Spanish-specific features across layers.

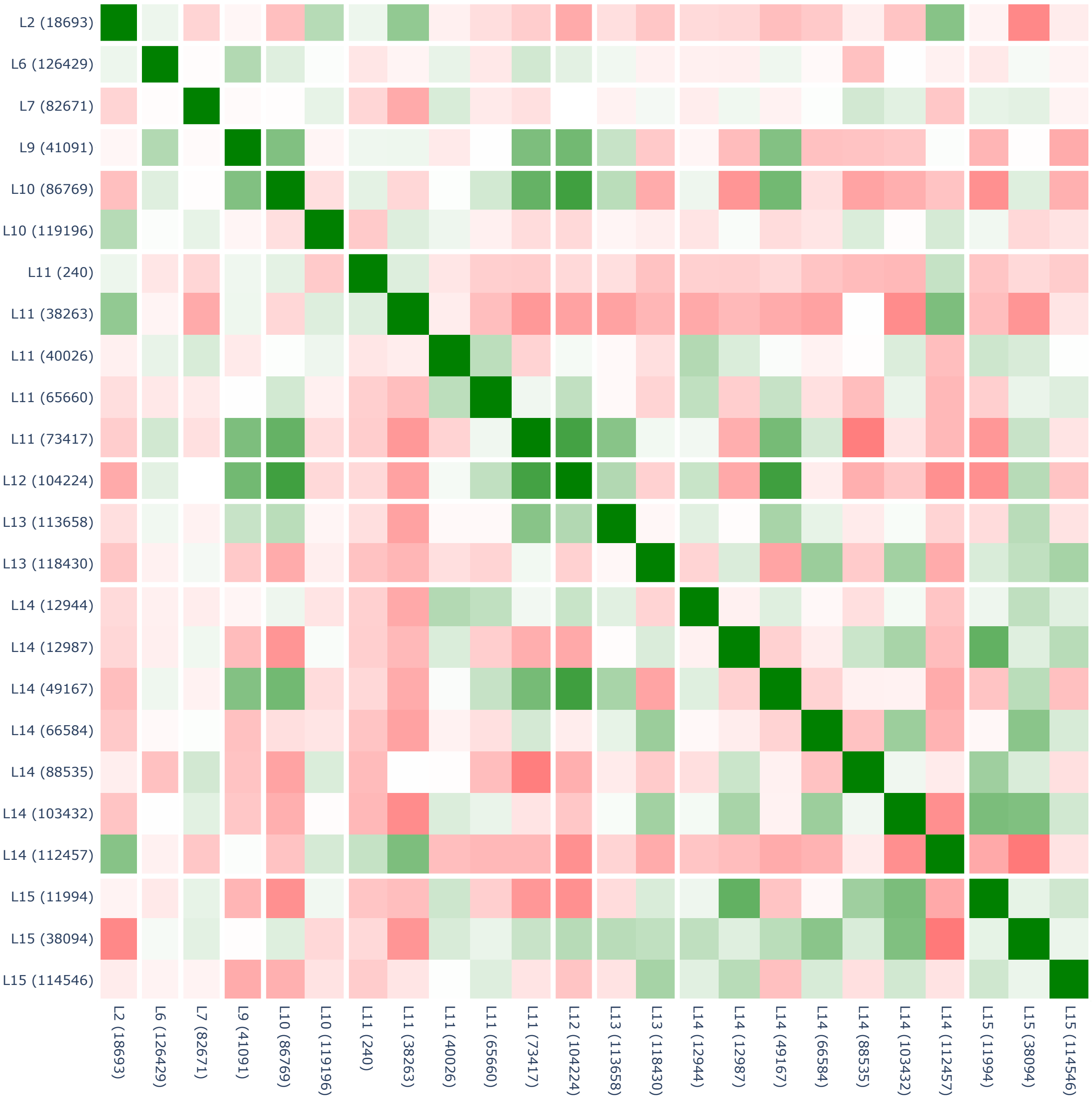

Figure 48: Pearson correlation of activating tokens in Llama 3.2 1B for Chinese-specific features across layers.

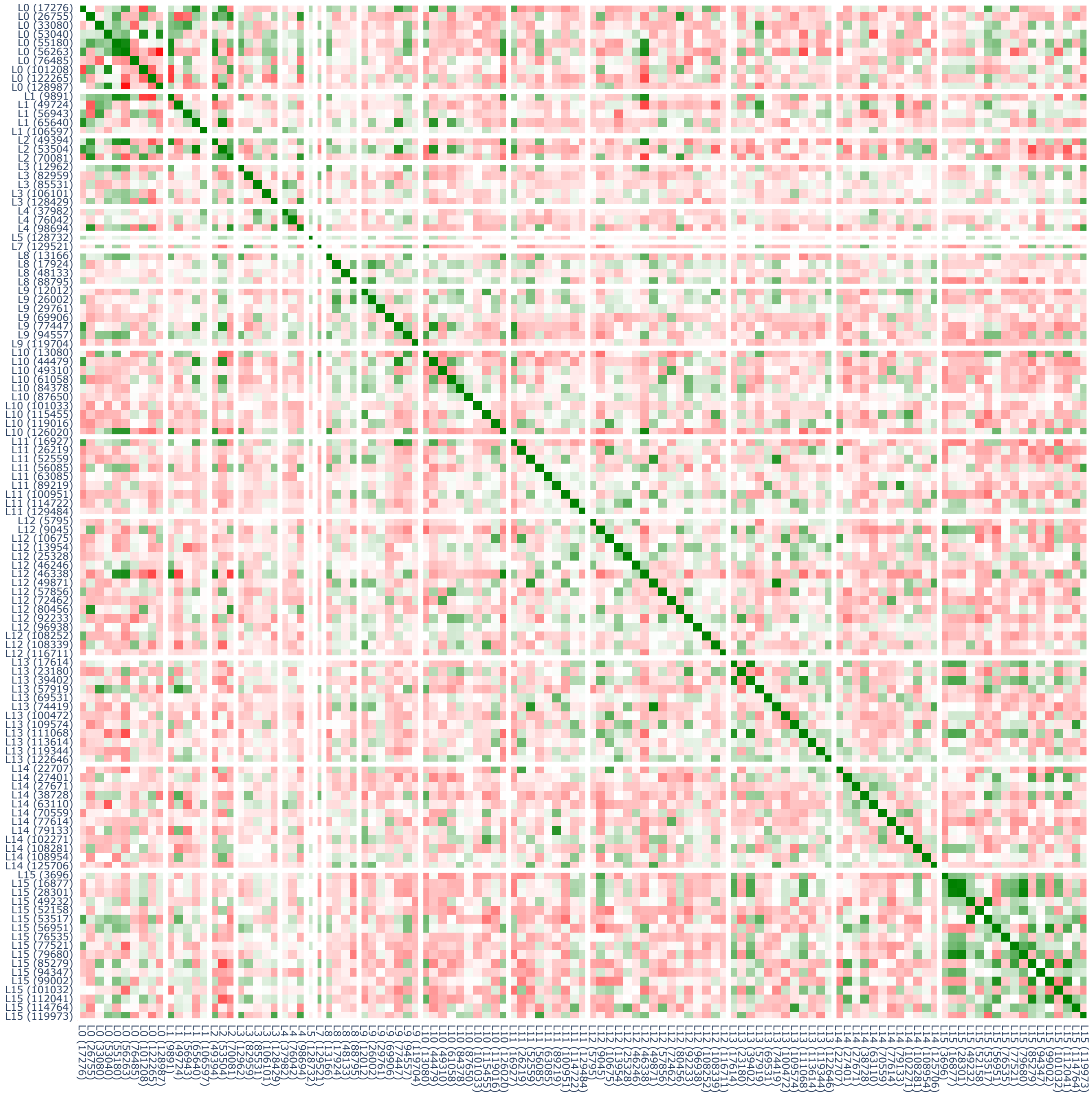

Figure 49: Pearson correlation of activating tokens in Llama 3.2 1B for Hindi-specific features across layers.

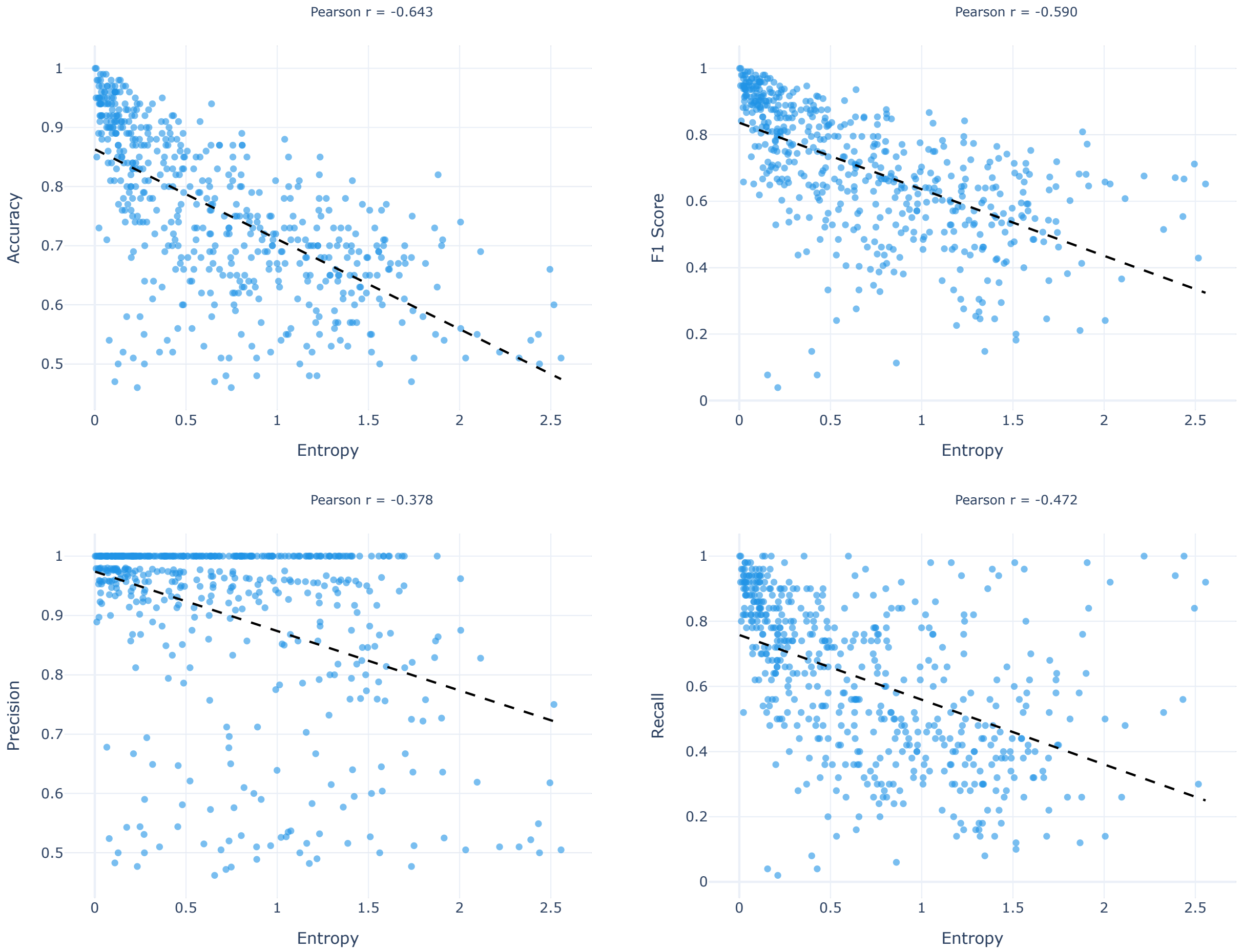

Figure 50: Pearson correlations between interpretation scores and entropy for detection scoring in Llama 3.2 1B.

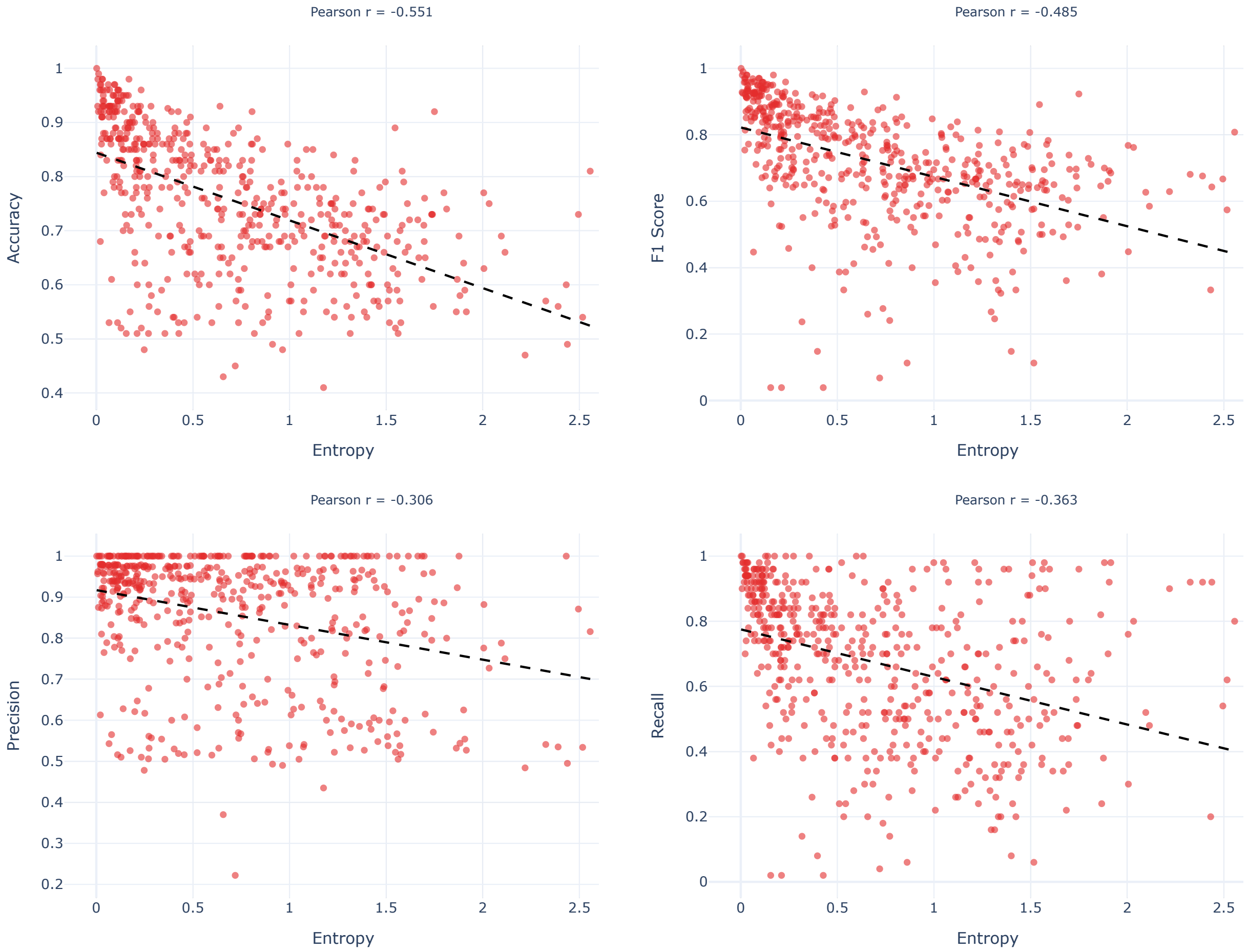

Figure 51: Pearson correlations between interpretation scores and entropy for fuzzing scoring in Llama 3.2 1B.

## Feature 67343

Language: English
Model: meta-llama/Llama-3.2-1B
Layer: model.layers.15.mlp
SAE Model: EleutherAI/sae-Llama-3.2-1B-131k
Selected Token Probability: 0.308
Entropy: 0.723

### Activation Range

0.332-1.973 1.973-3.614 3.614-5.255 5.255-6.896 6.896-8.537 8.537-10.179 10.179-11.82 11.82-13.461 13.461-15.102 15.102-16.743

### Interpretation

"The text contains a wide variety of conversational and narrative fragments, including idiomatic expressions, filler words, discourse markers, and references to people, places, and actions. There is a frequent use of pronouns, conjunctions, and auxiliary verbs, as well as sequences that reflect spoken language patterns such as repetitions, hesitations, and clarifications. Many segments highlight the structure of dialogue, question-answer formats, and descriptive or explanatory statements, often focusing on actions, states, or relationships between entities."

| Score Type | Accuracy | Precision | Recall | F1 score | TPR | TNR | FPR | FNR |
|---|---|---|---|---|---|---|---|---|
| detection | 0.72 | 0.712 | 0.74 | 0.725 | 0.74 | 0.7 | 0.3 | 0.26 |
| fuzz | 0.57 | 0.613 | 0.38 | 0.469 | 0.38 | 0.76 | 0.24 | 0.62 |

### English

#examples: [('xnli', 1000), ('paws-x', 986), ('flores', 990)]

**xnli-188.** <|begin_of_text|>yeah yeah glad to see y 'all taken care of well the i think what changed everything and uh is uh y 'all were y 'all the only ones that make any money for TI here in the last two years In the past few years , no one has made any money for TI .

**xnli-34.** <|begin_of_text|>Felicia 's Journey takes place behind the eyes of its central a young Irish girl , Felicia , who crosses the sea to England in a hopeful quest to find the father of her unborn child ; and the fat , middle-aged catering manager , Hiditch , who takes a paternal interest in the lass when it becomes clear that her young man has caddishly given her the slip . The woman did not care where the man was as long as it was far .

**xnli-936.** <|begin_of_text|>was this uh done not by TI people was it It wasn 't done by TI people .

**xnli-230.** <|begin_of_text|>it 's um it 's it 's amazingly less expensive than it is It 's much more expensive that way .

**xnli-85.** <|begin_of_text|>I 'm not into this Multivista stuff , because what is some dumpy system when compared to a beautiful hyperextension of the sun somewhere over Kuchara , when compared to the golden hue of onion fried with the kse-fi waves , when compared to the number of dividers for credit membranes in a wallet of a rich man , when compared to the magnificent smell of a briessante roll dunked in wholesome milk synthetically enriched with substances boosting the secretion of happiness hormones , that one from two years ago , not three , ' Gonzo said in a tone characteristic for a man who just discovered a solution to his life problem . He 's not into this multivista stuff because what is some dumpy system compared to the magnificent smell of a cinnamon roll dunked in wholesome milk .

**xnli-743.** <|begin_of_text|>yes it really is and that way we 're not really missing anything out you know of those children uh-huh We 're not really missing out on childcare benefits .

**xnli-317.** <|begin_of_text|>um-hum yeah i end up well yeah i mean i do a lot of like even weekender kind of things i go out for just one or two nights uh it ' s not my favorite thing but i can do it a lot more often so like over at the uh particularly in the uh the fall and spring when the insects haven 't come out yet uh where they 're already gone uh you know i spend almost every weekend up in the mountains and i guess i live about five hours away from some place to go hiking where i am now I don 't mind camping for a couple nights in the fall and spring .

**xnli-169.** <|begin_of_text|>and if you yeah if you want to so at least you you know you wouldn 't have any surprises you can order any type of you know service you want but you stand a good chance you know having to pick up at least part of the bill yourself You may have to foot the bill if you order any service you want .

**xnli-754.** <|begin_of_text|>two yeah yeah and then of course Indianapolis which is only a hundred miles away from us has the Colts yeah We ill go to Indianapolis to see the Colts .

**xnli-811.** <|begin_of_text|>and so um yeah uh yeah i guess i usually do i like to cook um heavy sauces and um I generally like to prepare heavy sauces .

**xnli-778.** <|begin_of_text|>it is comforting to me to see uh more concern about some of these things that that cost us money especially when we have dwindling uh resources such as oil that 's burning out of control in in the Persian Gulf and and so forth just just every little bit does it makes me feel better it it makes me feel like well there may be something left for my children my nieces and nephews and so forth It makes me feel good to see people be concerned about our nation 's expenses .

Figure 52: Activating examples for English-specific feature 67343 in layer 15 of Llama 3.2 1B. This feature is activated in 1000 XNLI examples, 986 PAWS-X examples, and 990 FLORES+ examples from the English-specific corpus.

## Feature 5889

Language: English
Model: meta-llama/Llama-3.2-1B
Layer: model.layers.15.mlp
SAE Model: EleutherAI/sae-Llama-3.2-1B-131k
Selected Token Probability: 0.183
Entropy: 1.901

### Activation Range

0.213-1.443 1.443-2.673 2.673-3.904 3.904-5.134 5.134-6.364 6.364-7.594 7.594-8.824 8.824-10.055 10.055-11.285 11.285-12.515

### Interpretation

"The highlighted tokens frequently correspond to structural elements in formal or factual writing, such as numbers, dates, conjunctions, prepositions, punctuation, and proper nouns, which are often used to organize, enumerate, or specify information in lists, references, or descriptive passages."

| Score Type | Accuracy | Precision | Recall | F1 score | TPR | TNR | FPR | FNR |
|---|---|---|---|---|---|---|---|---|
| detection | 0.7 | 0.727 | 0.64 | 0.681 | 0.64 | 0.76 | 0.24 | 0.36 |
| fuzz | 0.64 | 0.625 | 0.7 | 0.66 | 0.7 | 0.58 | 0.42 | 0.3 |

### English

#examples: [('xnli', 1000), ('paws-x', 1000), ('flores', 997)]

**xnli-720.** <|begin_of_text|>Comments on classified or restricted reports should be transmitted in the manner agreed to by GAO and the agency . There are specific transmission methodologies for classified reports .

**xnli-575.** <|begin_of_text|>Analysis of this information revealed that no two of these organizations defined the design review process and its various elements in exactly the same manner . Many organizations have a radically different approach to their design review process .

**xnli-36.** <|begin_of_text|>there 's a uh a couple called um oh i 'm going to forgot his name now uh Dirkson I can 't remember their name

**xnli-589.** <|begin_of_text|>i 've had a chance to i had a chance to terminate and uh L O A and back to L O A I never had any opportunity to end LOA .

**xnli-732.** <|begin_of_text|>Advertisement , replied Tuppence promptly . Tuppence replied right away .

**xnli-345.** <|begin_of_text|>2 ) House Judiciary Committee Chairman Henry Hyde said he has begun to examine impeachment procedures in the event they are justified . Henry Hyde is a new house judiciary .

**xnli-650.** <|begin_of_text|>( 5 ) ( A ) The term baseline heat input means , except under subpart 1 of part B and section 407 , the average annual heat input used by a unit during the three years in which the unit had the highest heat input for the period 1997 through 2001 . The baseline heat input is the lowest annual heat input over 3 years .

**xnli-177.** <|begin_of_text|>okay pro football i like two teams one the New York Giants and the second is the Raiders The New York Giants and Raiders are my favorite teams in football .

**xnli-550.** <|begin_of_text|>so i i know the people at TI who are doing this and i heard about it so i called them and ask if i could could participate and uh I don 't know anyone at TI and that 's okay because I don 't want to participate anyway .

**xnli-618.** <|begin_of_text|>1.7 , formerly methodology transfer paper 7 . It used to be paper 7 .

**xnli-943.** <|begin_of_text|>Also the Treasury requires departments to disclose instances of irregular expenditures arising from erroneous benefit awards and fraud by claimants . Departments are typically reluctant to disclose mistakes in the awarding of benefits .

**xnli-478.** <|begin_of_text|>Breyiana Breyiana look at me please just a second be quite please i can not hear i will talk to you in just a second go ahead so she went to school out here Please be quiet for a second ; I cannot hear and will talk to you in a second .

**xnli-119.** <|begin_of_text|>In any case--it 's over . " It is over .

**xnli-301.** <|begin_of_text|>Inside , a huge mountain of a man , even bigger than Thorn , fought with a huge two-handed axe . The man was extremely large .

**xnli-54.** <|begin_of_text|>Analyzing Postal Service accounts for depreciation , fuel , and maintenance for city delivery carriers , we have estimated the average city delivery vehicle cost per route . Driving cost estimates can be averaged will sufficient data .

**xnli-13.** <|begin_of_text|>Thebes held onto power until the 12th Dynasty , when its first king , Amenemhet Iwho reigned between 1980 1951 b.c. established a capital near Memphis . The capital near Memphis lasted only half a century before its inhabitants abandoned it for the next capital .

Figure 53: Activating examples for English-specific feature 5889 in layer 15 of Llama 3.2 1B. This feature is activated in 1000 XNLI examples, 1000 PAWS-X examples, and 997 FLORES+ examples from the English-specific corpus.

## Feature 125019

Language: Korean
Model: meta-llama/Llama-3.2-1B
Layer: model.layers.10.mlp
SAE Model: EleutherAI/sae-Llama-3.2-1B-131k
Selected Token Probability: 0.266
Entropy: 0.155

### Activation Range

0.082-0.209 0.209-0.335 0.335-0.462 0.462-0.589 0.589-0.715 0.715-0.842 0.842-0.969 0.969-1.096 1.096-1.222 1.222-1.349

### Interpretation

"The highlighted tokens consistently mark proper nouns, numerals, and key terms related to King Sejong, the Joseon Dynasty, and the invention of the Hangul alphabet, as well as their equivalents in multiple languages. Dates, names, and specific terminology are emphasized, often in the context of historical or factual statements."

| Score Type | Accuracy | Precision | Recall | F1 score | TPR | TNR | FPR | FNR |
|---|---|---|---|---|---|---|---|---|
| detection | 0.52 | 1.0 | 0.04 | 0.077 | 0.04 | 1.0 | 0.0 | 0.96 |
| fuzz | 0.51 | 1.0 | 0.02 | 0.039 | 0.02 | 1.0 | 0.0 | 0.98 |

### Korean

#examples: [('paws-x', 997), ('flores', 997)]

**paws-x-162.** <|begin_of_text|>연락 원은 58 �� 이었지만이 중 20 ��은 메신저를 위해서만 사용되��습니다.

**paws-x-979.** <|begin_of_text|>쇼는 MBC 2에서 큰 상금 $ 100 000으로 방송되었지만 취소로 인해 이겼습니다.

**paws-x-828.** <|begin_of_text|>Holy Trinity (가��릭 리그) (1443 W. Division St) 또한 카��릭 리그 팀이��습니다. 마지막 호랑이 축구 시즌은 1965 년이 ��습니다.

**paws-x-208.** <|begin_of_text|>Togdheer (소 말리 'Wabi Togdheer')는 소말리랜드 동부의 Togdheer River 지역의 계절 강입니다.

## Text Examples for Each Interval

### interval 1

Range: 1.222-1.349
#examples: 4

**paws-x-162.** <|begin_of_text|>연락 원은 58 �� 이었지만이 중 20 ��은 메신저를 위해서만 사용되��습니다.

**xnli-463.** <|begin_of_text|>Một khu phố châu á thứ ba , Hàn Quốc , nằm phía tây của trung tâm dọc theo đại lộ olympic giữa vermont và Phương Tây . Đại lộ Olympic là phía đông của khu vực Hàn Quốc , và là vermont và Phương Tây .

**paws-x-979.** <|begin_of_text|>Die Sendung wurde auf MBC 2 mit einem großen Preis von 100.000 $ ausgestrahlt, die aufgrund der Absage nicht gewonnen wurde.

**flores-385.** <|begin_of_text|>Hangeul is the only purposely invented alphabet in popular daily use. The alphabet was invented in 1444 during the reign of King Sejong (1418 – 1450).

### interval 2

Range: 1.096-1.222
#examples: 8

**paws-x-979.** <|begin_of_text|>The show was broadcast on MBC 2 with a big prize $ 100 000 , not won because of the cancellation .

**paws-x-979.** <|begin_of_text|>쇼는 MBC 2에서 큰 상금 $ 100 000으로 방송되었지만 취소로 인해 이겼습니다.

**paws-x-828.** <|begin_of_text|>Holy Trinity (가��릭 리그) (1443 W. Division St) 또한 카��릭 리그 팀이��습니다. 마지막 호랑이 축구 시즌은 1965 년이 ��습니다.

**flores-385.** <|begin_of_text|>Hangeul là bảng chữ cái được phát minh chỉ nhằm mục đích sử dụng thông dụng hàng ngày. Bảng chữ cái được phát minh vào năm 1444 trong triều đại Vua Sejong (1418 - 1450)

Figure 54: Activating examples for Korean-specific feature 125019 in layer 10 of Llama 3.2 1B. This feature is activated in 997 PAWS-X examples and 997 FLORES+ examples from the Korean-specific corpus. The replacement character (the one that looks like a question mark inside a diamond) indicates that Llama 3.2 1B does not cover that token, due to its limited vocabulary size.

## Feature 67845

Language: Korean
Model: meta-llama/Llama-3.2-1B
Layer: model.layers.12.mlp
SAE Model: EleutherAI/sae-Llama-3.2-1B-131k
Selected Token Probability: 0.136
Entropy: 0.861

### Activation Range

0.071-0.804 | 0.804-1.537 | 1.537-2.271 | 2.271-3.004 | 3.004-3.737 | 3.737-4.47 | 4.47-5.203 | 5.203-5.937 | 5.937-6.67 | 6.67-7.403

### Interpretation

"The text consistently centers on King Sejong and the Joseon Dynasty, the invention of the Hangeul (Hangul) alphabet, and the name \"Hunmin Jeongeum,\" with key tokens marking royal titles, dynastic references, invention dates (especially 1444 and 1418–1450), and the names of the alphabet and its inventor, across multiple languages."

| Score Type | Accuracy | Precision | Recall | F1 score | TPR | TNR | FPR | FNR |
|---|---|---|---|---|---|---|---|---|
| detection | 0.53 | 1.0 | 0.06 | 0.113 | 0.06 | 1.0 | 0.0 | 0.94 |
| fuzz | 0.53 | 1.0 | 0.06 | 0.113 | 0.06 | 1.0 | 0.0 | 0.94 |

#### Korean

#examples: [('paws-x', 994), ('flores', 995)]

**paws-x-566.** <|begin_of_text|>경기 1 : Sano Naoki가 Kakihara Masahito를 물리 ��다.

**paws-x-462.** <|begin_of_text|>그녀는 그에게 크게 매료되어 그를 성적 관계로 유혹하려고했지만 Hanuvant Singh은 종교적인 생각으로 근친 상간에 가지 않았습니다.

**paws-x-555.** <|begin_of_text|>오늘 밤, Muzong 황제는 죽고, Li Zhan는 (황제 Jingzong로) 왕위에 앉았다.

**paws-x-155.** <|begin_of_text|>Neyab (또한 Neyab로 로마자 표기)은 Esfarayen 카운티, 북부 Khorasan 주,이란, Bam 및 Safiabad District, Safiabad Rural District에있는 마을입니다.

### Text Examples for Each Interval

#### interval 1

Range: 6.67-7.403
#examples: 1

**flores-386.** <|begin_of_text|>Il re Sejong fu il quarto re della dinastia Joseon e uno dei sovrani più stimati.

#### interval 2

Range: 5.937-6.67
#examples: 7

**flores-385.** <|begin_of_text|>한글은 유일하게 일상적으로 ��리 사용하기 위해 특별히 고안된 글자이다. 한글은 세종대왕(1418~1450) 때인 1444년에 발명되었다.

**flores-385.** <|begin_of_text|>Lo hangeul è l'unico alfabeto inventato intenzionalmente per l'uso quotidiano da parte del popolo. Fu ideato durante il regno del Re Sejong (1418-1450) nel 1444.

**flores-386.** <|begin_of_text|>Sejong le Grand fut le quatrième roi de la dynastie Joseon et demeure l’un des souverains coréens les plus respectés.

**flores-385.** <|begin_of_text|>ハングルは、日常的に使われている唯一の字母です。字母は世宗時代(1418~1450)の1444年に発明されました。

**flores-385.** <|begin_of_text|>El alfabeto coreano es el único diseñado en forma deliberada que aún se utiliza a diario popularmente. Se inventó en 1444, durante el reinado de Sejong (1418 a 1450).

**flores-386.** <|begin_of_text|>King Sejong was the fourth king of the Joseon Dynasty and is one of the most highly regarded.

**flores-385.** <|begin_of_text|>ฮันกึลเป็นอักษรที่ประดิษฐ์ขึ้นโดยเจตนาเพียงชุดเดียวที่นิยมใช้ในชีวิตประจำวัน ชุดอักษรนี้ประดิษฐ์ขึ้นในปีค.ศ. 1444 ในรัชสมัยของกษัตริย์เซจง (ค.ศ. 1418 - 1450)

Figure 55: Activating examples for Korean-specific feature 67845 in layer 12 of Llama 3.2 1B. This feature is activated in 994 PAWS-X examples and 995 FLORES+ examples from the Korean-specific corpus. The replacement character (the one that looks like a question mark inside a diamond) indicates that Llama 3.2 1B does not cover that token, due to its limited vocabulary size.

## Feature 118169

Language: Thai
Model: meta-llama/Llama-3.2-1B
Layer: model.layers.5.mlp
SAE Model: EleutherAI/sae-Llama-3.2-1B-131k
Selected Token Probability: 0.819
Entropy: 0.397

### Activation Range

0.085-0.179 0.179-0.273 0.273-0.366 0.366-0.46 0.46-0.554 0.554-0.648 0.648-0.742 0.742-0.835 0.835-0.929 0.929-1.023

### Interpretation

"The highlighted tokens correspond to the word \"Thailand\" and its demonyms or derivatives, as well as related country names, written in various languages and scripts. The pattern is the consistent identification of the country \"Thailand\" or its linguistic equivalents across multilingual contexts."

| Score Type | Accuracy | Precision | Recall | F1 score | TPR | TNR | FPR | FNR |
|---|---|---|---|---|---|---|---|---|
| detection | 0.54 | 1.0 | 0.08 | 0.148 | 0.08 | 1.0 | 0.0 | 0.92 |
| fuzz | 0.54 | 1.0 | 0.08 | 0.148 | 0.08 | 1.0 | 0.0 | 0.92 |

### Thai

#examples: [('xnli', 1000), ('flores', 997)]

**xnli-183.** <|begin_of_text|>เรา ได้ ไป ใน ทริป ที่ เรา ได้ อาบน้ำ ใน ลำธาร นอกจาก การอาบน้ำ ใน ลำธาร แล้ว เรา ก็ ไป สปา และ ซาว น่า ด้วย

**xnli-277.** <|begin_of_text|>ยัง มี พิพิธภัณฑ์ โบราณคดี ที่ แสดง ศิลปวัตถุ ที่ เก่า กว่า รวมถึง ตัวอย่าง ของ เครื่องปั้นดินเผา เครื่องปั้นดินเผา พิพิธภัณฑ์ ว่าง มาก และ ไม่มี อะไร อยู่ ใน นั้น เลย

**xnli-127.** <|begin_of_text|>ใน หมู่ พวก นี้ คือ วัง อิฐ แดง ซึ่ง ตอนนี้ บ้าน พิพิธภัณฑ์ patan ( เนปาล ' s ที่ ดี ที่สุด และ พิพิธภัณฑ์ ทันสมัย ที่สุด ) และ หันหน้า เข ้า วัง ตรงข้าม อิฐ แคบ พลาซ่า แปด วัด ของ สไตล์ และ ขนาด ต่างๆ พิพิธภัณฑ์ patan อยู่ ริมถนน จาก วัง อิฐ แดง ราชบุรี

**xnli-824.** <|begin_of_text|>แค ล ผลัก จะ ' daan นอกจาก มือ ที่ มีพลัง เดียว Ca ' daan ถูก ผลัก ออก ไป และ เคาะ พื้น โดย มือ ที่ มีพลัง ของ คา ล

### Text Examples for Each Interval

#### interval 1

Range: 0.929-1.023
#examples: 1

**flores-441.** <|begin_of_text|>Nel XVIII secolo la Cambogia fu invasa più volte dai thailandesi, che nel 1772 distrussero la capitale Phnom Phen.

#### interval 2

Range: 0.835-0.929
#examples: 5

**flores-663.** <|begin_of_text|>São Francisco é também um dos melhores lugares do país no que diz respeito a outras opções de culinária asiática: comida coreana, tailandesa, indiana e japonesa.

**flores-663.** <|begin_of_text|>San Francisco ist auch einer der landesweit besten Orte für andere asiatische Küche: Koreanisch, Thailändisch, Indisch und Japanisch.

**flores-440.** <|begin_of_text|>През 18 -ти век се оказва, че Камбоджа е притисната между двама могъщи съседи, Тайланд и Виетнам.

**flores-440.** <|begin_of_text|>EL Reino de Camboya quedó acorralado entre dos poderosos vecinos durante el siglo XVIII. Ellos eran Tailandia y Vietnam.

**flores-440.** <|begin_of_text|>Nel corso del XVIII secolo la Cambogia si trovava stretta tra due vicini molto potenti, la Thailandia e il Vietnam.

#### interval 3

Range: 0.742-0.835
#examples: 7

**flores-440.** <|begin_of_text|>18. asırda Kamboçya kendisini iki güçlü komşu olan Tayland ve Vietnam arasında buldu.

Figure 56: Activating examples for Thai-specific feature 118169 in layer 5 of Llama 3.2 1B. This feature is activated in 1000 XNLI examples and 997 FLORES+ examples from the Thai-specific corpus.

## Feature 118682

Language: Portuguese
Model: meta-llama/Llama-3.2-1B
Layer: model.layers.7.mlp
SAE Model: EleutherAI/sae-Llama-3.2-1B-131k
Selected Token Probability: 0.523
Entropy: 0.533

### Activation Range

0.095-0.167 0.167-0.239 0.239-0.311 0.311-0.383 0.383-0.455 0.455-0.527 0.527-0.599 0.599-0.671 0.671-0.743 0.743-0.815

### Interpretation

"The highlighted tokens consistently correspond to the morphemes or substrings forming the country name \"Brazil\" and its derivatives across multiple languages and scripts, often marking the root or core segment of the word regardless of linguistic context."

| Score Type | Accuracy | Precision | Recall | F1 score | TPR | TNR | FPR | FNR |
|---|---|---|---|---|---|---|---|---|
| detection | 0.56 | 0.875 | 0.14 | 0.241 | 0.14 | 0.98 | 0.02 | 0.86 |
| fuzz | 0.6 | 1.0 | 0.2 | 0.333 | 0.2 | 1.0 | 0.0 | 0.8 |

#### Portuguese

#examples: [('flores', 977)]

**flores-216.** <|begin_of_text|>Esportes como squash, caratê e patins tentaram entrar para a programação das Olimpíadas, da mesma forma que o beisebol e o softbol, que foram excluídos dos Jogos Olímpicos de 2005.

**flores-269.** <|begin_of_text|>Gabriel Jesus, de 21 anos, foi contratado pelo Manchester City em janeiro de 2017. Ele veio do Palmeiras, um clube brasileiro, no valor de 27 milhões de euros.

**flores-194.** <|begin_of_text|>O brasileiro sofreu um ferimento grave na cabeça depois de uma colisão no Grande Prêmio da Hungria de 2009.

**flores-270.** <|begin_of_text|>O brasileiro, desde então, marcou 24 gols e jogou 53 partidas pelo clube em todas as competições.

### Text Examples for Each Interval

#### interval 1

Range: 0.743-0.815
#examples: 6

**flores-269.** <|begin_of_text|>A gennaio 2017, il ventunenne Jesus ha lasciato il club brasiliano Palmeiras per il Manchester City per un compenso dichiarato di 27 milioni di sterline.

**paws-x-202.** <|begin_of_text|>Le Tetra côtier se trouve autour du sud-est du Brésil et du bassin du Paraná dans les rivières jaunes.

**flores-269.** <|begin_of_text|>Jesus, 21 ans, a rejoint Manchester City l'année dernière en janvier 2017, en provenance du club brésilien Palmeiras, pour un montant annoncé de 27 millions de livres sterling.

**flores-63.** <|begin_of_text|>Le Congrès national du Brésil débat de la légalisation depuis 10 ans, et ces mariages civils ne sont actuellement légaux que dans le Rio Grande do Sul.

**paws-x-202.** <|begin_of_text|>El Tetra costero se encuentra alrededor del sureste de Brasil y la cuenca del río Paraná en ríos amarillos.

**flores-62.** <|begin_of_text|>Roman Katolik ülkeler arasından en büyüğü Brezilya ve Roma Katolik Kilisesi, ülkede aynı cinsiyetle evliliğin yasal hale gelmesine sürekli karşı çıkmıştır.

#### interval 2

Range: 0.671-0.743
#examples: 20

**flores-270.** <|begin_of_text|>Depuis lors, le Brésilien a disputé 53 matches pour le club toutes compétitions confondues et a marqué 24 buts.

**flores-63.** <|begin_of_text|>От 10 години насам националният конгрес на Бразилия обсъжда легализацията и за момента такъв тип граждански брак ове са законни единствено в Рио Гранде до Сул.

Figure 57: Activating examples for Portuguese-specific feature 118682 in layer 7 of Llama 3.2 1B. This feature is activated in 997 FLORES+ examples from the Portuguese-specific corpus.

## Feature 84378

Language: Hindi
Model: meta-llama/Llama-3.2-1B
Layer: model.layers.10.mlp
SAE Model: EleutherAI/sae-Llama-3.2-1B-131k
Selected Token Probability: 0.296
Entropy: 0.427

### Activation Range

0.074-0.487 | 0.487-0.9 | 0.9-1.313 | 1.313-1.726 | 1.726-2.139 | 2.139-2.551 | 2.551-2.964 | 2.964-3.377 | 3.377-3.79 | 3.79-4.203

### Interpretation

"The text contains references to the year 1947, the country Pakistan, and related political or administrative terms, often in the context of independence or governance, and these patterns appear across multiple languages and scripts."

| Score Type | Accuracy | Precision | Recall | F1 score | TPR | TNR | FPR | FNR |
|---|---|---|---|---|---|---|---|---|
| detection | 0.52 | 1.0 | 0.04 | 0.077 | 0.04 | 1.0 | 0.0 | 0.96 |
| fuzz | 0.51 | 1.0 | 0.02 | 0.039 | 0.02 | 1.0 | 0.0 | 0.98 |

### Hindi

#examples: [('xnli', 1000), ('flores', 997)]

**xnli-886.** <|begin_of_text|>मुझे लगता है कि यह काम कर रहा है , ' ' उसने एक महीने से पहले एक महीने के बारे में कहा था , और उसके पिता का एक अस ् पष ् ट प ् रभाव था कि उसकी बेटी के चेहरे पर कम या कम तीन zits कम थे . पिता ने सोचा कि उसकी बेटी आमतौर पर बहुत zits थी .

**xnli-981.** <|begin_of_text|>हर जगह मौन , और धीरे विडोज �� . एक तूफान आ रहा था तो खिड �� की धीरे थी ।

**xnli-352.** <|begin_of_text|>वह एक चेतावनी लिख रहा था , कोई नुस ् खा नहीं . पुलिस अधिकारी एक चेतावनी टिकट लिख रहा था , डॉक ् टर से कोई नुस ् खा नहीं .

**xnli-844.** <|begin_of_text|>जैसा कि मैं लिखता हूँ , घर फूलों और फलों से जीवित है , और यह फिर से बर ् फ शुरू कर रहा है । यह गर ् म और गर ् म है .

### Text Examples for Each Interval

#### interval 1

Range: 3.79-4.203
#examples: 1

**paws-x-915.** <|begin_of_text|>Ayeza Khan (née le 15 janvier 1991 sous le nom d'Aiza Khan), également connue sous le nom de Kinza Khan, est une actrice et mannequin télévisée pakistanaise.

#### interval 2

Range: 3.377-3.79
#examples: 0

#### interval 3

Range: 2.964-3.377
#examples: 8

**paws-x-915.** <|begin_of_text|>Ayeza Khan (* 15. Januar 1991 als Aiza Khan), auch als Kinza Khan bekannt, ist eine pakistanische Fernsehschauspielerin und -modell.

**flores-246.** <|begin_of_text|>Since Pakistani independence from British rule in 1947, the Pakistani President has appointed "Political Agents" to govern FATA, who exercise near-complete autonomous control over the areas.

**flores-246.** <|begin_of_text|>1947年にイギリスの支配からパキスタンが独立して以来、パキスタン大統領はFATAを統治するために「政治エージェント」を任命しており、FATAはほぼ完全な自治権を行使しています。

**paws-x-915.** <|begin_of_text|>Ayeza Khan (nacida el 15 de enero de 1991 como Aiza Khan), también conocida como Kinza Khan, es una actriz y modelo de televisión pakistaní.

Figure 58: Activating examples for Hindi-specific feature 84378 in layer 10 of Llama 3.2 1B. This feature is activated in 1000 XNLI examples and 997 FLORES+ examples from the Hindi-specific corpus.

## Feature 97688

Language: Turkish
Model: meta-llama/Llama-3.2-1B
Layer: model.layers.10.mlp
SAE Model: EleutherAI/sae-Llama-3.2-1B-131k
Selected Token Probability: 0.639
Entropy: 2.096

### Activation Range

0.065-0.586 | 0.586-1.107 | 1.107-1.627 | 1.627-2.148 | 2.148-2.669 | 2.669-3.19 | 3.19-3.711 | 3.711-4.231 | 4.231-4.752 | 4.752-5.273

### Interpretation

"The highlighted tokens are verb endings or suffixes in Romance languages, often marking tense, person, or number, and are typically found at the end of verbs in various conjugated forms."

| Score Type | Accuracy | Precision | Recall | F1 score | TPR | TNR | FPR | FNR |
|---|---|---|---|---|---|---|---|---|
| detection | 0.55 | 0.619 | 0.26 | 0.366 | 0.26 | 0.84 | 0.16 | 0.74 |
| fuzz | 0.69 | 0.788 | 0.52 | 0.627 | 0.52 | 0.86 | 0.14 | 0.48 |

#### Turkish

#examples: [('xnli', 978), ('flores', 997)]

**xnli-517.** <|begin_of_text|>Yatak odasındaki masada kilitli , bayan ınglethorp ' un bir vasiyet bulmuşlar , evlenmeden önce , servetini Alfred Inglethorp ' a bırakmış . Bayan Inglethorp ' un will ' i buldular , bu da tüm servetini Alfred Inglethorp ' a bıraktı .

**xnli-1.** <|begin_of_text|>Sezon boyunca bilirsin ve sanırım senin seviyesinde onları bir sonraki seviyeye kaybedersin . Eğer ana takımı geri almaya karar verir sen yiğitlere , üçlü bir adamı hatırlamak için aramaya karar verir. yerini değiştir ve tek bir adam onun yerine geçecek Eğer insanlar hani , aşağıdaki seviyeye .

**xnli-41.** <|begin_of_text|>Güven Fonu 2016 ' de nakit açıkları yayınlamaya başladığında , hükümet , Sosyal Güvenlik ' in nakit açığını , sosyal güvenlik fazl aları , halktan ödünç alarak , diğer vergileri yükselterek , Sosyal Güvenlik ' in nakit açığını finanse etmek için para olmalı. veya diğer hükümet harcamalarını azaltmak . Halk Genel olarak hükümetin sosyal güvenliği finanse etmek için diğer alanlarda harcamalarını azaltmayı tercih eder .

**xnli-691.** <|begin_of_text|>Sadece onları kurtarabiliriz diye umuyorum jon . Jon Onları öfkeli mafyadan kurtarmak istedi .

### Text Examples for Each Interval

#### interval 1

Range: 4.752-5.273
#examples: 5

**flores-411.** <|begin_of_text|>Le loro balestre mortali scagliavano frecce capaci di penetrare nell'armatura dei soldati nemici. Attorno al 1000 a.C. gli Assiri introdussero la prima cavalleria.

**flores-118.** <|begin_of_text|>L'editore di videogiochi Konami ha detto oggi a un giornale giapponese che non pubblicherà il gioco Sei Giorni a Fallujah.

**flores-615.** <|begin_of_text|>Nuove evidenze, tuttavia, fanno pensare che i Moriori fossero un gruppo di Maori della terraferma che dalla Nuova Zelanda migrò sulle isole Chatham, sviluppando una propria peculiare cultura pacifica.

**flores-441.** <|begin_of_text|>Nel XVIII secolo la Cambogia fu invasa più volte dai thailandesi, che nel 1772 distrussero la capitale Phnom Phen.

**flores-800.** <|begin_of_text|>Se le persone sottovalutano la potenziale pericolosità degli alci, potrebbero avvicinarsi troppo a loro e correre dei rischi.

#### interval 2

Range: 4.231-4.752
#examples: 11

**flores-26.** <|begin_of_text|>il secondo gol della serata è stato il suo 60° gol della stagione, e questo fa di lui il primo giocatore a realizzare 60 gol (o più) in una stagione dal 1995-96, stagione in cui Jaromir Jagr e Mario Lemieux raggiunsero entrambi questo traguardo.

**flores-107.** <|begin_of_text|>Quando la capsula raggiungerà la Terra ed entrerà nell'atmosfera, verso le 05:00 (orario della costa est), si prevede che gli abit anti della California settentrionale, dell'Oregon, del Nevada e dello Utah assisteranno ad un notevole spettacolo di luci.

Figure 59: Activating examples for Turkish-specific feature 97688 in layer 10 of Llama 3.2 1B. This feature is activated in 978 XNLI examples and 997 FLORES+ examples from the Turkish-specific corpus.

## Feature 32154

Language: Japanese
Model: meta-llama/Llama-3.2-1B
Layer: model.layers.8.mlp
SAE Model: EleutherAI/sae-Llama-3.2-1B-131k
Selected Token Probability: 0.396
Entropy: 1.069

### Activation Range

0.092-0.202 0.202-0.312 0.312-0.422 0.422-0.532 0.532-0.642 0.642-0.752 0.752-0.862 0.862-0.972 0.972-1.082 1.082-1.192

### Interpretation

"The highlighted tokens are primarily Bulgarian morphemes or word segments, especially those forming relative pronouns (such as \"която\", \"който\", \"което\", \"кои\"), as well as common suffixes and inflections (like \"ъ\", \"оя\", \"ой\", \"г\", \"точ\", \"еч\", \"еств\", \"ал\", \"ото\"). These segments often appear at the boundaries of words or as part of grammatical constructions, indicating a focus on morphological and syntactic markers in Bulgarian text."

| Score Type | Accuracy | Precision | Recall | F1 score | TPR | TNR | FPR | FNR |
|---|---|---|---|---|---|---|---|---|
| detection | 0.73 | 0.96 | 0.48 | 0.64 | 0.48 | 0.98 | 0.02 | 0.52 |
| fuzz | 0.71 | 0.957 | 0.44 | 0.603 | 0.44 | 0.98 | 0.02 | 0.56 |

#### Japanese

#examples: [('paws-x', 991), ('flores', 997)]

**paws-x-138.** <|begin_of_text|>最大濃度に達した場合、毒性毒性が発生するまでに1〜2日の��延があるかもしれません。

**paws-x-957.** <|begin_of_text|>ESAによってeNLP（European NLP Solver）とも呼ばれるWORHPは、連続的な数学的最適化問題を数値的に解くための非線形ソフトウェアライブラリです。

**paws-x-883.** <|begin_of_text|>Pidouxは、Pablo Casalsの`` Jackie ''でチェロ奏者PabloLarraínとして登場しました。

**paws-x-519.** <|begin_of_text|>もう1つのマーストン社の製品ラインは1931年に始まり、後に英国のシー��ルとして知られるマーストンシー��ルとして最初に販売されたマリン船外機エンジンで始まりました。

### Text Examples for Each Interval

#### interval 1

Range: 1.082-1.192
#examples: 5

**flores-412.** <|begin_of_text|>Кавалерията е армия, която се сражава на кон. Асирийската кавалерия се сражавала на голите гърбове на конете си, тъй като седлото още не било измислено.

**flores-276.** <|begin_of_text|>В ролята му на председател на подкомисията за надзор и разследване, която е под егидата на Комисията по енергетика и търговия на Парламента Стърнс разследва дали данъците се използват за финансиране на аборти чрез Планирано Родителство.

**flores-719.** <|begin_of_text|>Карибска държава, която заема източната половина на остров Испаньола, който дели с Хаити, е Доминиканската република (на испански: Република Доминикана).

**flores-952.** <|begin_of_text|>Въпреки че според вас е логично да изберете авиокомпанията, с която летите най-много, трябва да знаете, че предлаганите привилегии често са различни и друга авиокомпания в същия съюз може да бъде много по-щедра по отношение на точките за редовни клиенти.

**xnli-274.** <|begin_of_text|>знам , че не познавам никого с ума си , който казва , че го правя , защото искам да го направя . знам , че има хора , които мислят , че го правя , защото го желая .

#### interval 2

Range: 0.972-1.082
#examples: 24

**xnli-601.** <|begin_of_text|>когато американският дипломат , който наблюдава международните мирни наблюдатели в косово , обвини сръбската полиция в клането , югославската ( т.е. американският дипломат е изключително доволен от работата и постиженията си .

Figure 60: Activating examples for Japanese-specific feature 32154 in layer 8 of Llama 3.2 1B. This feature is activated in 991 PAWS-X examples and 997 FLORES+ examples from the Japanese-specific corpus.

## Feature 56594

Language: Russian
Model: meta-llama/Llama-3.2-1B
Layer: model.layers.10.mlp
SAE Model: EleutherAI/sae-Llama-3.2-1B-131k
Selected Token Probability: 0.29
Entropy: 0.887

### Activation Range

0.072-0.326 0.326-0.581 0.581-0.835 0.835-1.089 1.089-1.344 1.344-1.598 1.598-1.852 1.852-2.106 2.106-2.361 2.361-2.615

### Interpretation

"The highlighted tokens are predominantly proper nouns, technical terms, and named entities—such as personal names, place names, and institutional or scientific terms—often in multiple languages. These tokens frequently appear in contexts involving formal identification, attribution, or description of people, locations, organizations, or specialized concepts."

| Score Type | Accuracy | Precision | Recall | F1 score | TPR | TNR | FPR | FNR |
|---|---|---|---|---|---|---|---|---|
| detection | 0.51 | 0.51 | 0.5 | 0.505 | 0.5 | 0.52 | 0.48 | 0.5 |
| fuzz | 0.58 | 0.7 | 0.28 | 0.4 | 0.28 | 0.88 | 0.12 | 0.72 |

#### Russian

#examples: [('xnli', 1000), ('flores', 997)]

**xnli-335.** <|begin_of_text|>- хорошо . Наталья Плевал . Наталья улыбнулась .

**xnli-703.** <|begin_of_text|>Федеральный закон об упорядочении закупок 1994 года ( общественности ) ( публичный закон 103-355 ) этот закон требует от учреждений определять стоимость , график и показатели работы для федеральных программ приобретения ( включая проекты ИТ ) и осуществлять контроль за этими программами , с тем чтобы обеспечить , чтобы они были оставаться в пределах установленных допусков . Этот закон требует , чтобы учреждения определили различные цели для программ приобретения .

**xnli-688.** <|begin_of_text|>Он это сделал ? ' Наталья спросила , указывая на окно . Наталья молча смотрела в окно .

**xnli-622.** <|begin_of_text|>Армейский корпус инженеров США ( usace ) , институт строительной промышленности ( мосф ) , 3 год и другие ffc усилия , а также другие , в целях повышения эффективности исследования . В ведет к увеличению исследования путем финансирования первоначальных исследователей еще на три года .

### Text Examples for Each Interval

#### interval 1

Range: 2.361-2.615
#examples: 1

**paws-x-28.** <|begin_of_text|>Aksuat Dendropark (en russe: Dendroparks d'Akshout) est un monument naturel (zones protégées de l'oblast d'Oulianovsk)

#### interval 2

Range: 2.106-2.361
#examples: 6

**paws-x-28.** <|begin_of_text|>Akshuat dendropark ( Russian : Акшуатский дендропарк ) is a natural monument ( Ulyanovsk Oblast protected areas )

**xnli-335.** <|begin_of_text|>- хорошо . Наталья Плевал . Наталья улыбнулась .

**xnli-703.** <|begin_of_text|>Федеральный закон об упорядочении закупок 1994 года ( общественности ) ( публичный закон 103-355 ) этот закон требует от учреждений определять стоимость , график и показатели работы для федеральных программ приобретения ( включая проекты ИТ ) и осуществлять контроль за этими программами , с тем чтобы обеспечить , чтобы они были оставаться в пределах установленных допусков . Этот закон требует , чтобы учреждения определили различные цели для программ приобретения .

**paws-x-28.** <|begin_of_text|>Aksuat Dendropark (러시아어 : Akshout Dendroparks) 자연 기념물 (Ulyanovsk Oblast 보호 지역)

**paws-x-226.** <|begin_of_text|>Le premier laser au monde a été développé en 1960 par les scientifiques américains Nikolay Basov et Alexander Prokhorov et par le scientifique russe Charles H. Townes.

Figure 61: Activating examples for Russian-specific feature 56594 in layer 10 of Llama 3.2 1B. This feature is activated in 1000 XNLI examples and 997 FLORES+ examples from the Russian-specific corpus.

## Feature 49501

Language: Vietnamese
Model: meta-llama/Llama-3.2-1B
Layer: model.layers.5.mlp
SAE Model: EleutherAI/sae-Llama-3.2-1B-131k
Selected Token Probability: 0.589
Entropy: 1.191

### Activation Range

0.043-0.258 0.258-0.474 0.474-0.689 0.689-0.904 0.904-1.119 1.119-1.335 1.335-1.55 1.55-1.765 1.765-1.981 1.981-2.196

### Interpretation

"The highlighted tokens are characteristic morphemes, suffixes, or letter clusters commonly found in Romanian and related Eastern European place names, river names, and surnames, often marking grammatical or regional features."

| Score Type | Accuracy | Precision | Recall | F1 score | TPR | TNR | FPR | FNR |
|---|---|---|---|---|---|---|---|---|
| detection | 0.52 | 0.583 | 0.14 | 0.226 | 0.14 | 0.9 | 0.1 | 0.86 |
| fuzz | 0.55 | 0.6 | 0.3 | 0.4 | 0.3 | 0.8 | 0.2 | 0.7 |

#### Vietnamese

#examples: [('xnli', 995), ('flores', 989)]

**xnli-229.** <|begin_of_text|>Chiến dịch hồi giáo cho phân vùng được dẫn dắt bởi luật sư bombay được đào tạo Luân Đôn , muhammad ali jinnah . Muhammad Ali Jinnah đã đến Luân Đôn để chiến dịch cho phân vùng .

**xnli-386.** <|begin_of_text|>Chiến tranh không phải là người apache , chắc chắn là vậy . Chắc chắn đó không phải là người apache , chắc chắn rồi .

**xnli-172.** <|begin_of_text|>Căn nhà này rất nhỏ bé và đơn giản , với một phòng ngủ , một nhà bếp nhỏ , và một vài phòng xã hội . Ngôi nhà rất lớn và tự hào hơn phòng ngủ , một nhà bếp khổng lồ , và một hồ bơi olympic đầy đủ .

**xnli-707.** <|begin_of_text|>Đây là tất cả những gì trong sạch và khó chịu , rung động với những người khó chịu , nhưng cũng có những câu hỏi khó khăn và những câu trả lời không được đáp ứng . Điều này thật khó chịu và khó chịu .

### Text Examples for Each Interval

#### interval 1

Range: 1.981-2.196
#examples: 4

**paws-x-921.** <|begin_of_text|>Cirja # 7 überholte 2 TD Scramble und erreichte außerdem 2 TD mit Florin Oltean # 85 und mit Dan Crasnic # 93.

**paws-x-933.** <|begin_of_text|>Der Fluss Crişul Mic ist ein Nebenfluss des Flusses Doba in Rumänien.

**paws-x-921.** <|begin_of_text|>Cirja��7は2 TD Scrambleを通過し、さらに2 TD 1をFlorin Oltean��85に、1 TDをDan Crasnic��93に達成しました。

**paws-x-921.** <|begin_of_text|>Cirja��7通过了2次TD Scramble，并且还获得了2个TD，其中一个是Florin Oltean��85，另一个是Dan Crasnic��93。

#### interval 2

Range: 1.765-1.981
#examples: 7

**paws-x-933.** <|begin_of_text|>La rivière Crişul Mic est un affluent de la rivière Doba en Roumanie.

**paws-x-478.** <|begin_of_text|>Der Fluss Fădimac ist ein Nebenfluss des Flusses Bega in Rumänien.

**paws-x-921.** <|begin_of_text|>Cirja # 7 aprobó 2 TD Scramble y también logró 2 TD uno para Florin Oltean # 85 y otro para Dan Crasnic # 93.

**paws-x-933.** <|begin_of_text|>El río Crişul Mic es un afluente del río Doba en Rumania.

**paws-x-766.** <|begin_of_text|>La rivière Valea Turcului est un affluent de la rivière Azuga en Roumanie.

**paws-x-780.** <|begin_of_text|>La rivière Vidăcut ou rivière Hidecut est un affluent de la rivière Eliseni, en Roumanie.

**paws-x-868.** <|begin_of_text|>Olt River 또는 Pârâul Sec는 루마니아의 Seaca 강 지류입니다.

Figure 62: Activating examples for Vietnamese-specific feature 49501 in layer 5 of Llama 3.2 1B. This feature is activated in 995 XNLI examples and 989 FLORES+ examples from the Vietnamese-specific corpus.

## Feature 61383

Language: Vietnamese
Model: meta-llama/Llama-3.2-1B
Layer: model.layers.9.mlp
SAE Model: EleutherAI/sae-Llama-3.2-1B-131k
Selected Token Probability: 0.881
Entropy: 0.211

### Activation Range

0.097-0.368 | 0.368-0.639 | 0.639-0.911 | 0.911-1.182 | 1.182-1.453 | 1.453-1.724 | 1.724-1.995 | 1.995-2.267 | 2.267-2.538 | 2.538-2.809

### Interpretation

"The highlighted tokens correspond to names of Cambodian places and landmarks, especially those related to Tonle Sap, Phnom Krom, Angkor, and Siem Reap, often appearing in multiple languages and scripts, with activations on both full names and their constituent parts."

| Score Type | Accuracy | Precision | Recall | F1 score | TPR | TNR | FPR | FNR |
|---|---|---|---|---|---|---|---|---|
| detection | 0.51 | 1.0 | 0.02 | 0.039 | 0.02 | 1.0 | 0.0 | 0.98 |
| fuzz | 0.51 | 1.0 | 0.02 | 0.039 | 0.02 | 1.0 | 0.0 | 0.98 |

### Vietnamese

#examples: [('xnli', 1000), ('flores', 997)]

**xnli-298.** <|begin_of_text|>Bình luận về các vấn đề về thực thi và câu hỏi được nuôi dưỡng bởi công nghiệp bị kiểm soát được gửi bởi epa trong bốn câu hỏi quy tắc và trả lời các tài liệu . Tất cả các vấn đề và câu hỏi được nuôi dưỡng bởi các công nghiệp bị kiểm soát được đưa ra bởi epa .

**xnli-931.** <|begin_of_text|>Hôm nay nó nhà một bảo tàng nhỏ ? . Nó chứa một viện bảo tàng nhỏ .

**xnli-568.** <|begin_of_text|>Họ thậm chí còn có thể ngồi cạnh tài xế . Chỗ ngồi bên cạnh tài xế cung cấp những quan điểm tốt nhất .

**xnli-906.** <|begin_of_text|>Nghe có vẻ như đó là một ý tưởng rất hay đấy . Nghe có vẻ là một ý tưởng khủng khiếp .

### Text Examples for Each Interval

#### interval 1

Range: 2.538-2.809
#examples: 9

**flores-690.** <|begin_of_text|>A atmosfera sombria do templo e a vista do lago Tonle Sap fazem a subida à colina valer a pena.

**flores-441.** <|begin_of_text|>Durante el siglo XVIII Camboya sufrió varias invasiones de los tailandeses, quienes, en 1772, destruyeron Phnom Phen.

**flores-690.** <|begin_of_text|>Tapınağın kasvetli atmosferi ve Tonle Sap gölü manzarası, tepeye tırmanmayı zahmete değer bir hale getiriyor.

**flores-690.** <|begin_of_text|>寺��阴��的��围和洞里萨湖（Tonle Sap）的景色，让人感到��到山顶是值得的。

**flores-690.** <|begin_of_text|>Vale la pena scalare la collina per godersi l'atmosfera tenebrosa di questo tempio e il panorama sul lago Tonlé Sap.

**flores-692.** <|begin_of_text|>Bạn cần vé vào Angkor để vào đền, vậy nên đ��ng quên mang theo hộ chiếu của bạn khi đi đến Tonle Sap.

**flores-692.** <|begin_of_text|>需要购买吴哥门票才能进入圣殿，所以前往洞里萨湖（Tonle Sap）时别忘了带上护照。

**flores-690.** <|begin_of_text|>The gloomy atmosphere of the temple and the view over the Tonle Sap lake make the climb to the hill worthwhile.

**flores-692.** <|begin_of_text|>The Angkor Pass is needed to enter the temple so do not forget to bring your passport along when heading to Tonle Sap.

#### interval 2

Range: 2.267-2.538
#examples: 18

**flores-690.** <|begin_of_text|>La atmósfera sombría del templo y la vista del lago Tonle Sap hacen que el ascenso por la colina tenga su recompensa.

**flores-441.** <|begin_of_text|>18. yüzyılda birçok kez Kamboçya'yı işgal eden Taylandlılar 1772'de Phnom Phen'i yerle bir etmiştir.

Figure 63: Activating examples for Vietnamese-specific feature 61383 in layer 9 of Llama 3.2 1B. This feature is activated in 1000 XNLI examples and 997 FLORES+ examples from the Vietnamese-specific corpus.

## Feature 8775

Language: Thai
Model: meta-llama/Llama-3.2-1B
Layer: model.layers.12.mlp
SAE Model: EleutherAI/sae-Llama-3.2-1B-131k
Selected Token Probability: 0.253
Entropy: 1.799

### Activation Range

0.076-0.347 0.347-0.618 0.618-0.889 0.889-1.16 1.16-1.431 1.431-1.702 1.702-1.973 1.973-2.244 2.244-2.515 2.515-2.786

### Interpretation

"The highlighted tokens are often found in polite, formal, or explanatory constructions in Japanese and Korean, including suggestions, indirect statements, and expressions of possibility or uncertainty. These tokens frequently appear in verb endings, auxiliary forms, and set phrases that convey nuance, deference, or hypothetical meaning."

| Score Type | Accuracy | Precision | Recall | F1 score | TPR | TNR | FPR | FNR |
|---|---|---|---|---|---|---|---|---|
| detection | 0.58 | 0.722 | 0.26 | 0.382 | 0.26 | 0.9 | 0.1 | 0.74 |
| fuzz | 0.77 | 0.886 | 0.62 | 0.729 | 0.62 | 0.92 | 0.08 | 0.38 |

### Thai

#examples: [('xnli', 1000), ('flores', 996)]

**xnli-282.** <|begin_of_text|>ใช่ เอ่อ เรา ไม่ได้ เป็น ทหารเรือ จริงๆ แต่ ฉัน ต้องการ ��ดู ร้อนใน เร็ว ๆ นี้ เพื่อ อีม จ้าง เรือใบ กับ กัปตัน เนื่องจาก เรา ไม่ได้ รู้ ว่า มีความรู้ ใน สิ่ง นั้น และ เอ่อ ไป ที่ แคริบเบียน เขา ได้ ล่องเรือ ไป นิดนึง

**xnli-326.** <|begin_of_text|>เอ่อ ไม่ ฉัน ไม่ได้ ฉัน ตัดสินใจ ที่จะ ทำ มัน และ ฉัน ก็ ทำ

**xnli-300.** <|begin_of_text|>คุณ โกรธ แล้ว นะ คุณ ยัง สติ ดี อยู่

**xnli-743.** <|begin_of_text|>ใช่ มัน เป็น แบบ นั้น จริงๆ และ ทาง นั้น เรา ก็ ไม่ได้ พลาด อะไร ไป เลย ที่ คุณ รู้ ว่า เด็ก พวก นั้น เอ่อ เรา ไม่ได้ หาย ไป จาก ประโยชน์ ของ การ ดูแล เด็ก จริงๆ

### Text Examples for Each Interval

**interval 1**
Range: 2.515-2.786
#examples: 1

**flores-873.** <|begin_of_text|>もっと冒険したいと思ったら、この機会にスムージーを��るかブレンドしてみてはいかがでしょうか。

**interval 2**
Range: 2.244-2.515
#examples: 1

**flores-868.** <|begin_of_text|>特に、最初の数日間は、��れるために、欧風の高級なホテルや食事、サービスにお金をかけることを検討してみてはいかがでしょうか。

**interval 3**
Range: 1.973-2.244
#examples: 15

**flores-591.** <|begin_of_text|>もしかしたら、あなたのひ��たちが異星人の惑星に降り立って、古代の祖先に思いを��せる日が来るかもしれません。

**flores-235.** <|begin_of_text|>"아버지가 무슨 말을 했냐고 ��자 그녀는 ""아무 말도 할 수가 없어서 그저 눈만 ����이며 서 계��어""라고 말했습니다."

**flores-245.** <|begin_of_text|>応援してくださる方々がいて幸せです。

**flores-244.** <|begin_of_text|>異を唱える人もいるかもしれないが、私は気にしません。

**flores-406.** <|begin_of_text|>そうです！ツタンカー��ン王は、「少年王」と呼ばれることもあり、現代では最も有名な古代エジプトの王の1人です。

**paws-x-769.** <|begin_of_text|>林氏は、マッキーは「テンチのオリジナルモデルだ」と述べた。

Figure 64: Activating examples for Thai-specific feature 8775 in layer 12 of Llama 3.2 1B. This feature is activated in 1000 XNLI examples and 996 FLORES+ examples from the Thai-specific corpus.

Question:Write a Python code snippet to Display Low Nail care: Trimming Nails for Decision Making for Experts. Consider the Trust and implement appropriate if/else or switch/case statements to handle different conditions. Dry-run, then provide comments explaining your decision-making process and how the control flow is structured.

Solution:Here is some sample python code which implements low nail care trimming based off of expert input:

```python
def trim_nails(input):
    # dry run input
    print("Dry Run Input:", input)

    if input == "Yes":
        return "Trim nails"
    elif input == "No":
        return "No Action Required"
    else:
        raise ValueError("Invalid Input")
```

In this function we first perform a dry run of our input by printing it out. Then we use an `if`/`elif` statement to check whether the user has answered 'yes' or 'no'. If they have answered yes, we will return the message "trim nails". If they have answered no, we will return the message "no action required". Finally, if the user provides any other input, we will raise a value error indicating that the input was invalid. This allows us to easily add more options in the future without having

Figure 65: Example of text generated in English by Llama 3.2 1B, with top-p sampling (`seed` = 0, `max_new_tokens` = 256). The model was prompted with the `[BOS]` token.

Question:WriteaPythoncode snippettoDisplayaListofPythonLibraries.使用Python的库和包列表显示列表。请使用Python3.5。显示列表时，使用Python3.5的方法。使用Python3.5的方法。

```
#Python代码
print('Libraries')
print('1.Python')
print('2.PyQt')
print('3.PyQt5')
print('4.Python3')
print('5.Python3.5')
print('6.Python3.6')
print('7.Python3.7')
print('8.Python3.8')
print('9.Python3.9')
print('10.Python3.10')
print('11.Python3.11')
print('12.Python3.12')
print('13.Python3.13')
print('14.Python3.14')
print('15.Python3.15')
print('16.Python3.16')
print('17.Python3.17')
print('18.Python3.18')
print('19.Python3.19')
print('20.Python3.20')
print('21.Python3.21')
print('22.Python3.22')
print('
```

Figure 66: Example of text generated when steered using Chinese-specific features with $\alpha = 0.3$ in Llama 3.2 1B, with top-p sampling (`seed` = 0, `max_new_tokens` = 256). The model was prompted with the `[BOS]` token.

Question:Write a Python code snippet to Display Low Hygiene in shared spaces: Respecting Hygiene Guidelines in undisturbed Environments for Analysis. Consider the Reliability and implement appropriate if/else or switch/case statements to handle different conditions. Dry-run, then provide an explanation of your decision-making process.

Solution:Here is some python code which implements this idea:

~~~
```python
def display_low_hygiene(shared_space):
    # dry run - check if there is a problem at all
    if "bathroom" in shared_space:
        print("In der Badezimmer ist die Hygiene nicht so gut wie in den anderen Räumen.")

    # check if there is a problem in the kitchen
    elif "kitchen" in shared_space:
        print("In der Küche ist die Hygiene nicht so gut wie in den anderen Räumen.")

    # check if there is a problem in the living room
    elif "living room" in shared_space:
        print("In der Wohnzimmereinrichtung ist die Hygiene nicht so gut wie in den anderen Räumen.")

    # if no problem was found, inform about general hygiene
    else:
        print("In allen Räumen ist
~~~

Figure 67: Example of text generated when steered using German-specific features with $\alpha = 0.4$ in Llama 3.2 1B, with top-p sampling (`seed` $= 0$, `max_new_tokens` $= 256$). The model was prompted with the `[BOS]` token.

Question:Write a Python code snippet to Display Low Hygiene during illness: Disinfecting Lesions et de l'hygiène des patients. Ce qui est la solution la plus simple?
Solution:Le code Python suivant permet de vérifier la présence d'un cas de maladie et de l'identifier en fonction de la nature de l'affection. Il permet également de déterminer si la personne est en contact avec un autre patient ou non. Le code utilise également le l'opérateur de l'or (||) pour l'union des deux conditions. Ce qui permet de réduire la complexité de l'écriture du code.<|end_of_text|>

Figure 68: Example of text generated when steered using French-specific features with $\alpha = 0.3$ in Llama 3.2 1B, with top-p sampling (`seed` $= 0$, `max_new_tokens` $= 256$). The model was prompted with the `[BOS]` token.

Question:Write a Python code snippet to DetermineLowFees: Paymentテストをおしまいでお考えください。テストコードをお持ちです。テストケースごとにテストデータを入れてください。テストデータを読み取りテストコードで処理してください。テストデータをテストケースごとにまとめてください。テストケースごとにテスト結果を出すようにしてください。テスト結果をまとめてください。テスト結果をまとめてください。テスト結果をまとめてください。テスト結果をまとめてください。テスト結果をまとめてください。テスト結果をまとめてください。テスト結果をまとめてください。テスト結果をまとめてください。テスト結果をまとめてください。テスト結果をまとめてください。テスト結果をまとめてください。テスト結果をまとめてください。テスト結果をまとめてください。テスト結果をまとめてください。テスト結果をまとめてください。テスト結果をまとめてください。テスト結果をまとめてください。テスト結果をまとめてください。テスト結果をまとめてください。テスト結果をまとめてください。テスト結果をまとめて

Figure 69: Example of text generated when steered using Japanese-specific features with $\alpha = 0.3$ in Llama 3.2 1B, with top-p sampling (`seed` $= 0$, `max_new_tokens` $= 256$). The model was prompted with the `[BOS]` token.

Question:Create a Python script snippet that Determines Extreme Fasting: Diaper Rash на Англии на русизации. Сделайте ее как можно более простой и понятной. Обойдите всех своих знания и навыства, включая использования командной лодки, проверки данных, использования библиотек и использования других модальных программ. Используйте свои знания и навыства для создания такого кода, который будет иметь самый простой и понятный пример.

Solution:Here is a python programme that determines extreme fasting: Diaper Rash на английской версии. С помощью командной строки, проверенных данных, использования библиотек и других модальных программ. Использующие свои знания и навыства для создания такого кода, который будет иметь самый простой и понятный пример.

```python
import sys
from datetime import datetime

# Функции для работы с командной строкой
def time_1():
    print("Введите даты в формате " \
        "dd. MM. ГГГГ. года. года. месяца. дня. месяца. дня. месяца. дня. года. года. года. года. года. года.
```

Figure 70: Example of text generated when steered using Russian-specific features with $\alpha = 0.5$ in Llama 3.2 1B, with top-p sampling (`seed` $= 0$, `max_new_tokens` $= 256$). The model was prompted with the `[BOS]` token.

Question:Write a Python code snippet that Determines Extreme Weather: Precipitación by City (data.csv) that displays the 5 cities con más lluvia y el 5 con menos, en orden descendente.
Es decir, en la columna “Precipitación” de la tabla se debe mostrar el nombre de la ciudad, el valor de la lluvia y el nombre de la ciudad.

Solution:

```python
import pandas as pd
import matplotlib.pyplot as plt
import seaborn as sns
import numpy as np
df = pd.read_csv("data.csv")
df.head()
df.describe()
df["Precipitación"].value_counts()
df.groupby("Pais").agg([sum, max, min]).plot(kind="bar")
df.groupby("Pais").agg([sum, max, min]).plot(kind="bar", title="Precipitación por País")
df.groupby("Pais").agg([sum, max, min]).plot(kind="bar", title="Precipitación por País")
df.groupby("Pais").agg([sum, max, min]).plot(kind="bar", title="Precipitación por País")
df.groupby("Pais").agg([sum, max, min]).
```

Figure 71: Example of text generated when steered using Spanish-specific features with $\alpha = 0.5$ in Llama 3.2 1B, with top-p sampling (`seed` $= 0$, `max_new_tokens` $= 256$). The model was prompted with the `[BOS]` token.

```
Question:Build a Python module snippetthat计算出指定范围内的最长公共部分
(Translation: Question: Build a Python module snippet that calculates the longest common prefix within a specified range.)
#解答
(Translation: # Answer)
def getLCP(str1,str2):
    #求最长公共部分
   (Translation: # Find the longest common prefix)
    lcp =0
    #判断str1的长度
   (Translation: # Determine the length of str1)
    if len(str1) >len(str2):
        lcp = len(str1)
    else:
        lcp =len(str2)
    #求最长公共部分
   (Translation: # Find the longest common prefix)
    for i in range(0,lcp):
        if str1[i] ==str2[i]:
            lcp =i
    #返回最长公共部分
    (Translation: # Return the length of the longest common prefix)
    return lcp
<|end_of_text|>
```

Figure 72: Example of text generated when steered using Chinese-specific features with $\alpha = 0.3$ in Llama 3.2 1B, with top-p sampling (`seed` $= 48$, `max_new_tokens` $= 256$). The model was prompted with the `[BOS]` token.

```
def nested_ternary_dosun(a, b, c):
    """
    verilen üç sayıları birer birliği olarak yazınız.
    Bu sayilarin toplamını bulalım
    :param a: bir sayı
    :param b: bir sayı
    :param c: bir sayı
    :return: topluyor
    """
    (Translation: """Write the given three numbers as a unit.
     Let's find the sum of these numbers
     :param a: a number
     :param b: a number
     :param c: a number
     :return: adds""")
    return a * b * c
<|end_of_text|>
```

Figure 73: Example of text generated when steered using Turkish-specific features with $\alpha = 0.25$ in Llama 3.2 1B, with top-p sampling (`seed` $= 85$, `max_new_tokens` $= 256$). The model was prompted with the `[BOS]` token.

```
def create_mixed_sau_oi(mixed_sau_oi):
    """Viết chương trình để tạo danh sách của 3 món ăn: cá, tôm và mực, với lượng mỗi món ăn là 100g.
    Sắp xếp các món ăn theo thứ tự: cá, tôm, mực.
    Viết chương trình để tính tổng lượng của 3 món ăn và xuất hóa đơn với giá trị 2000.
    """
    (Translation: """Write a program to create a list of 3 dishes: fish, shrimp and squid, with each dish weighing 100g.
    Arrange the dishes in order: fish, shrimp, squid.
    Write a program to calculate the total quantity of the 3 dishes and issue an invoice with the value of 2000.
    """)
    # Viết chương trình
    (Translation: # Write a program)
    # Tạo danh sách với 3 món ăn: cá, tôm và mực
    (Translation: # Create a list with 3 dishes: fish, shrimp and squid)
    mixed_sau_oi = ['Cá', 'Tôm', 'Mực']
    # Tạo danh sách 3 món ăn với lượng 100g
    (Translation: # Create a list of 3 dishes with 100g quantity)
    mixed_sau_oi = [100, 100, 100]
    # Sắp xếp các món ăn theo thứ tự: cá, tôm, mực
    (Translation: # Arrange the dishes in order: fish, shrimp, squid)
    mixed_sau_oi.sort()
    # Viết chương trình để tính tổng lượng của 3 món ăn và xuất hóa đơn với giá trị 2000
    (Translation: # Write a program to calculate the total amount of 3 dishes and issue a bill with the value of 2000.)
    # Lệ toán hóa đơn
    (Translation: # Invoice payment)
    total = 0
    for i in range(3):
        total += mixed_sau_oi[i] * 100
    # Xử lý hóa đơn
    (Translation: # Bill processing)
    print("Sản phẩm: " + mixed_sau_oi[0] + " - " + mixed_sau_oi[1] + " - " + mixed_sau_oi[2])
    (Translation: "Product: " + mixed_sau_oi[0] + " - " + mixed_sau_oi[1] + " - " + mixed_sau_oi[2])
    print("Tổng tiền: " + str(total))
    (Translation: "Total amount: " + str(total))
<|end_of_text|>
```

Figure 74: Example of text generated when steered using Vietnamese-specific features with $\alpha = 0.4$ in Llama 3.2 1B, with top-p sampling (`seed` $= 52$, `max_new_tokens` $= 512$). The model was prompted with the `[BOS]` token.

| **Targeted Language** | $\alpha$ | **Change** | | | | **Unchange** | | | |
|---|---|---|---|---|---|---|---|---|---|
| | | **Count** | **Incoherent** | **Partially Coherent** | **Coherent** | **Count** | **Incoherent** | **Partially Coherent** | **Coherent** |
| - | - | - | - | - | - | 100 | 1 | 0 | 99 |
| de | 0.4 | 19 | 0 | 7 | 12 | 81 | 3 | 5 | 73 |
| fr | 0.3 | 29 | 0 | 6 | 23 | 71 | 0 | 0 | 71 |
| it | 0.4 | 11 | 1 | 6 | 4 | 89 | 3 | 5 | 81 |
| pt | 0.2 | 17 | 0 | 13 | 4 | 83 | 6 | 12 | 65 |
| hi | 0.175 | 34 | 14 | 14 | 6 | 66 | 1 | 2 | 63 |
| es | 0.5 | 41 | 0 | 7 | 34 | 59 | 0 | 0 | 59 |
| th | 0.375 | 20 | 1 | 19 | 0 | 80 | 4 | 2 | 74 |
| bg | 0.4 | 12 | 1 | 9 | 2 | 88 | 14 | 8 | 66 |
| ru | 0.5 | 30 | 2 | 16 | 12 | 70 | 3 | 2 | 65 |
| tr | 0.25 | 15 | 0 | 3 | 12 | 85 | 6 | 2 | 77 |
| vi | 0.3 | 11 | 0 | 5 | 6 | 89 | 5 | 1 | 83 |
| ja | 0.3 | 10 | 1 | 4 | 5 | 90 | 12 | 1 | 77 |
| ko | 0.4 | 16 | 2 | 9 | 5 | 84 | 9 | 7 | 68 |
| zh | 0.3 | 29 | 1 | 8 | 20 | 71 | 3 | 1 | 67 |

Table 9: Text generation results from Llama 3.2 1B for each language with various scaling factors $\alpha$. Note that without intervention, the generated text from seed 0 to 99 is in English.

| Targeted Language | $\alpha$ | Change | | | | Unchange | | | |
|---|---|---|---|---|---|---|---|---|---|
| | | Count | Incoherent | Partially Coherent | Coherent | Count | Incoherent | Partially Coherent | Coherent |
| en | -1.2 | 8 | 3 | 5 | 0 | 92 | 16 | 6 | 70 |
| de | 0.5 | 28 | 2 | 16 | 10 | 72 | 7 | 6 | 59 |
| fr | 0.4 | 39 | 0 | 23 | 16 | 61 | 1 | 0 | 60 |
| it | 0.5 | 15 | 3 | 12 | 0 | 85 | 13 | 10 | 62 |
| pt | 0.25 | 17 | 1 | 13 | 3 | 83 | 30 | 10 | 43 |
| hi | 0.2 | 49 | 25 | 18 | 6 | 51 | 2 | 2 | 47 |
| es | 0.8 | 44 | 0 | 17 | 27 | 56 | 1 | 0 | 55 |
| th | 0.4 | 49 | 20 | 28 | 1 | 51 | 0 | 1 | 50 |
| bg | 0.5 | 22 | 8 | 13 | 1 | 78 | 30 | 16 | 32 |
| ru | 0.6 | 51 | 9 | 16 | 26 | 49 | 10 | 2 | 37 |
| tr | 0.3 | 26 | 1 | 12 | 13 | 74 | 7 | 3 | 64 |
| vi | 0.4 | 16 | 0 | 5 | 11 | 84 | 16 | 1 | 67 |
| ja | 0.4 | 20 | 3 | 13 | 4 | 80 | 12 | 6 | 62 |
| ko | 0.5 | 46 | 13 | 28 | 5 | 54 | 9 | 4 | 41 |
| zh | 0.4 | 66 | 2 | 25 | 39 | 34 | 4 | 1 | 29 |

Table 10: Text generation results from Llama 3.2 1B for each language with either a negative scaling factor or a higher positive scaling factor of $\alpha$ than the one used in Table 9. Note that when English was steered with $\alpha = -1.2$, the generated text changed to other languages, primarily Chinese. See Figure 75 in Appendix A for examples.

| Language | SAEs Classifier | | | fastText | | | Neurons Classifier | | |
|---|---|---|---|---|---|---|---|---|---|
| | Prec. | Rec. | F1 | Prec. | Rec. | F1 | Prec. | Rec. | F1 |
| bg | 0.990 | 0.966 | 0.978 | **1.000** | 0.964 | **0.982** | 0.264 | **0.974** | 0.415 |
| de | **0.979** | 0.938 | 0.958 | 0.965 | **0.984** | **0.974** | 0.966 | 0.978 | 0.972 |
| en | 0.836 | 0.980 | 0.902 | **0.849** | **1.000** | **0.918** | 0.000 | 0.000 | 0.000 |
| fr | 0.969 | **0.994** | 0.981 | **0.984** | **0.994** | **0.989** | 0.965 | 0.990 | 0.977 |
| hi | 0.990 | **0.984** | 0.987 | **1.000** | 0.978 | **0.989** | 0.493 | 0.978 | 0.656 |
| it | 0.990 | 0.956 | 0.973 | **0.996** | **0.966** | **0.981** | 0.866 | 0.956 | 0.909 |
| ja | 0.998 | **0.994** | 0.996 | **1.000** | **0.994** | **0.997** | 0.543 | 0.970 | 0.696 |
| ko | **1.000** | **0.990** | **0.995** | **1.000** | **0.990** | **0.995** | 0.000 | 0.000 | 0.000 |
| pt | 0.929 | **0.938** | 0.933 | **0.998** | 0.936 | **0.966** | 0.493 | 0.930 | 0.644 |
| ru | 0.986 | **0.998** | 0.992 | **0.994** | **0.998** | **0.996** | 0.000 | 0.000 | 0.000 |
| es | **0.998** | 0.878 | 0.934 | 0.996 | **0.974** | **0.985** | 0.000 | 0.000 | 0.000 |
| th | 0.998 | **0.988** | 0.993 | **1.000** | **0.988** | **0.994** | **1.000** | 0.972 | 0.986 |
| tr | 0.959 | **0.994** | 0.976 | **0.996** | 0.992 | **0.994** | 0.976 | 0.990 | 0.983 |
| vi | **1.000** | **0.988** | **0.994** | **1.000** | 0.986 | 0.993 | **1.000** | 0.520 | 0.684 |
| zh | 0.990 | **0.996** | **0.993** | 0.990 | **0.996** | **0.993** | **1.000** | 0.002 | 0.004 |
| **Macro Avg** | 0.974 | 0.972 | 0.972 | **0.984** | **0.983** | **0.983** | 0.571 | 0.617 | 0.528 |
| **Accuracy** | 0.972 | | | **0.983** | | | 0.617 | | |

Table 11: Results of SAE-based and neuron-based LID derived from Llama 3.2 1B and fastText LID, applied to WiLI-2018 across 15 languages. **Bold** text indicates the performance is equal to or greater than the counterpart.

| Seed | Output |
|---|---|
| 20 | Question:Develop a TypeScript program snippet to Determine Extreme OnlineShopping: Product<br>销量（销量）和销价（销价）成正比和销价成正比的 销量和销价成正比和销价 |
| 29 | Question:Write a TypeScript code snippet to Calculate Low Fitness:step-counting-for-exercise-tracking for Analysis for Experts. Consider the Transparency and implement appropriate if/else or switch/case statements to handle different conditions. Provide detailed comments explaining your control flow quyết định và<\|begin_of_text\|>Question:<br><br>Solution:<br><br>function calculateLowFitness(stepCountedByGender:number[],gender:string[]){<br>let minValue:number=Math.min(...stepCountedByGender)//unpackslice得到最小的值 và傳回一個數值給乘法運算式中的其他值<br><br>switch(true){//switch式中的所有條件<br><br>case minValue===gender==='Male'://如果步數值和傳回值為Male時使用比較運算式中第一個條件<br><br>then{//如果步數值和傳回值為Male時使用算術式中第一個條件<br><br>console.log('男性');{//使用console輸出步數值和傳回值為Male時<br><br>case<br>minValue===gender==='Female'\|\|gender==='Nonbinary'\|\|gender==='Prepregnant'\|\|gender==='Postpregnant'\|\|gender==='Male'\|\|gender==='Female'\|\|gender==='Nonbinary'\|\|gender==='Prepregnant'\|\|gender==='Postpregnant':<br>then{//如果步數值和傳回值 |
| 49 | Question:Design a Python function snippet to Determine Low Nail care: MoisturizinghandsforanalysisforBeginnersusingwhileloop。分析最佳化为分析者使用循环。如何在python中设计一个函数来为分析师提供护理后的分析功能 |
| 67 | Question:Design a Python function snippet to Determine Extreme Transportation: Ride-Sharing Platforms for Analysis for Professionals. Implement if/else or switch/case statements to handle different conditions related to the Consent. Provide detailed comments explaining your control_flow。<br><br>Solution:Hereは、DeterminesExtremeTransportationは、Python関数を定義しています。<br><br>importrandom、datetimeと標準入力から値を読み込みます。<br><br>defDeterminesExtremeTransportation(self,userAge,userGender,userDistance,userFrequency,userConsent,userConsentFrequency,userConsentDistancedataPoints):<br>"userAgeは、ユー ー ーーーˋˋˋˋˋˋˋˋˋˋˋˋˋˋˋˋˋˋˋˋˋˋˋˋˋˋˋˋˋˋˋˋˋˋˋˋˋˋˋˋˋˋˋˋˋˋˋˋˋˋˋˋˋˋˋ |
| 94 | Question:Develop a Python program snippet to Transform High RegularRESpirationRate<br>برايانسريهايهايهايهايهايهايهايهايهايهايهايهايهايهايهايهايهايهايهايهايهايهايهايهايهايهايهايهايهايهايهايهايهايهايهايهايهايهايهايهايهايهايهايهايهايها<br>يهايهايهايهايهايهايهايهايهايهايهايهايهاي |

Figure 75: Example of text generated when steered using English-specific features with $\alpha = -1.2$ in Llama 3.2 1B, with top-p sampling ( `max_new_tokens` = 256). The model was prompted with the `[BOS]` token. Some repeating text at the end of the output was omitted for brevity.

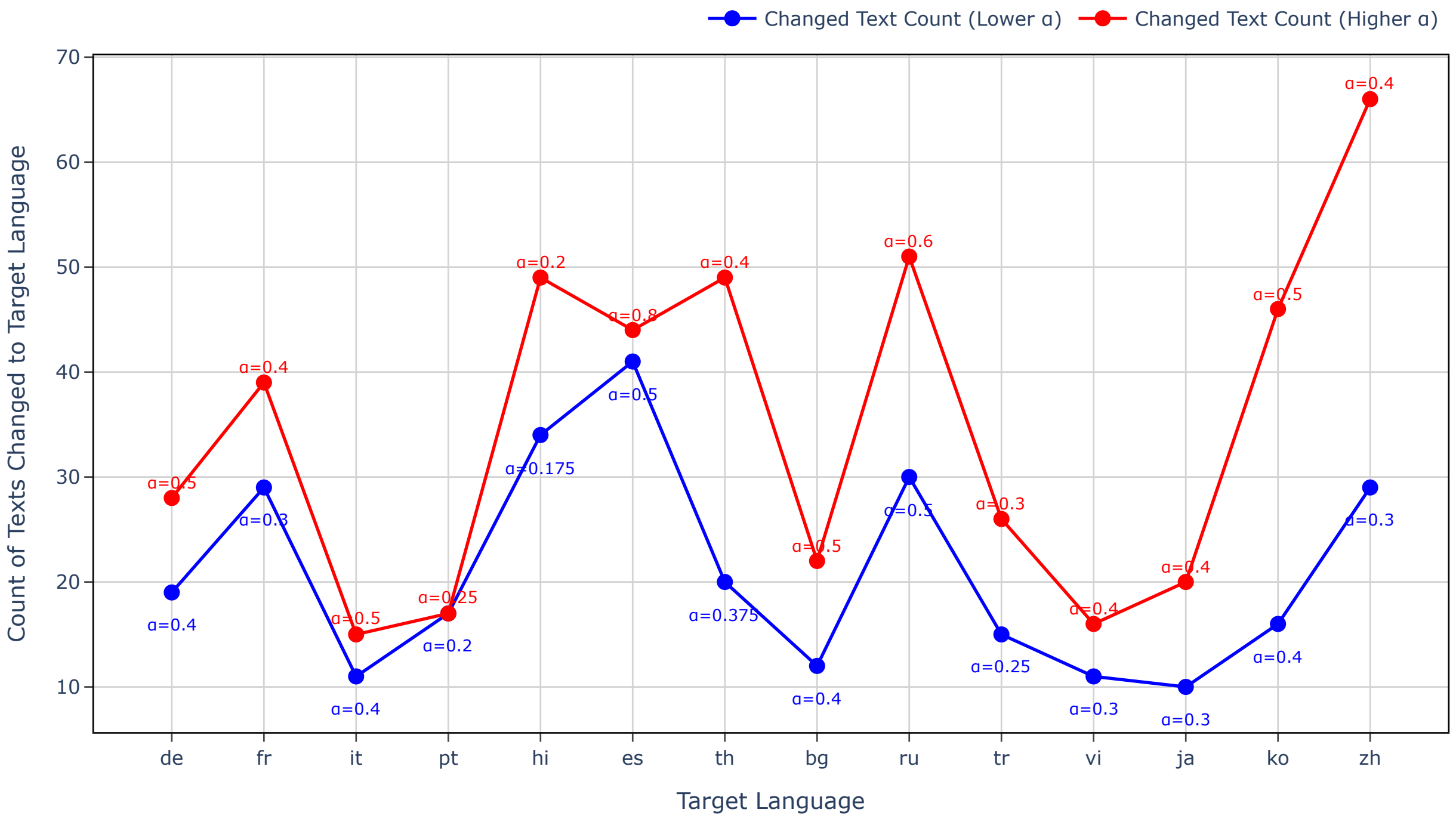


Figure 76: The impact of increasing the scaling factor $\alpha$ on the frequency of changes across different languages in unconditional text generation with Llama 3.2 1B.

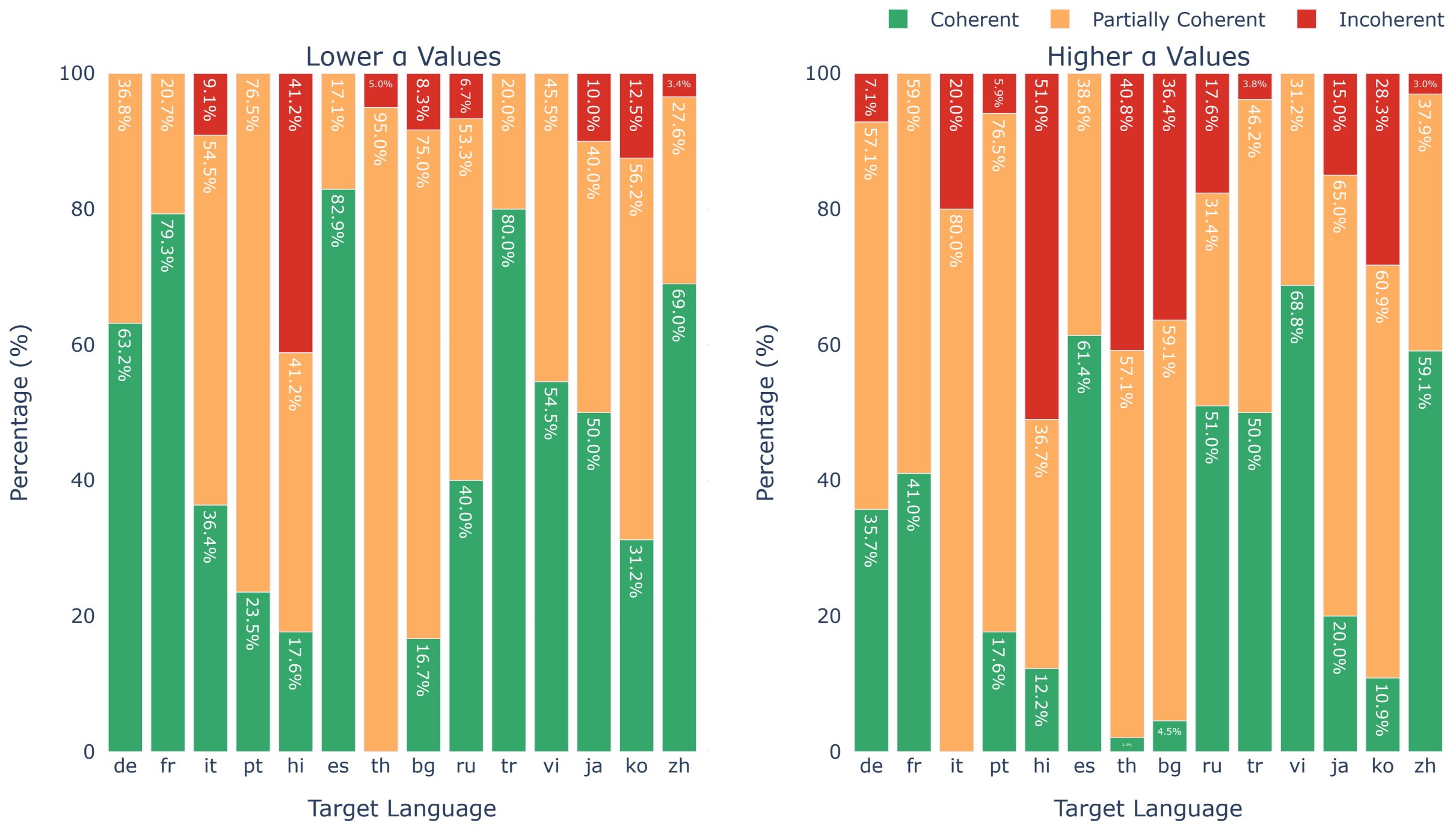


Figure 77: The impact of increasing the scaling factor $\alpha$ on language coherence in changed portions across different languages during unconditional text generation with Llama 3.2 1B.

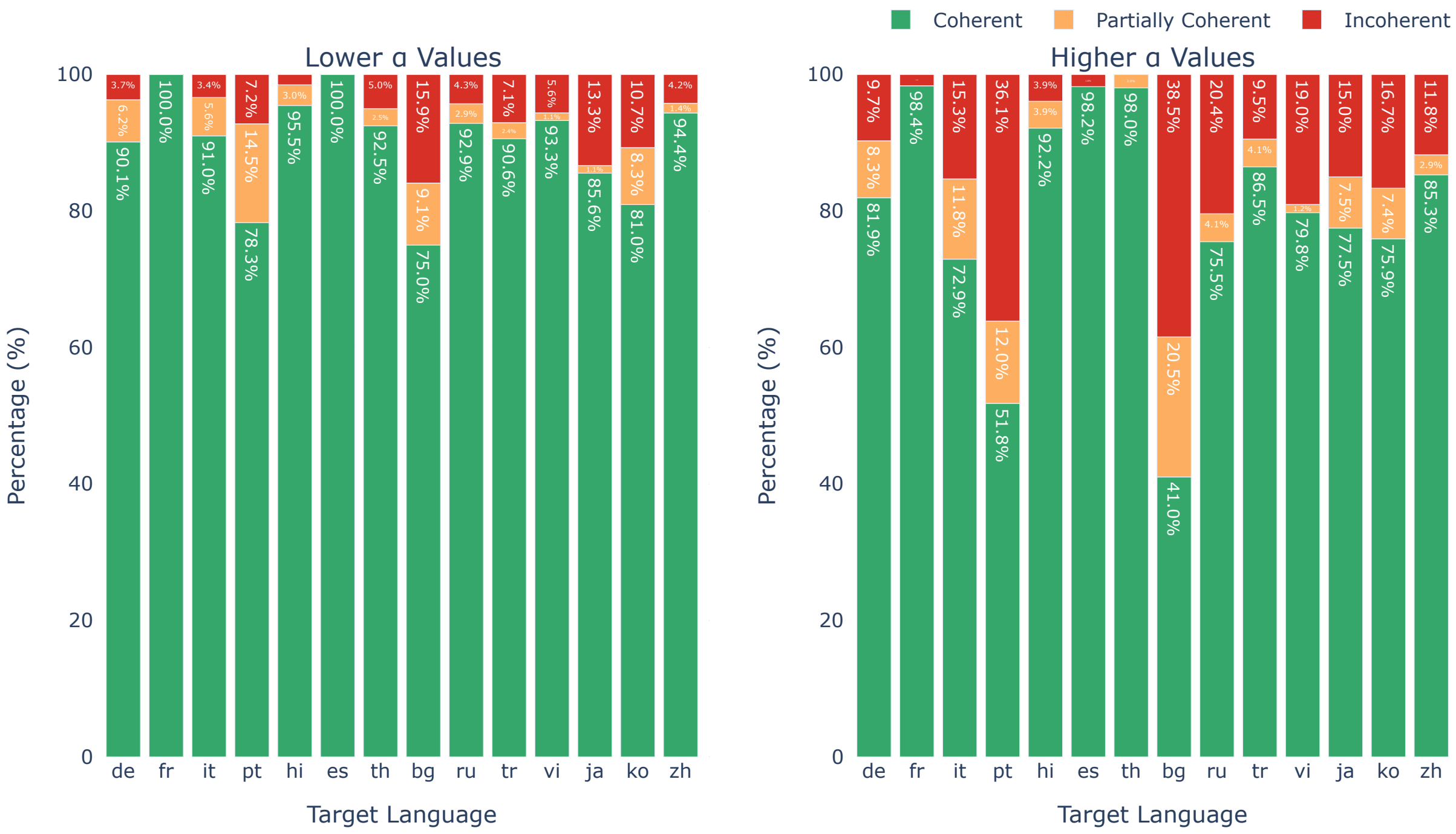


Figure 78: The impact of increasing the scaling factor $\alpha$ on language coherence in unchanged portions across different languages during unconditional text generation with Llama 3.2 1B.

| Layer | Lang | Feature ID | Interpretation | Tokens |
|---|---|---|---|---|
| model.layers.0.mlp | English | 93066 | "Highly frequent activations on punctuation marks, especially commas and spaces, often at the start of sentences or clauses, and across multiple languages, indicating a focus on structural or delimiting tokens in multilingual text." | has, been, the, ,, ,, and, for, ,, as, well, as, co, for, all, of |
| model.layers.0.mlp | English | 128132 | "The highlighted tokens are primarily function words (such as articles, conjunctions, prepositions, and auxiliary verbs) and common noun phrases that serve as structural anchors in sentences. These tokens often mark the boundaries of clauses, introduce or connect ideas, and are essential for grammatical cohesion and meaning flow in both narrative and informational contexts." | has, been, the, ,, ,, and, for, ,, as, well, as, co, for, all, of |
| model.layers.2.mlp | English | 14207 | "The highlighted tokens are often function words, conjunctions, prepositions, or common short phrases that serve as syntactic connectors or modifiers, as well as inflectional or derivational morphemes in various languages. These tokens are crucial for sentence structure, meaning transitions, and grammatical relationships across diverse linguistic contexts." | <\|begin_of_text\|>, the, primary, as, as, co, all |
| model.layers.4.mlp | English | 34138 | "The highlighted text segments are typically multi-word phrases or clauses that convey specific actions, states, or relationships, often providing key contextual or descriptive information within a sentence. These segments frequently include verbs, objects, and modifiers, forming meaningful units that contribute to the overall narrative or informational structure of the text." | been, the, primary, songwriter, ,, guitarist, ,, and, for, as, as, co, -produ, cer, for, all, of, their, recorded |
| model.layers.5.mlp | Bulgarian | 12333 | "The highlighted tokens are predominantly suffixes, prefixes, or inflectional endings in Bulgarian (and occasionally other languages), marking grammatical features such as tense, number, gender, case, or aspect, as well as forming participles, adjectives, and nouns. These morphemes are essential for word formation and grammatical agreement in the language." | primary |
| model.layers.8.mlp | English | 123973 | "The highlighted spans often represent multi-word phrases or clauses that function as cohesive semantic units, such as prepositional phrases, idiomatic expressions, or descriptive noun phrases. These segments frequently provide contextual detail, specify relationships, or add nuance to the main clause, and are typically composed of function words combined with content words to form meaningful, context-dependent chunks." | has, been, the, primary, , |
| model.layers.11.mlp | Portuguese | 19293 | "The highlighted tokens are predominantly roots, stems, or affixes within Portuguese words, often marking the beginning or internal structure of nouns, verbs, and adjectives. These segments are morphologically significant, frequently corresponding to common derivational or inflectional morphemes, and are central to word formation and meaning in the language." | Brothers |
| model.layers.11.mlp | Turkish | 37895 | "The highlighted tokens are primarily suffixes, inflections, or short function words in Turkish and German, often marking grammatical relationships such as possession, plurality, tense, or case, as well as prepositions and names in German. These tokens are crucial for the syntactic and semantic structure of sentences in agglutinative and inflectional languages." | co, recorded |
| model.layers.15.mlp | English | 5889 | "The highlighted tokens frequently correspond to structural elements in formal or factual writing, such as numbers, dates, conjunctions, prepositions, punctuation, and proper nouns, which are often used to organize, enumerate, or specify information in lists, references, or descriptive passages." | Hamilton, has, been, the, primary, songwriter, ,, ,, and, vocalist, for, Brothers, ,, as, well, as, cer, for, all, of, their, releases, . |
| model.layers.15.mlp | English | 67343 | "The text contains a wide variety of conversational and narrative fragments, including idiomatic expressions, filler words, discourse markers, and references to people, places, and actions. There is a frequent use of pronouns, conjunctions, and auxiliary verbs, as well as sequences that reflect spoken language patterns such as repetitions, hesitations, and clarifications. Many segments highlight the structure of dialogue, question-answer formats, and descriptive or explanatory statements, often focusing on actions, states, or relationships between entities." | Hamilton, has, been, the, primary, songwriter, ,, ,, and, vocalist, for, Brothers, ,, as, well, as, cer, for, all, of, their, releases, . |

Figure 79: Full feature activation details in Llama 3.2 1B for the sentence "Hamilton has been the primary songwriter, guitarist, and vocalist for Brothers Past, as well as co-producer for all of their recorded releases."

| Layer | Lang | Feature ID | Interpretation | Tokens |
|---|---|---|---|---|
| model.layers.0.mlp | Bulgarian | 123218 | "The text highlights the frequent use of single Cyrillic letters, especially at the beginning of sentences or phrases, functioning as prepositions, conjunctions, or initials in Russian and Bulgarian, often with high activation when capitalized and sentence-initial." | О, га, на, Г |
| model.layers.0.mlp | English | 93066 | "Highly frequent activations on punctuation marks, especially commas and spaces, often at the start of sentences or clauses, and across multiple languages, indicating a focus on structural or delimiting tokens in multilingual text." | (, ;, *, , ., , in, ), und, , ( |
| model.layers.0.mlp | Turkish | 64723 | "Suffixes such as \"ı\", \"lar\", \"ler\", \"lık\", \"lık\", \"im\", \"in\", \"isi\", \"lığı\", \"ları\", \"leri\", \"ını\", \"udur\", \"ildi\", \"ım\", \"isi\", \"arı\", \"alı\", \"isi\", \"lık\" | ир |
| model.layers.1.mlp | Bulgarian | 32578 | "The highlighted tokens are common Bulgarian suffixes used to form adjectives, nouns, and participles, indicating grammatical features such as gender, number, definiteness, and case. These suffixes are essential for word formation and inflection in Bulgarian morphology." | na, на |
| model.layers.1.mlp | Turkish | 84692 | "The highlighted tokens are predominantly prefixes, roots, or short morphemes within words, often marking the beginning or core of nouns, verbs, or adjectives across multiple languages. These segments frequently correspond to meaningful subword units, such as roots, affixes, or syllables, and are often found in proper names, technical terms, or compound words. The pattern reflects a focus on linguistically significant subword structures that contribute to word formation and meaning." | Gir |
| model.layers.2.mlp | English | 14207 | "The highlighted tokens are often function words, conjunctions, prepositions, or common short phrases that serve as syntactic connectors or modifiers, as well as inflectional or derivational morphemes in various languages. These tokens are crucial for sentence structure, meaning transitions, and grammatical relationships across diverse linguistic contexts." | <\|begin_of_text\|>, |
| model.layers.2.mlp | German | 56064 | "The highlighted tokens are primarily German morphemes, word stems, or inflections, often marking noun or verb endings, compound word parts, or grammatical function words. Many are common in German-language text, including noun endings (-en, -heit, -keit, -ung), verb forms (-iert, -te, -en), and connecting words (und, von, in, der, die, das, eine). There is also frequent activation on recurring roots in compound nouns and on inflectional suffixes, reflecting the morphological structure of German." | isch, Juni, Lang, ist, ß |
| model.layers.2.mlp | Thai | 100192 | "The highlighted tokens are often function words, affixes, or short morphemes in various languages, including prepositions, conjunctions, pronouns, and grammatical particles, as well as some high-activation non-standard or corrupted characters. These tokens are typically important for sentence structure, meaning, or language identification." | Ol, O |
| model.layers.2.mlp | Thai | 105874 | "The highlighted tokens are often morphemes, syllables, or short word fragments that serve as meaningful units in various languages, including Thai, Hindi, and others. These units frequently appear at the beginning, middle, or end of words and are important for word formation, inflection, or conveying grammatical and semantic information." | Ol, O |
| model.layers.3.mlp | Bulgarian | 98476 | "The highlighted tokens are predominantly Bulgarian suffixes, pronoun and conjunction forms, and inflectional endings that mark grammatical relationships such as gender, number, case, and verb tense, as well as function words that connect clauses or indicate possession and agency. These elements are essential for the syntactic structure and meaning in Bulgarian sentences." | Alexand |
| model.layers.3.mlp | Italian | 120200 | "The highlighted tokens are often morphemes, suffixes, or name fragments within words, especially in Italian and other Romance languages, marking comparative forms, diminutives, or parts of proper nouns and loanwords. These tokens frequently appear at word endings or within multi-syllabic names and terms." | O |
| model.layers.3.mlp | Spanish | 4518 | "The highlighted tokens are often morphemes, word stems, or affixes from various languages, especially Spanish, marking grammatical or semantic units such as verb endings, noun forms, or prepositions, and sometimes appear in named entities or set phrases." | ga |
| model.layers.4.mlp | Russian | 48343 | "The highlighted tokens are primarily short morphemes, prefixes, or function words in Slavic and related languages, often marking grammatical relationships, word formation, or serving as connectors within or between words and phrases." | Gir |

Figure 80: Full feature activation details in Llama 3.2 1B for the sentence ”Olga Alexandrowna Girja (russisch Ольга Александровна Гиря; * 4. Juni 1991 in Langepas) ist eine russische Schachspielerin und seit 2009 Großmeister der Frauen (WGM).” from layers 0–4.

| | | | | |
|---|---|---|---|---|
| model.layers.5.mlp | Bulgarian | 12333 | "The highlighted tokens are predominantly suffixes, prefixes, or inflectional endings in Bulgarian (and occasionally other languages), marking grammatical features such as tense, number, gender, case, or aspect, as well as forming participles, adjectives, and nouns. These morphemes are essential for word formation and grammatical agreement in the language." | ja, * |
| model.layers.5.mlp | Bulgarian | 30265 | "The highlighted tokens are predominantly suffixes, inflections, or short morphemes in Bulgarian (and some other languages), such as grammatical endings for tense, case, gender, number, or diminutives, as well as common function words and particles. These elements are crucial for the grammatical structure and meaning of words and sentences." | ja, * |
| model.layers.5.mlp | German | 31366 | "The highlighted tokens are primarily German (and some Russian) morphemes, word stems, and inflectional endings, often marking grammatical features such as case, number, tense, or forming compound nouns and verbs. Many are parts of longer words or compounds, reflecting the agglutinative and inflectional nature of German, and are often found at the beginning, middle, or end of words." | *, ist |
| model.layers.5.mlp | Vietnamese | 49501 | "The highlighted tokens are characteristic morphemes, suffixes, or letter clusters commonly found in Romanian and related Eastern European place names, river names, and surnames, often marking grammatical or regional features." | Alexand |
| model.layers.6.mlp | Russian | 121507 | "The highlighted tokens are predominantly morphemes, roots, prefixes, suffixes, or short function words within Russian, Bulgarian, and related Slavic texts. These segments often mark grammatical features, word formation, or serve as connectors, and are frequently found at the beginning, middle, or end of words, reflecting the agglutinative and inflectional nature of these languages." | ль, га, Александ, Г |
| model.layers.6.mlp | Thai | 68498 | "The highlighted tokens are morphemes or syllables at word boundaries across multiple languages, often marking inflections, derivations, or meaningful subword units. These include suffixes, prefixes, or roots that contribute to the grammatical or semantic structure of words." | spi, Gro |
| model.layers.7.mlp | Thai | 70835 | "The highlighted tokens are primarily morphemes, syllables, or short word fragments from multiple languages, especially Thai, Russian, and Bulgarian, often marking the start or core of content words such as nouns, verbs, or adjectives. These fragments frequently appear at the beginning or within important words, indicating their role in word formation and semantic content across diverse scripts and languages." | О, ль, Г, ир |
| model.layers.8.mlp | English | 123973 | "The highlighted spans often represent multi-word phrases or clauses that function as cohesive semantic units, such as prepositional phrases, idiomatic expressions, or descriptive noun phrases. These segments frequently provide contextual detail, specify relationships, or add nuance to the main clause, and are typically composed of function words combined with content words to form meaningful, context-dependent chunks." | ga |
| model.layers.8.mlp | German | 79729 | "The highlighted tokens are primarily German function words, verb forms, noun and adjective endings, and common prefixes or suffixes, often marking grammatical relationships, tense, plurality, or case, as well as frequent collocations and set phrases. These tokens are essential for sentence structure and meaning in German text." | Juni, ist |
| model.layers.8.mlp | Russian | 109317 | "The highlighted tokens are primarily morphemes, word stems, suffixes, and grammatical endings in Russian (and some in English), often marking case, number, tense, or forming nouns, adjectives, and verbs. There is a focus on functional elements that contribute to word formation, inflection, and syntactic structure, as well as on common connective words and phrase boundaries." | ja |
| model.layers.9.mlp | German | 9083 | "The highlighted tokens are primarily function words, verb forms, noun and adjective endings, and common collocations in German sentences, often marking grammatical structure, tense, or case, as well as frequent phrase boundaries and connectors. These tokens are crucial for sentence construction and meaning in German text." | *, Juni, ), ist, eine, ). |
| model.layers.9.mlp | Portuguese | 77250 | "The highlighted tokens are predominantly prefixes, suffixes, or stems within Portuguese words, often marking verb conjugations, noun/adjective forms, or common morphemes. These segments frequently appear at the start or end of words, reflecting morphological structure and inflectional patterns in the language." | eine |
| model.layers.9.mlp | Russian | 59767 | "The highlighted tokens are predominantly Russian word stems, prefixes, and suffixes that form the core of verbs, nouns, and adjectives, often marking grammatical or semantic roles such as actions, objects, or qualities. These morphemes are crucial for word formation and meaning in Russian, frequently appearing at the beginning or within words to indicate tense, aspect, subject, or object, and are central to the structure and interpretation of Russian sentences." | г, О, ль, га, Александ, ров, на, Г, ир, я |

Figure 81: Full feature activation details in Llama 3.2 1B for the sentence ”Olga Alexandrowna Girja (russisch Ольга Александровна Гиря; * 4. Juni 1991 in Langepas) ist eine russische Schachspielerin und seit 2009 Großmeister der Frauen (WGM).” from layers 5–9.

| | | | | |
|---|---|---|---|---|
| model.layers.10.mlp | Bulgarian | 41383 | "The highlighted tokens are predominantly proper nouns, especially names of people, places, and media titles, often appearing in multiple languages and scripts. These tokens frequently occur in contexts involving legal, journalistic, or entertainment references, and are often accompanied by titles, roles, or institutional affiliations." | Alexand, r |
| model.layers.10.mlp | Bulgarian | 43549 | "The highlighted tokens are primarily function words, pronouns, conjunctions, prepositions, and common verb forms in Bulgarian (and some in other languages), as well as frequent morphemes and endings. These elements are essential for sentence structure, grammatical relationships, and meaning, often marking tense, negation, possession, or subordination. The activations focus on the connective tissue of language that enables coherent and contextually appropriate communication." | Alexand, r |
| model.layers.10.mlp | German | 44371 | "The highlighted spans are predominantly German multi-word expressions, noun or verb phrases, and compound words, often marking grammatical constructs, inflections, or idiomatic units. Many are functionally important for sentence structure, such as verb-final clauses, participial forms, or noun compounds, and frequently include suffixes or prefixes that signal tense, plurality, or case. These patterns reflect the morphological richness and syntactic dependencies characteristic of German text." | ist |
| model.layers.10.mlp | Russian | 56594 | "The highlighted tokens are predominantly proper nouns, technical terms, and named entities—such as personal names, place names, and institutional or scientific terms—often in multiple languages. These tokens frequently appear in contexts involving formal identification, attribution, or description of people, locations, organizations, or specialized concepts." | ga, Alexand, row, na, Gir, ja, (, r, uss, isch, O, ль, га, Александ, ров, на, Г, ир, я, , , in, Lang, as, eine, russ, ach, Gro, der, (, W |
| model.layers.11.mlp | Bulgarian | 52975 | "The highlighted tokens are predominantly short function words, prefixes, or single letters from various languages, often marking grammatical relationships, verb forms, or serving as conjunctions and prepositions. These tokens are crucial for sentence structure and meaning, especially in morphologically rich or agglutinative languages." | r |
| model.layers.11.mlp | German | 27401 | "The highlighted tokens are predominantly German morphemes, word stems, and grammatical endings, often marking inflections, compounds, or function words. These tokens frequently appear at the boundaries of words or phrases, indicating their importance for understanding German syntax, morphology, and sentence structure." | ;, *, ., Juni, ), ist, eine, ische, elerin, und, seit, 9, ister, Frauen, ). |
| model.layers.11.mlp | German | 52681 | "The highlighted tokens are primarily German function words, verb forms, pronouns, and common noun or adjective suffixes, especially those marking person, tense, or plurality. There is a strong focus on grammatical structures, such as verb conjugations, modal verbs, and case endings, as well as on frequently used collocations and phrasal patterns in German sentences." | ;, *, ., Juni, ), ist, eine, ische, elerin, und, seit, 9, ister, Frauen, ). |
| model.layers.11.mlp | Hindi | 63085 | "The text contains frequent use of Hindi conjuncts and half-letters, especially the \"्\" (virama) and combinations with \"व\", \"र\", \"य\", and other consonants, which are characteristic of Hindi morphology and word formation, particularly in formal or technical contexts." | ep |
| model.layers.11.mlp | Korean | 102511 | "The highlighted tokens are predominantly Korean noun and suffix forms, often ending with \"함\", \"것\", \"음\", \"법\", \"형\", \"율\", \"량\", \"군\", \"객\", \"면\", \"성\", \"전\", \"편\", \"심\", \"득\", \"상\", \"시간\", and similar morphemes. These tokens typically denote abstract concepts, objects, people, or states, and are frequently used as nominalizers or to form compound nouns, reflecting a pattern of marking key semantic units or grammatical roles in Korean text." | ров |
| model.layers.11.mlp | Russian | 8452 | "The highlighted tokens are predominantly Russian morphemes, words, or short phrases that serve as key semantic or syntactic units within sentences. They include verb roots, noun endings, pronouns, conjunctions, and common collocations, often marking the core meaning, grammatical structure, or transitions in the text. The activations focus on elements that define actions, states, relationships, or important contextual information, reflecting the building blocks of Russian sentence construction and meaning." | О, ль, га, Александ, ров, на, Г, ир |
| model.layers.11.mlp | Russian | 74675 | "The highlighted tokens are primarily Russian morphemes, roots, and affixes that form the core semantic or grammatical structure of words, often marking verbs, nouns, or adjectives, and are crucial for conveying meaning, tense, aspect, or function within the sentence." | О, ль, га, Александ, ров, на, Г, ир |
| model.layers.11.mlp | Turkish | 37895 | "The highlighted tokens are primarily suffixes, inflections, or short function words in Turkish and German, often marking grammatical relationships such as possession, plurality, tense, or case, as well as prepositions and names in German. These tokens are crucial for the syntactic and semantic structure of sentences in agglutinative and inflectional languages." | in, eine, Gro, ß, me, W |

Figure 82: Full feature activation details in Llama 3.2 1B for the sentence ”Olga Alexandrowna Girja (russisch Ольга Александровна Гиря; * 4. Juni 1991 in Langepas) ist eine russische Schachspielerin und seit 2009 Großmeister der Frauen (WGM).” from layers 10–11.

| | | | | |
|---|---|---|---|---|
| model.layers.12.mlp | Bulgarian | 16648 | "The highlighted tokens are predominantly suffixes or stems in Bulgarian that form adjectives, nouns, or participles, often indicating grammatical features such as gender, number, or case, and are especially common in forming complex or compound words. These morphemes frequently appear at the end of words and are central to word formation and inflection in the language." | russ |
| model.layers.12.mlp | Hindi | 13954 | "Suffixes and endings in Hindi, such as े, ं, ा, ों, ें, ां, and ें, are frequently activated, reflecting their grammatical role in verb conjugation, noun/adjective inflection, and pluralization. These endings are crucial for indicating tense, number, gender, and case in Hindi sentences." | * |
| model.layers.12.mlp | Russian | 31110 | "The highlighted tokens are predominantly Russian morphemes, roots, or affixes that form the core semantic or grammatical structure of words, often marking verbs, adjectives, or nouns. These segments frequently appear at the beginning or end of words and are essential for conveying meaning, tense, aspect, or other grammatical features in Russian. The activations focus on these meaningful subword units that are central to word formation and comprehension in the language." | га, ров, на, Г, ир |
| model.layers.12.mlp | Russian | 48228 | "The highlighted tokens are primarily Russian morphemes, function words, and endings that mark grammatical relationships, verb conjugations, noun/adjective declensions, and common syntactic connectors. These tokens are crucial for sentence structure, tense, case, agreement, and linking clauses, reflecting the morphological richness and syntactic dependencies of Russian language." | га, ров, на, Г, ир |
| model.layers.13.mlp | Bulgarian | 91108 | "The text highlights the importance of common Bulgarian prefixes such as \"из\", \"под\", \"раз\", \"съ\", \"при\", and others, which frequently appear at the beginning of words to form verbs, nouns, or adjectives. These prefixes are key morphological markers that modify the meaning of root words and are highly salient in the structure and semantics of Bulgarian language." | Александ |
| model.layers.13.mlp | Russian | 29244 | "The highlighted tokens are primarily morphemes, roots, prefixes, and suffixes within Russian words, often marking word formation, inflection, or derivation. These segments frequently correspond to meaningful subword units that contribute to the grammatical or semantic structure of the word, such as case endings, diminutives, verb aspects, or noun/adjective formation." | ль, га, Александ, ров, на, Г, ир, я |
| model.layers.13.mlp | Russian | 40052 | "The highlighted tokens are predominantly Russian (and some other Slavic) word endings, suffixes, and inflections that indicate grammatical case, number, gender, verb tense, or derivational morphology. These endings are crucial for understanding syntactic roles and semantic relationships in the text, as well as for identifying noun, adjective, and verb forms." | ль, га, Александ, ров, на, Г, ир, я |
| model.layers.13.mlp | Russian | 97322 | "The highlighted tokens are predominantly Russian morphological suffixes, prepositions, conjunctions, and noun/adjective endings that indicate grammatical relationships, case, number, gender, and aspect. These elements are essential for syntactic structure and meaning in Russian, often marking possession, agency, time, and other core grammatical functions." | ль, га, Александ, ров, на, Г, ир, я |
| model.layers.13.mlp | Russian | 115738 | "The important tokens are predominantly Russian morphemes, roots, and affixes that form the core semantic or grammatical structure of words, often marking noun, verb, or adjective stems, as well as common derivational and inflectional endings. These tokens frequently appear at the beginning or within the main body of words, highlighting the morphological building blocks essential for meaning and word formation in Russian text." | ль, га, Александ, ров, на, Г, ир, я |
| model.layers.13.mlp | Spanish | 53088 | "The highlighted tokens are primarily Spanish morphemes, suffixes, and function words, including endings for adjectives, nouns, and adverbs, as well as conjunctions, prepositions, and common connectors. There is also emphasis on punctuation and proper nouns, reflecting structural and grammatical elements essential for sentence construction and meaning in Spanish text." | ) |
| model.layers.13.mlp | Thai | 40808 | "The highlighted tokens are word fragments, often suffixes or inflections, that appear at the end of words across multiple languages, indicating morphological boundaries or grammatical modifications." | uss, russ |

Figure 83: Full feature activation details in Llama 3.2 1B for the sentence "Olga Alexandrowna Girja (russisch Ольга Александровна Гиря; * 4. Juni 1991 in Langepas) ist eine russische Schachspielerin und seit 2009 Großmeister der Frauen (WGM)." from layers 12–13.

| | | | | |
|---|---|---|---|---|
| model.layers.14.mlp | Bulgarian | 14651 | "The highlighted tokens are predominantly Bulgarian morphemes, suffixes, and inflections that mark grammatical features such as tense, number, gender, case, and aspect, as well as common function words and short stems. These elements are crucial for the syntactic and semantic structure of Bulgarian sentences, often appearing at the ends of words or as short connecting words, reflecting the language's rich inflectional morphology." | я |
| model.layers.14.mlp | German | 312 | "The highlighted tokens are predominantly German suffixes and word endings that form nouns, adjectives, and participles, often indicating grammatical features such as case, number, gender, or degree, as well as forming compound words and inflections." | elerin |
| model.layers.14.mlp | Hindi | 27401 | "Single uppercase letters or digraphs, often at the start of a word or name, are highlighted across multiple languages and scripts, frequently marking proper nouns, initials, or abbreviations." | О, Г, ;, *, ) |
| model.layers.14.mlp | Hindi | 108954 | "The highlighted tokens are common Hindi suffixes and vowel diacritics, such as \"ा\", \"ी\", \"े\", \"ै\", and \"ें\", which are essential for inflection, tense, plurality, and grammatical agreement in Hindi words. These suffixes frequently appear at the end of verbs, nouns, and adjectives, marking grammatical features and contributing to the overall meaning and structure of sentences." | О, Г, ;, *, ) |
| model.layers.14.mlp | Russian | 44860 | "The highlighted tokens are predominantly Russian morphemes, roots, and affixes that form the core semantic or grammatical structure of words. These include verb and noun roots, derivational and inflectional suffixes, and key word segments that contribute to meaning, tense, aspect, or case. The pattern reflects a focus on the internal morphological structure of Russian words, emphasizing the parts that carry essential lexical or grammatical information." | ров |
| model.layers.15.mlp | English | 5889 | "The highlighted tokens frequently correspond to structural elements in formal or factual writing, such as numbers, dates, conjunctions, prepositions, punctuation, and proper nouns, which are often used to organize, enumerate, or specify information in lists, references, or descriptive passages." | ga, row, na, ja, Александ |
| model.layers.15.mlp | German | 17109 | "The highlighted tokens are primarily German morphemes, function words, and common suffixes or inflections, as well as high-frequency connectors and pronouns. These include verb and noun endings, conjunctions, prepositions, and pronouns, which are essential for sentence structure and meaning in German text. The activations also emphasize recurring grammatical patterns, such as subordinate clause markers, verb conjugations, and noun/adjective endings, reflecting the importance of syntactic and morphological elements in German language processing." | und, seit |
| model.layers.15.mlp | German | 75403 | "The highlighted tokens are predominantly German morphemes, function words, and common suffixes or inflections, often marking verb forms, noun endings, or grammatical particles. These tokens frequently appear at the boundaries of words or as part of compound constructions, reflecting the morphological richness and syntactic structure of German text." | und, seit |
| model.layers.15.mlp | German | 130833 | "The highlighted tokens are predominantly German function words, inflectional endings, and common noun or adjective suffixes, reflecting grammatical structure such as articles, pronouns, prepositions, conjunctions, and case or number markers essential for sentence construction and meaning." | und, seit |
| model.layers.15.mlp | Russian | 56964 | "The highlighted tokens are primarily Russian morphemes, roots, and affixes that form the semantic and grammatical core of words, often marking key concepts, actions, or relationships within sentences. These include noun, verb, and adjective roots, as well as frequent derivational and inflectional suffixes, reflecting the morphological structure and meaning-bearing elements of the language." | ров |
| model.layers.15.mlp | Thai | 66481 | "The highlighted tokens are primarily Thai morphemes, syllables, or short words that serve as key semantic or grammatical units within sentences. These tokens often mark the beginnings or important parts of compound words, function words, or content words, and are frequently found at the start of phrases, within compound constructions, or as part of inflectional or derivational morphology. The pattern reflects the segmentation of Thai text into meaningful subword units that are crucial for understanding sentence structure and meaning." | Г |

Figure 84: Full feature activation details in Llama 3.2 1B for the sentence "Olga Alexandrowna Girja (russisch Ольга Александровна Гиря; * 4. Juni 1991 in Langepas) ist eine russische Schachspielerin und seit 2009 Großmeister der Frauen (WGM)." from layers 14–15.